\documentclass{egpubl}
\usepackage[T1]{fontenc}
\usepackage{dfadobe}  

\biberVersion
\BibtexOrBiblatex
\usepackage[backend=biber,bibstyle=EG,citestyle=alphabetic,backref=true]{biblatex} 
\electronicVersion
\PrintedOrElectronic

\ifpdf \usepackage[pdftex]{graphicx} \pdfcompresslevel=9
\else \usepackage[dvips]{graphicx} \fi

\usepackage{egweblnk} 

\usepackage{hyperref}
\usepackage{subfig}
\usepackage{amsmath}
\usepackage{listings}
\usepackage{float}
\usepackage{xcolor}
\definecolor{codegreen}{rgb}{0,0.6,0}
\definecolor{codegray}{rgb}{0.5,0.5,0.5}
\definecolor{Maroon}{rgb}{0.95,0,0.5}
\definecolor{RoyalBlue}{rgb}{0.58,0,0.82}
\definecolor{CTlink}{rgb}{0.58,0,0.82}
\definecolor{codepurple}{rgb}{0.58,0,0.82}
\definecolor{backcolour}{rgb}{0.95,0.95,0.92}

\lstdefinestyle{stepstyle}{
	backgroundcolor=\color{backcolour},   
	commentstyle=\color{codegreen},
	keywordstyle=\color{magenta},
	numberstyle=\tiny\color{codegray},
	stringstyle=\color{codepurple},
	basicstyle=\ttfamily\tiny,
	breakatwhitespace=false,   
	breaklines=true,           
	captionpos=b,              
	keepspaces=true,           
	numbersep=5pt,         
	showspaces=false,          
	showstringspaces=false,
	showtabs=false,        
	tabsize=2,
	emph={np},          
	emphstyle=\ttfamily\color{blue!80!black},
	otherkeywords={self},
    keywordstyle=\color{purple},
}
\usepackage{caption}
\title[CAD 3D Model classification by Graph Neural Networks: A new approach based on STEP format]%
      {CAD 3D Model classification by Graph Neural Networks: A new approach based on STEP format}

\author[L. Mandelli \& S. Berretti
]
{
\parbox{\textwidth}{\centering Lorenzo Mandelli $^{1}$\orcid{}
        and Stefano Berretti$^{1}$\orcid{0000-0003-1219-4386} 
        }
        \\
{\parbox{\textwidth}{\centering $^1$University of Florence, Italy
       }
      }
}

\begin{document}


\maketitle
\begin{abstract}
In this paper, we introduce a new approach for retrieval and classification of 3D models that directly performs in the Computer-Aided Design (CAD) format without any conversion to other representations like point clouds or meshes, thus avoiding any loss of information. Among the various CAD formats, we consider the widely used \textit{STEP} extension, which represents a standard for product manufacturing information. This particular format represents a 3D model as a set of primitive elements such as surfaces and vertices linked together. In our approach, we exploit the linked structure of \textit{STEP} files to create a graph in which the nodes are the primitive elements and the arcs are the connections between them. We then use Graph Neural Networks (GNNs) to solve the problem of model classification. Finally, we created two datasets of 3D models in native CAD format, respectively, by collecting data from the Traceparts model library and from the Configurators software modeling company. We used these datasets to test and compare our approach with respect to state-of-the-art methods that consider other 3D formats. 
Our code is available at  \url{https://github.com/divanoLetto/3D_STEP_Classification}


\end{abstract}  

\section{Introduction}
3D model designers spend a large amount of their time searching for the right information during the product design process and most of their design could be created by modifying an already existing Computer Aided Design (CAD) model. For this reason, the retrieval and reuse of CAD models play a very important role in the domain of CAD model management. 
However, huge repositories of CAD models need a priori categorization on engineering data or require significant organization making design reuse a hard task. As a matter of fact, traditionally, 3D CAD model classification and retrieval were achieved by a time-consuming and troublesome human manual process of labeling that is prone to errors. These problems are even more frequent if the models are generated by product development activities and they are so specified by different labeling and tags that would require a difficult process of data harmonization. 
Furthermore, due to the intrinsic complexity of 3D CAD model definition, no rigid and general rules can be applied in model classification. In fact, depending on product origin, the features and parameters of the models may vary significantly.
An automatic classification and retrieval approach is then needed to overcome these difficulties.

In most of the previous works, the 3D model classification problem was approached using machine learning techniques. However, in those methods, 3D data were usually taken from sensors that can scan real objects. 3D data from sensors is typically represented in various formats such as depth images, point clouds, meshes, or grid of voxels, whereas 3D CAD data created with modeling software is typically stored in one of the CAD formats.
The main state-of-the-art methods that work on 3D models are based on models taken from sensors without considering the possibilities of directly using data in CAD formats. In general, the information contained in a CAD model is greater than that of the same model expressed in other 3D formats, as it allows access to all the information stored during model creation. For example, a model with a cylindrical shape can be represented in the CAD format by the equations describing the surface of the cylinder, while in other 3D formats it is represented by a collection of many primitive elements such as point clouds, grid of voxels or triangular meshes. Therefore, the conversion from one CAD format to other 3D formats is achievable simply by sampling respective elements in the model, while the opposite is not always possible due to the loss of information.

In this paper, we focus on models represented in a native CAD format, and we propose a deep learning approach to automatically classify 3D CAD models. In particular, among the various CAD formats, we consider the widely used STEP extension, which represents a standard for product manufacturing information. This format represents a 3D model as a set of primitive elements such as surfaces and vertices linked together. 
In our approach, we exploit the linked structure of STEP files to create a graph in which the nodes are the primitive elements and the arcs are the connections between them. We then use Graph Neural Networks (GNNs) to solve the problem of model classification and retrieval. 
To the best of our knowledge, no past research does exist that tries to exploit the STEP format for this purpose. 
Furthermore, as there are no large datasets of native 3D CAD models, we built two STEP datasets by finding models from the Traceparts online library~\cite{Traceparts}, and by collecting models from the Configurators~\cite{Configuratori} software modeling company. 
Finally, we tested our approach with the collected datasets, and compared it with classic state-of-the-art methods based on 3D data.

The main contributions of this paper are summarized as follows:
\begin{itemize}
    \item We proposed a new retrieval approach that considers data directly in a CAD format by transforming STEP files into graphs, and propose the use of GNNs for graph classification and retrieval;
    \item We constructed two datasets of native CAD models whole elements were collected from the Traceparts open CAD library and from the Configurators~\cite{Configuratori} software company, and organized them into classes;
    \item We tested our approach and compared it with state-of-the-art methods that work on general 3D models. 
\end{itemize}

The remainder of the paper is organized as follows: in Section~\ref{sec:relatedwork}, we summarize methods that address the problem of retrieval and classification of 3D data in general and CAD models in particular. In Section~\ref{sec:approach}, we present our proposed approach; in particular, in Section~\ref{sec:Step2Graph}, we discuss the characteristics of the STEP format and how to convert 3D models to graphs, while in Section~\ref{sec:GraphClassificationRetrieval}, we describe the proposed GNN for comparing and classifying the obtained graphs. In Section~\ref{sec:experiments}, we describe the collected datasets and the experiments carried on them to validate our approach. Finally, in Section~\ref{sec:conclusion}, conclusions and future research directions are reported. 

\section{Related work}\label{sec:relatedwork}
There are not much works in the literature that performed 3D model retrieval or classification by directly operating on CAD models. 
In the following, we first summarize general methods that worked on 3D data dividing them according to the fact they are based on extracting low-level hand-crafted features, or they directly learn from point clouds, voxel grids, or model views. 
Subsequently, we focus on methods specifically designed for CAD models starting from general solutions, then focusing on graph-based methods and approaches that directly work on the STEP format.

\subsection{Feature-based approaches}
A general approach to handling 3D models is to extract some engineered features to summarize the content of the models in vectors (feature-based techniques) or consider the models as polygon meshes and obtain different transformation invariants to measure the similarity among 3D models (shape-based techniques). In both the approaches, the descriptors obtained are then compared or used as input for supervised machine learning techniques to solve the problem of classification and retrieval. 

In~\cite{CADClassification1}, several shape descriptors (\textit{e.g.}, Zernike descriptor~\cite{zernike}) Multi-resolution Reeb graph descriptor~\cite{MRG} and spectral properties) were extracted to represent a model as $n$ numerical attributes, then a nearest neighbor classifier or a Support Vector Machine was applied. In~\cite{CADClassification3}, for each model a shape distribution vector~\cite{ShapeDistribution} was computed by sampling points and creating a histogram from theirs properties. The histograms were then compared with curve matching techniques (\textit{e.g.}, Minkowski $L_N$ distance). 
In~\cite{CADClassification4} and~\cite{CADClassification2}, the LFD (Light Field Descriptor)~\cite{CADClassification4} view-based features were used, assuming that two 3D models belonging to the same class look similar from all viewing angles. Using LFD, features were extracted further (using the Zernike moments and the Fourier descriptor) after 2D images were generated from 3D models through light field projection. The vectors obtained were finally passed to a deep neural network to solve the classification problem. In~\cite{Frequency}, the CAD model files were converted to a frequency domain representation and then the corresponding spectrograms were compared. 

\subsection{Learning-based methods}
\textbf{Voxels:} 
In~\cite{VoxNet}, trying to overcome the feature-based methods, authors developed a new approach by considering the models as 3D grids and processing them through a supervised network that uses 3D convolutions (3D Convolutional Neural Network, CNN). This machine learning end-to-end approach allows the neural network to learn which are the most significant features instead of defining them manually. More recently, in~\cite{ROCA} authors sampled from the CAD models a grid $32^3$ and extracted in the same way a descriptor. 
In~\cite{LightNet}, authors proposed LightNet, a volumetric CNN architecture with a small number of training parameters, to address the real-time 3D object recognition problem by leveraging multitask learning.

\noindent
\textbf{Point-clouds}: 
Point clouds are becoming popular approaches for representing 3D objects. This representation allows feeding point clouds to deep networks directly since it does not need any transformation of the input data, such as voxelization or projection so that any loss of information is avoided. The difficulty lays on the fact that point cloud data is spatially irregular and permutation invariant, which is essentially different from rasterized data as grid of voxels or pixels. 
In~\cite{PointNet} and~\cite{PointNet++},  Qi~\textit{et al.} directly used point cloud data, represented by three coordinates to feed a network able to perform feature transformations and aggregate data points by max pooling. In~\cite{KnowledgeGraph}, authors utilized the PointNet++ model to segment each 3D model into a set of shapes and extracted their features. The $K$-means method was then utilized to construct from each different shape a node of a 3D shape knowledge graph. Then, a graph embedding was performed based on the 3D shape knowledge graph and an entities’ retrieval method was used to handle the 3D model retrieval problem. 
In~\cite{Kd-Network}, authors used a $kd$-tree to represent point clouds. The network performed the operations using a CNNs adapted to the use of data structures as an input. 

\begin{figure*}[ht]
    \centering
    \includegraphics[width=0.85\linewidth]{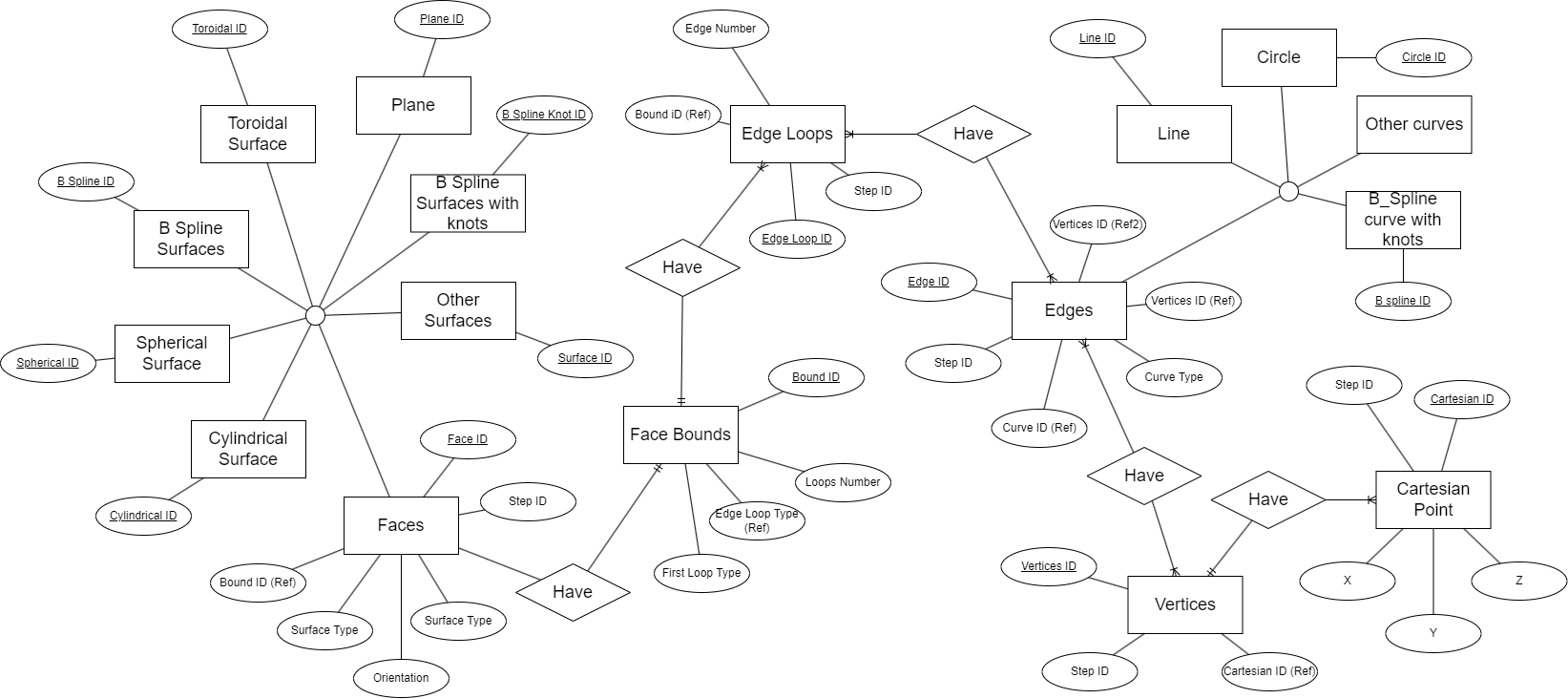}
    \caption{Entity relationship diagram that shows some of the entities defined in the STEP standard.}
  \label{fig:step_schema}
\end{figure*}

\noindent
\textbf{Model views}:
In the Multi View CNN (MVCNN) work~\cite{MVCNN}, exploiting the success of CNNs, authors utilized images taken from 12 or 80 multiple 2D views around a 3D model, aggregated them by a view-pooling layer, and passed the result of the aggregation to a CNN pre-trained on ImageNet to generate a compact shape descriptor. In this way, the problem of 3D model classification was redefined as a view recognition problem. 
In~\cite{PANORAMA}, and in~\cite{PANORAMA_ENN}, a special view of the 3D model, the PANORAMA representation, was obtained and then passed to a CNN to classify the models. 
In the RotationNet approach~\cite{RotationNet}, multiple 2D views of an object were used as input and the network jointly estimated their pose and the object category.

\subsection{CAD model retrieval}
Many works considered the problem of retrieval of CAD models with a sketch-based approach: a 2D sketch is given as an input query and a feature vector is extracted from it using image descriptors. The vector is then used to measure the similarity with the feature vectors extracted from the 3D models stored in the database. These sketch-based methods need therefore to project the 3D model into a 2D-view with the best angle and select an image descriptor capable of matching the view and the hand-drawn sketch. 
Following this idea, Hou~\textit{et al.}~\cite{CADSketch1},~\cite{CADSketch2} proposed a sketch-based retrieval method using three views of a 3D model and combining image descriptors such as spherical harmonics, Fourier transform and Zernike moments. 
In~\cite{CADSketch3}, authors proposed an approach based on sketches and unsupervised learning that extracted geometric information and structural semantic information from the views and the sketch.

Other CAD retrieval methods directly used the 3D structure of the data without resorting to a 2D matching problem. 
Bai \textit{et al.}~\cite{BAI:2010} proposed an approach for partial retrieval of 3D CAD models with the aim of permitting design reuse. 
The multi-mode partial retrieval was achieved by performing multi-mode matching and similarity assessment between the query and the design reusable subparts in the library. 
Osada~\textit{et al.}~\cite{Osada:2008} proposed local 2D visual features integrated with the bag-of-features approach for CAD model retrieval using the data made available in the SHREC’08 CAD model retrieval track.
The SHREC'17 track in~\cite{Hua:2017} had the goal of studying and evaluating the performance of 3D object retrieval algorithms using RGB-D data. This was inspired from the practical need to pair an object acquired from a consumer-grade depth camera to CAD models available in public datasets on the Internet. To support the study, the ObjectNN dataset was proposed with segmented and annotated RGB-D objects from the SceneNN~\cite{HPN:2016} and CAD models from ShapeNet~\cite{ShapeNet}.

\subsubsection{Graph-based analysis}
Several methods used a graph-based approach to analyze CAD models by transforming them into graphs starting from their boundary representation (B-rep).

Tao \textit{et al.}~\cite{Tao:2011} proposed a CAD model retrieval approach by using the Lagrange multiplier method to solve attribute adjacency graph matching. 3D CAD models were first transformed to attribute adjacency graph using B-rep information in 3D CAD models. Then, the vertex compatibility matrix and edge compatibility matrix between the attribute adjacency graph of the target and searched model were calculated, and the measurement of the two models similarity was created from the compatibility matrices. Last, after relaxing objective function as equality constraints, the Lagrange multiplier method was used to solve it.
Tao~\cite{Tao:2015} transformed CAD models into face adjacency graphs (FAGs) by considering only their faces and the edges between them, then applied a graduated assignment algorithm for CAD model retrieval. 
Tao \textit{et al.}~\cite{TAO:2013} proposed a CAD model retrieval method based on local surface region decomposition on FAGs. First, according to the salient geometric features of the mechanical part, the surface boundary of a solid model was divided into local convex, concave and planar regions. Then, region codes were given that describe the surface regions and their links in the CAD model. Finally, the model retrieval was realized based on the similarity measurement between two models’ region codes. 
Following a similar approach, in~\cite{Ma:2019} Ma \textit{et al.} transformed the CAD models into graphs taking into account the faces and the edges between them. The type of the faces and of the edges, and their concavity were also embedded as attributes of nodes and edges, respectively. Then, a two-stage retrieval method was proposed using the TF-IMF (Term Frequency-Inverse Model Frequency) vector method and the ACO (Ant Colony Optimization) between the attribute adjacency graphs.
In~\cite{Li:2011}, Li \textit{et al.} proposed a design reusability assessment method to support retrieval and reuse of CAD models. First, model knowledge was extracted to construct Feature dependency Directed Acyclic Graph (FDAG). Then, these models were progressively simplified based on the design-knowledge-based FDAG. These simplified shapes were finally compared with 3D queries given as inputs, to retrieve CAD models by measuring the general shape similarity.
El-Mehalawi \textit{et al.}~\cite{ELMEHALAWI:2003-I},~\cite{ELMEHALAWI:2003-II} represented CAD components using attributed graphs in which the nodes correspond to the surfaces of the component and the links correspond to the edges of the component. Their graph was based on the STEP physical file of the component. Then, an approach was proposed for retrieving similar designs in a database of mechanical components by abstracting the graphs into some geometric entities and using them to match the similar graphs.
You \textit{et al.}~\cite{You:2010} presented a retrieval architecture that can be utilized to acquire similar mechanical artifacts based on the local feature correspondence. In this work, an attributed graph was created from a B-rep structure by considering the faces and the edge between them. Local feature correspondence was evaluated by identifying the size of the common subgraph from the graph descriptor. This work applied an Independent Maximal Cliques (IMC) detection, and a simulated annealing algorithm to solve the graph-matching problem.
Giannini \textit{et al.}~\cite{Giannini:2017} transformed CAD models into graphs by adding a node for each face and an edge between two faces if they are adjacent or they satisfy some geometric criteria, such as planar with concordant normals. Then, the authors adopted a solution based on the simulated annealing process to detect a maximum clique and solve the graph matching problem. 
Ding \textit{et al.}~\cite{Bo:2014} proposed a novel graph representation of 3D CAD models called Hierarchical Graph (HG). The model descriptors were divided into shape feature descriptors and topology relationship descriptors that can be extracted from the HG. Then, a 3D model retrieval method was proposed based on Genetic Algorithms (GA) and ACO, which were employed to detect the common sub-graph in the corresponding hierarchical graphs of different models.

Compared to the methods mentioned above, we proposed a new approach for the generation of graphs and for the analysis of CAD models. Our strategy for graphs generation is not limited to considering only faces and edges of the B-rep, but also includes every structural element contained in STEP files such as vertices or edge loops. Furthermore, to the best of our knowledge, we are the first to apply a GNN method to the analysis and classification of graphs extracted from CAD models in STEP format.

\section{The proposed approach}\label{sec:approach}
Our approach consists of two steps: first, we transformed the STEP files into graphs as described in Section~\ref{sec:Step2Graph}, then we designed a GNN for the classification and retrieval tasks in Section~\ref{sec:GraphClassificationRetrieval}.

\subsection{From STEP to graph}\label{sec:Step2Graph}
Here, we first illustrate the main characteristics of the STEP format, then we propose a solution to pass from the STEP format to a graph representation. 

\subsubsection{STEP format}
The Standard for the Exchange of Product model data (STEP format) is a widely used ISO standard for data exchange that can represent 3D objects in CAD and related information. 
In this format, a CAD model is expressed by its topological components such as faces, edges or vertices and the connections between them. This collection of connected surface elements, which define the boundary between interior and exterior points is called boundary representation (B-rep).

A generic STEP-file expresses its topological components by a list of instances of standard entities that are stored in the file as lines, as shown in Code~\ref{code:step-example}. Every instance has its own unique $id$ and it is characterized by a series of attributes passed as arguments that can be numeric values, string values or references to other instances. The values of the attributes differentiate between instances of the same entity: two instances of the entity \textit{Point} vary for the numeric values of the coordinates $x$, $y$, and $z$ passed as arguments. 

\begin{lstlisting}[language=C++, caption=Example of a STEP file., label=code:step-example]
ISO-10303-21;
HEADER;
FILE_DESCRIPTION(
/* description */ ('A minimal AP214 example with a single part'),
/* implementation_level */ '2;1');
FILE_NAME(
/* name */ 'demo',
/* time_stamp */ '2003-12-27T11:57:53',
/* author */ ('Lothar Klein'),
/* organization */ ('LKSoft'),
/* preprocessor_version */ ' ',
/* originating_system */ 'IDA-STEP',
/* authorization */ ' ');
FILE_SCHEMA (('AUTOMOTIVE_DESIGN { 1 0 10303 214 2 1 1}'));
ENDSEC;
DATA;
#10=ORGANIZATION('O0001','LKSoft','company');
#11=PRODUCT_DEFINITION_CONTEXT('part definition',#12,'manufacturing');
#12=APPLICATION_CONTEXT('mechanical design');
#13=APPLICATION_PROTOCOL_DEFINITION('','automotive_design',2003,#12);
#14=PRODUCT_DEFINITION('0',$,#15,#11);
#15=PRODUCT_DEFINITION_FORMATION('1',$,#16);
#16=PRODUCT('A0001','Test Part 1','',(#18));
#17=PRODUCT_RELATED_PRODUCT_CATEGORY('part',$,(#16));
#18=PRODUCT_CONTEXT('',#12,'');
#19=APPLIED_ORGANIZATION_ASSIGNMENT(#10,#20,(#16));
#20=ORGANIZATION_ROLE('id owner');
ENDSEC;
END-ISO-10303-21;
\end{lstlisting}

The entities that are considered in the format are geometric primitives such as \textit{points}, complex geometric surfaces such as \textit{splines}, and semantic properties of the model such as \textit{name}, \textit{security level} and domain. 
An entity relationship diagram that shows some of the entities defined in the STEP standard is shown in Figure~\ref{fig:step_schema}.

Compared with other formats for the representation of 3D objects, using the STEP format has the following advantages:
\begin{itemize}
    \item It contains more information than just the geometry of the objects, such as the domain to which they belong to and semantic information;
    \item It allows defining a complex surface through a few nodes, reducing the redundancy of information. For example, a cylindrical surface is defined just by its radius, its height and a spatial point for the center of the base. On the contrary, to represent such a surface in a point cloud or grid format it would be necessary to sample the space in a large number of elements;
    \item Compared to formats that represent an object with multiple 2D views, the STEP format allows considering information on internal elements or, in general, parts of the object not visible from the outside;
    \item In real business contexts, CAD models are often created by assembling smaller components in order to facilitate reusability. From the structure of the STEP file, it is possible to derive its hierarchical structure and use it as additional information. 
\end{itemize}

The presence of geometric implicit surfaces, the semantic information and the hierarchical structure of the CAD files therefore offer greater information than classic formats such as mesh, point cloud or voxels.

\subsubsection{Graph conversion}\label{Graph conversion}
\begin{figure}[ht]
 \centering
  \includegraphics[width=0.8\linewidth]{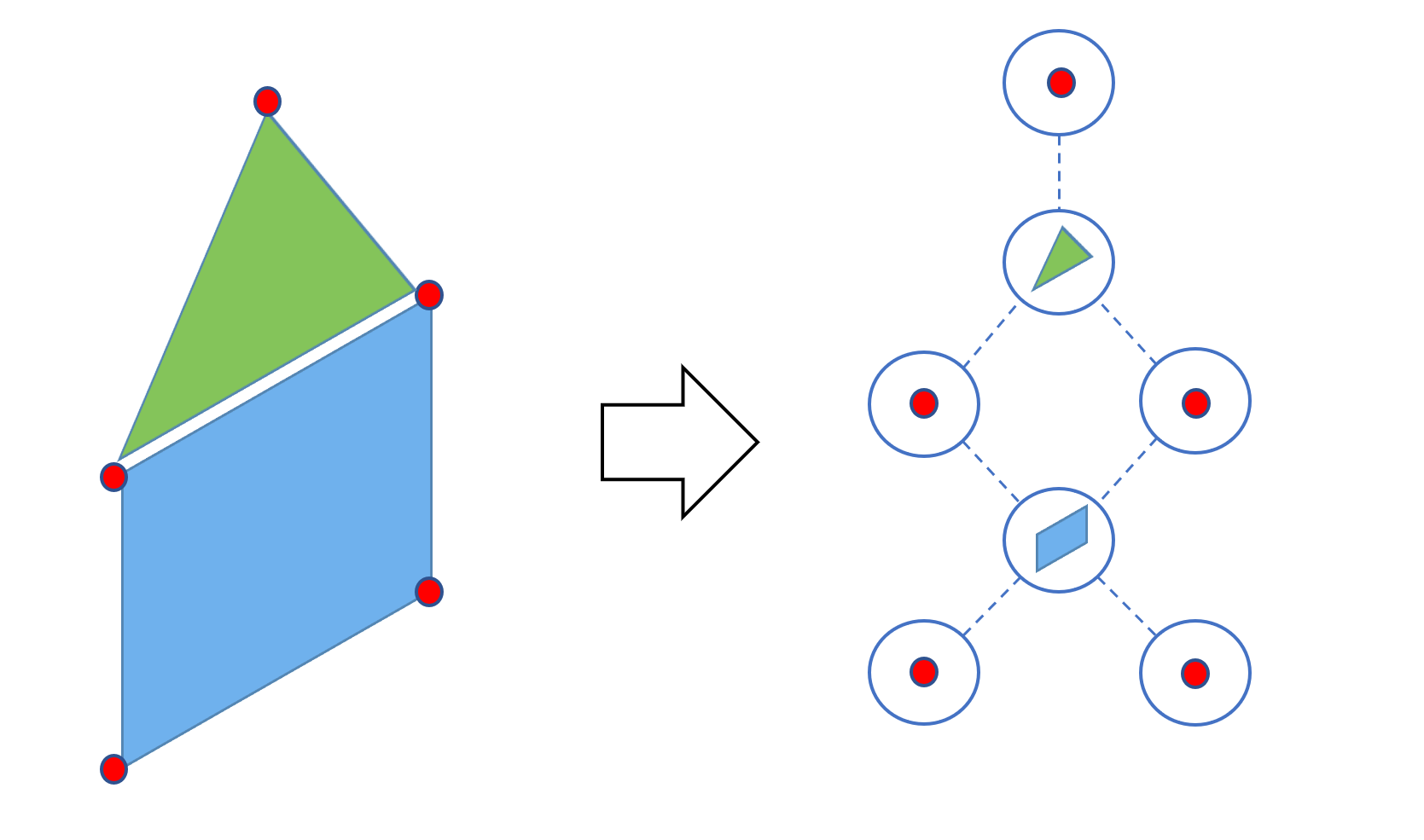}
  \caption{Example of conversion from the geometric information of a STEP file to a graph structure.}
  \label{fig:step2graph}
\end{figure}

From each STEP file, we aim to obtain a graph. This is achieved by parsing the file line by line and each entity instance which is present in each line is coded as a node. Each instance is then characterized by a series of attributes passed as an argument, which can be numeric, strings or references to other instances. The references between instances are encoded as arcs in the graph, numerical values and strings as the attributes of the nodes.
An example of the conversion between a STEP file and a graph is shown in Figure~\ref{fig:step2graph}.

\begin{figure}[ht]
\centering
  \includegraphics[width=0.8\linewidth]{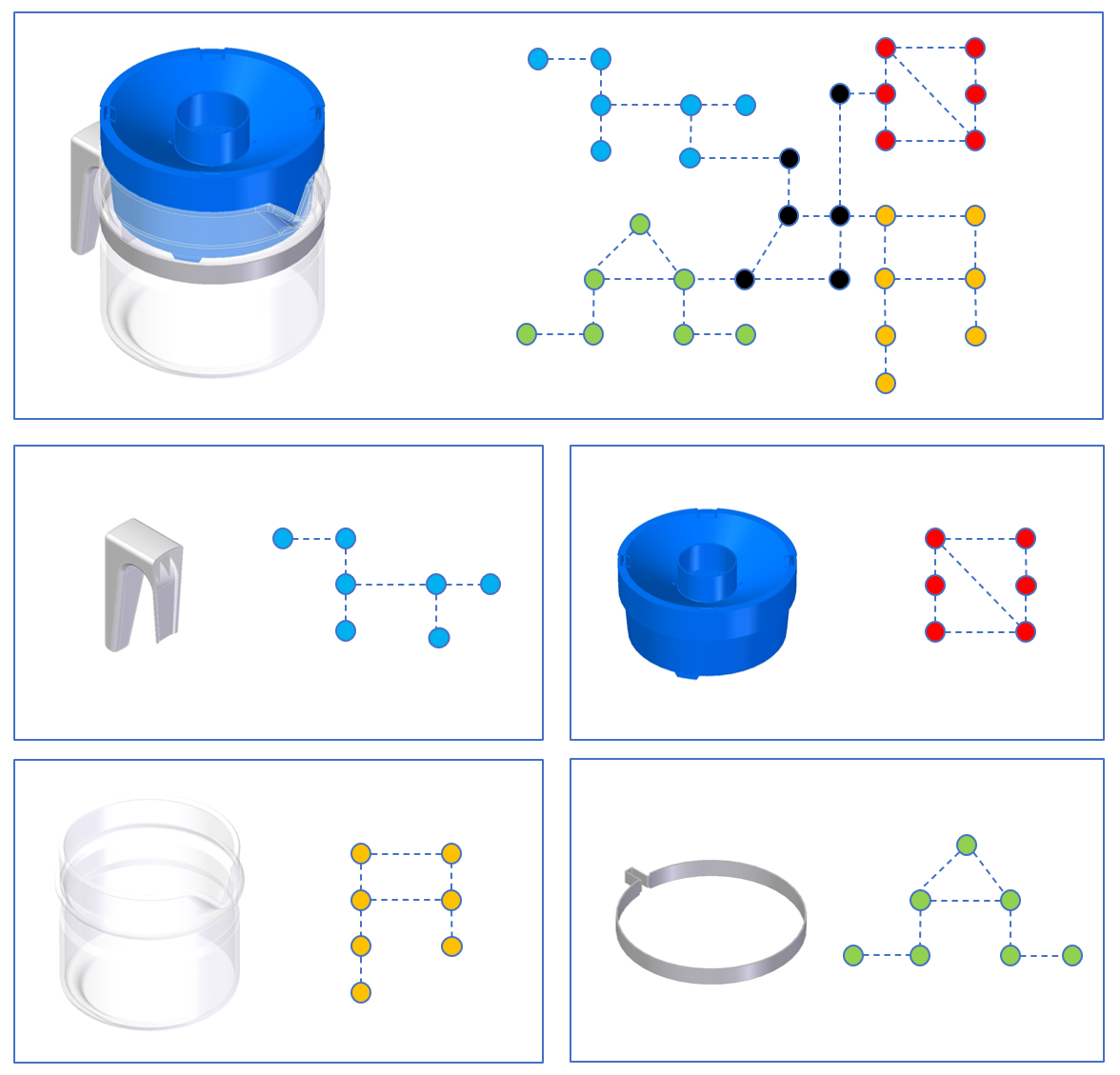}
  \caption{Example of the decomposition into components of a CAD file and its related sub-graphs.}
  \label{fig:step-hierarchy}
\end{figure}

From the STEP file, it is also possible to derive whether the 3D model is made starting from a sub-component structure. Generally, during a CAD modeling phase, the overall model is created by assembling smaller components previously made. For example, a bicycle model can consist of two wheel components and a frame component, which can be composed by even more primitive components. In this way, a single component can be easily reused several times in the same model or in different models without having to create it every time from scratch. 
The hierarchical structure of the STEP files allows us to decompose a model encoded as a graph into several smaller sub-graphs, as it is shown in Figure~\ref{fig:step-hierarchy}. 

Each sub-graph represents a component of the overall model. The breakdown into components provides two advantages:
\begin{itemize}
    \item We can use the additional information of the hierarchical structure of the models;
    \item Comparing the various sub-graphs with each other is much faster than comparing two overall graphs.
\end{itemize}

\subsection{Graph classification and retrieval}\label{sec:GraphClassificationRetrieval}
After creating graphs from 3D CAD models in STEP format, we want to classify them using an end-to-end graph neural network. The classification network can then be used for the retrieval task by extracting feature vectors from intermediate layers of the network and comparing them using various metrics.

\begin{figure*}
\centering
  \includegraphics[width=0.9\linewidth]{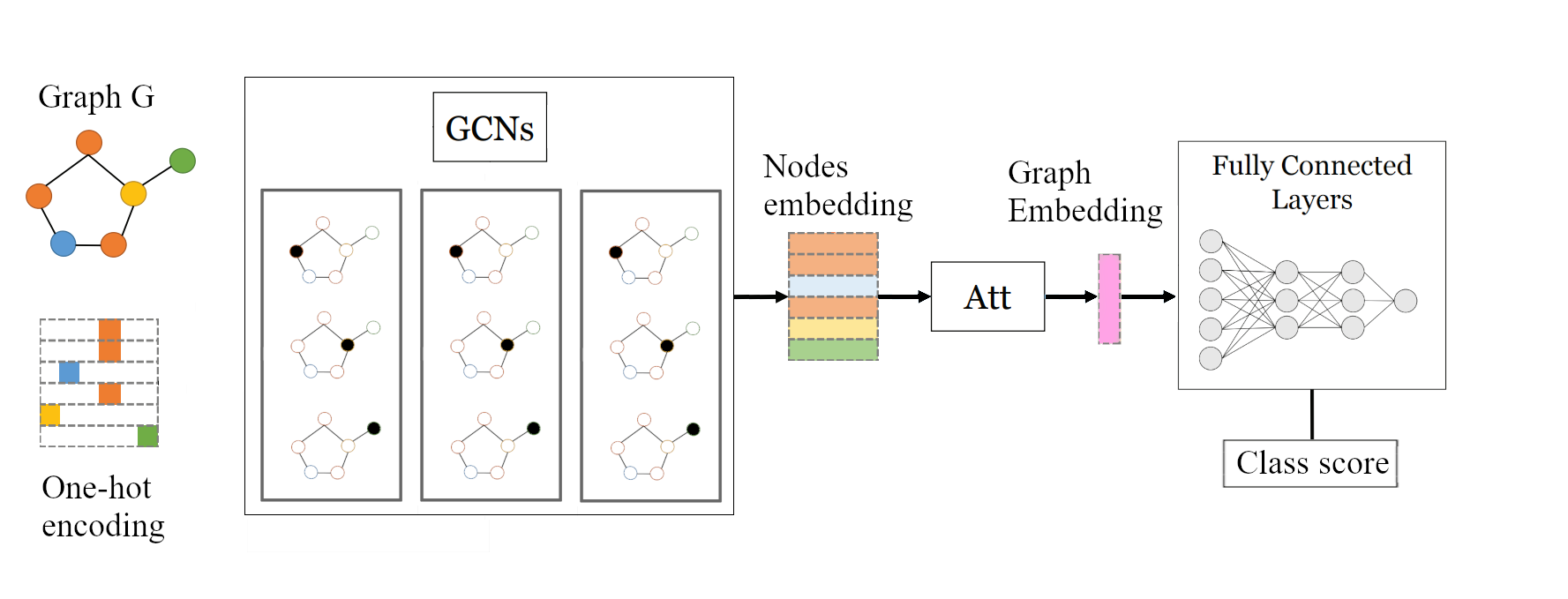}
  \caption{Structure of the proposed GCN network. The GCN layers generate the nodes embedding that are then passed together through an attention mechanism, which generates an embedding of the entire graph. Two fully connected (FC) layers and the softmax function lead to the classification score.}
  \label{fig:GCN}
\end{figure*} 

For this task, we hypothesized that similar models can be defined by a finite set of possible combinations of primitives and therefore of different graphs. For example, in the design phase, a sphere can be realized by a single sphere or by two half-spheres depending on the modeling process. However, it will hardly be realized by infinite representations of unique combinations of primitive elements. Hence, the number of possible graphs describing a geometry will be limited.

For simplicity, in this study that fist explores the potentiality of considering STEP files as graph input data, we did not consider the attributes of the nodes but only their typology. Each node is then represented by a vector of dimension $|P|$ that indicates the one-hot coding of the node type, where $P$ is the set of possible entities defined in the STEP standard. To restrict even more the size of the vector, we considered $P{'} \subseteq P$ representing all the entities contained in the STEP models of the considered dataset.

\textbf{Graph Neural Network}: GNNs are a class of deep learning methods designed to perform inference on data described by graphs in order to provide an easy way to do node-level, edge-level, and graph-level prediction tasks. 

\textbf{Graph Convolutional Networks}: GCNs are a type of GNNs based on an efficient variant of CNN that operates directly on graphs. Let $G(\mathcal{V}, \mathcal{E})$ be an undirected graph with $\mathcal{V}$ the set of nodes and $\mathcal{E})$ the set of the edges. Let $A$ the adjacency matrix of $G$, $I$ the identity matrix and $\tilde{A} = A + I$ the adjacency matrix with self-connections. Formally, the graph convolution can be modeled as:
\begin{equation}
H^{(l+1)} = \sigma(\tilde{D}^{-\frac{1}{2}} \tilde{A}\tilde{D}^{-\frac{1}{2}} H^{(l)} W^{(l)}),
\end{equation}

\noindent 
where $\tilde{D}$ and $W^{(l)}$ are trainable weight matrices and $\sigma$ denotes an activation function of the $l$-th layer. 
$H^{(l)} \in \mathcal{R}^{N\times D}$ is the matrix of activations in the layer, where $N=|\mathcal{V}|$.

This method scales linearly in the number of graph edges, and it is therefore suitable in our case, where graphs can consist of thousands of nodes and arcs. This operation allows us to learn hidden layer representations that encode both local graph structures and the node features. The GCN produces a reduced and meaningful representation of the nodes of a graph.

In order to obtain an overall graph descriptor, the information of the nodes must be summarized into a single vector that represents the whole graph. Instead of performing an unweighted average of node embeddings or a weighted sum, where the weight associated with a node is determined by its degree, we implemented an attention mechanism to let the model learn weights following the ideas in~\cite{SimGNN}. In this way, the nodes that are more important will receive more weights and the model will learn by itself the best weights through the backpropagation steps. 

Let's $U \in \mathcal{R}^{N \times D}$ be the nodes representation, where the $n$-th row represents the embedding $u_n$ of node $n$. For a node $n$ the node attentive mechanism $h$ is calculated as follow:
\begin{equation}
\begin{split}
h & = \sum_{n=1}^N \sigma(u_n^T c)u_n \\
c & = tanh( \frac{1}{N} W \sum_{n=1}^{N} u_n),
\end{split}
\end{equation}

\noindent
where $c$ represents a global context of the graph and it is obtained by a simple average of node embeddings followed by a nonlinear transformation $C \in \mathcal{R}^D$. The function $\sigma$ is the sigmoid function and $W \in \mathcal{R}^{D\times D}$ is a learnable weight matrix. The attention weight for each node is calculated by an inner product between $c$ and its node embedding. In this way, nodes that are more similar to the global context will receive higher attention weights. The resulted vector, which represents the information of the whole graph is then passed to two fully connected layers, reducing the vector size to the number of classes specified by the classification task. A summary scheme of the proposed architecture is shown in Figure~\ref{fig:GCN}.

\section{Experiments}\label{sec:experiments}
In order to evaluate our proposed approach, we first collected two datasets of CAD models in STEP format as discussed in Section~\ref{sec:dataset}. Then, we performed classification and retrieval experiments on those datasets, respectively, in Section~\ref{sec:classification} and Section~\ref{sec:retrieval}. Classification results are also compared with state-of-the-art solutions that work on different data format. Evaluations while varying different parameters and settings for our approach are also reported, 

\subsection{Datasets}\label{sec:dataset}
Large datasets of native CAD models do not exist. While converting the CAD format to other formats such as point clouds, views, voxels or meshes is possible, the reverse process is not since CAD models contain more information than that available in the other formats. Not being possible to use the major 3D model datasets such as ShapeNet~\cite{ShapeNet} or ModelNet~\cite{ModelNet} for our purposes, we created and used for our tests two CAD datasets, the TraceParts and the Configurators ones, and made the first one available at the link \url{https://mega.nz/folder/YfESgYDA#FNFLcCsg8mOkb6O2y0o-tA}.

\textbf{TraceParts CAD dataset}:  
The first dataset consists of 6 classes with 100 CAD models each. The models belong to the STEP 242 application protocol (AP) and were obtained from the free online TraceParts CAD library~\cite{Traceparts}, which stores many STEP models of various categories produced by different companies and with different modeling processes. Unfortunately, the components of the models in the library are not defined, but the models are treated as a single piece. Therefore, it was not possible to exploit the hierarchical composition of the CAD models as discussed in Section~\ref{Graph conversion}.

\begin{figure}[ht]
\centering
\begin{tabular}{cccccc}
\hline
\textbf{Class 0} & \textbf{Class 1} & \textbf{Class 2} & \textbf{Class 3} & \textbf{Class 4} & \textbf{Class 5} \\
\hline \\
\subfloat{\includegraphics[width = .12\columnwidth]{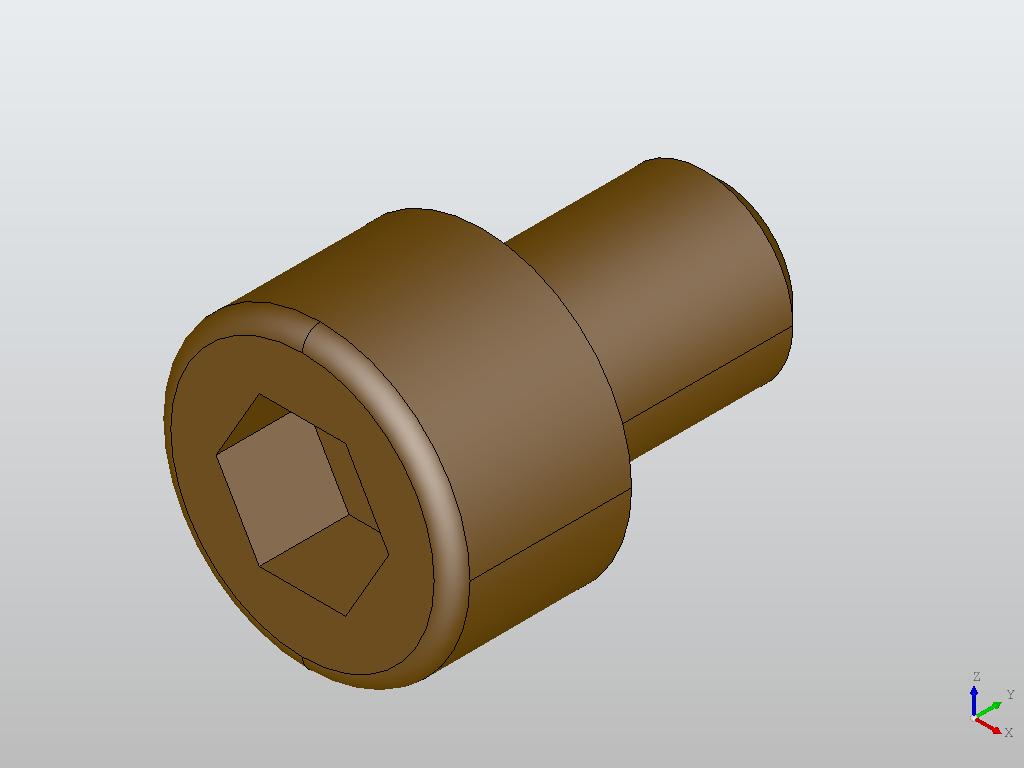}} &
\subfloat{\includegraphics[width = .12\linewidth]{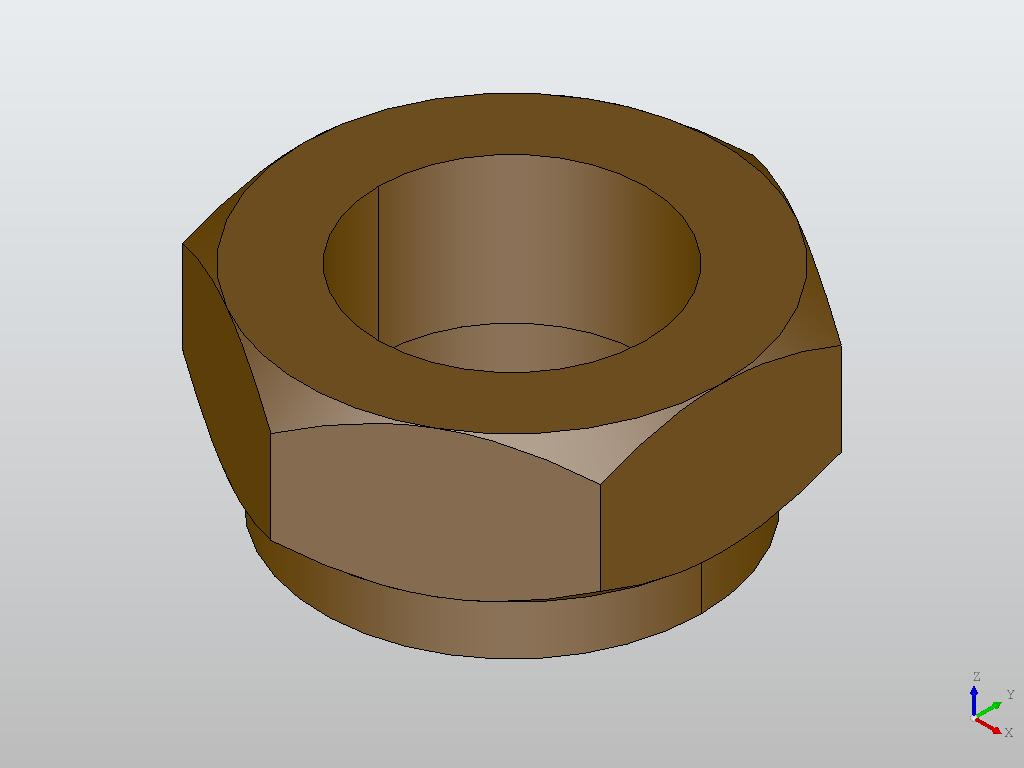}} &
\subfloat{\includegraphics[width = .12\linewidth]{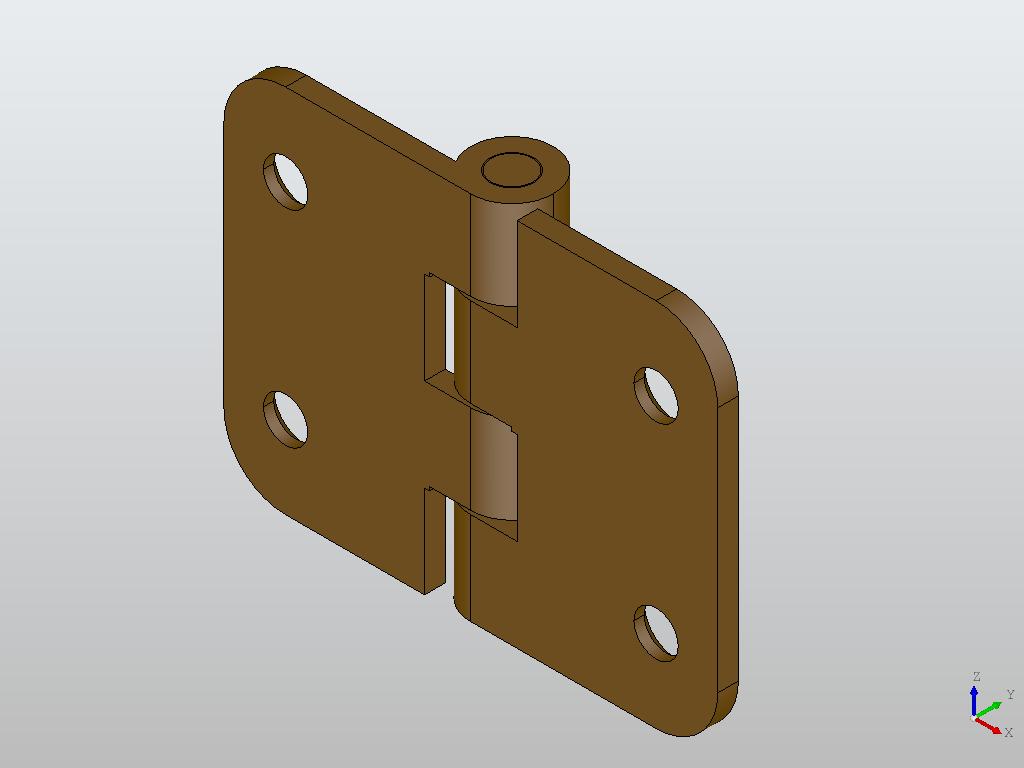}} &
\subfloat{\includegraphics[width = .12\linewidth]{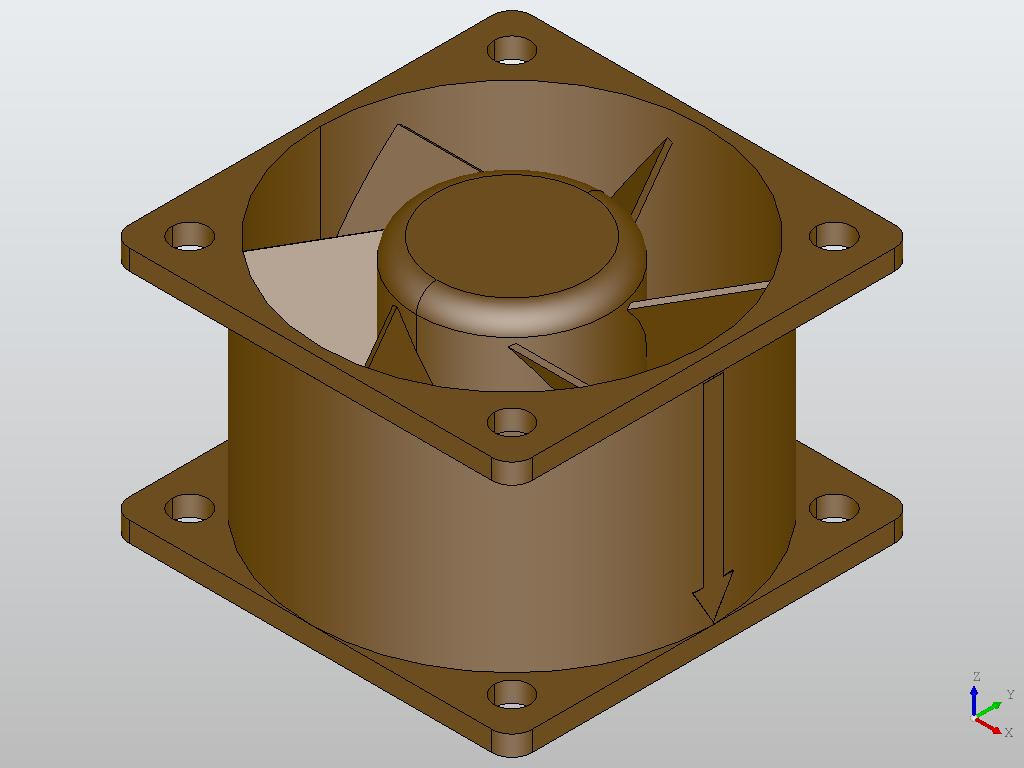}} &
\subfloat{\includegraphics[width = .12\linewidth]{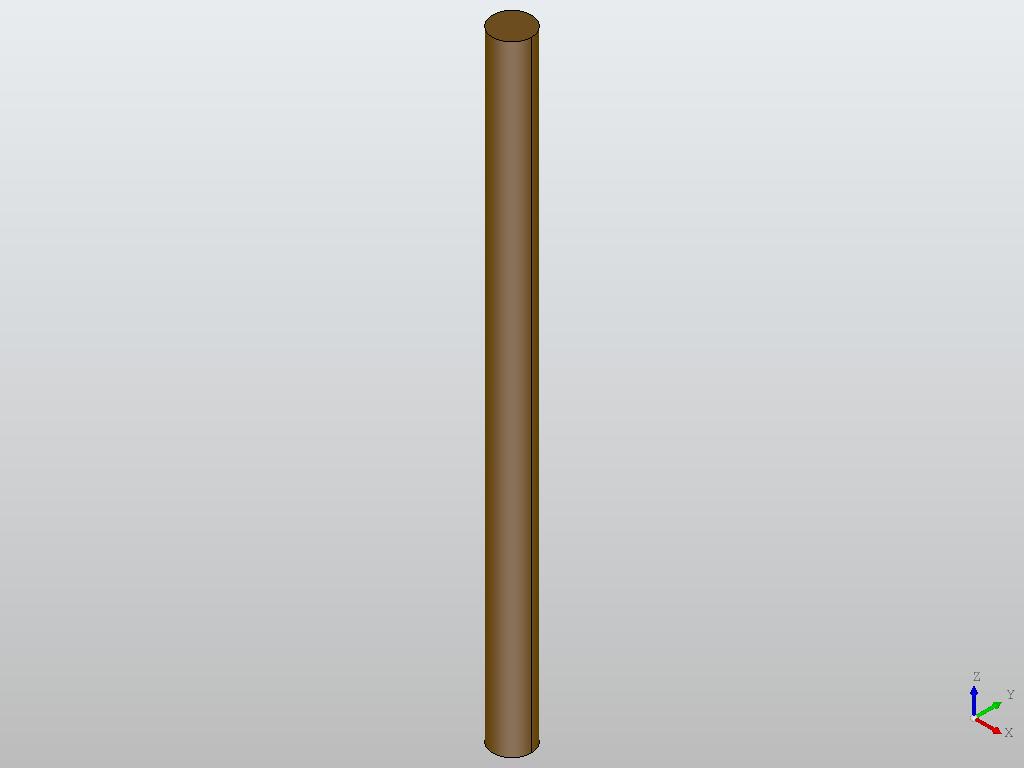}} &
\subfloat{\includegraphics[width = .12\linewidth]{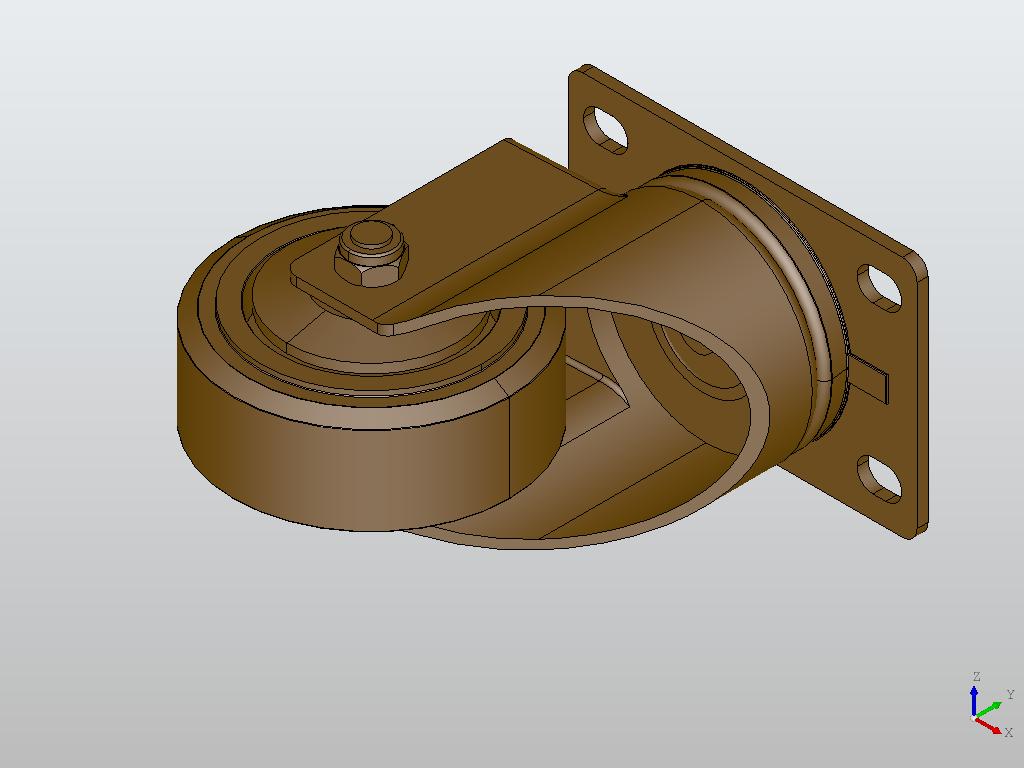}} \\
\subfloat{\includegraphics[width = .12\linewidth]{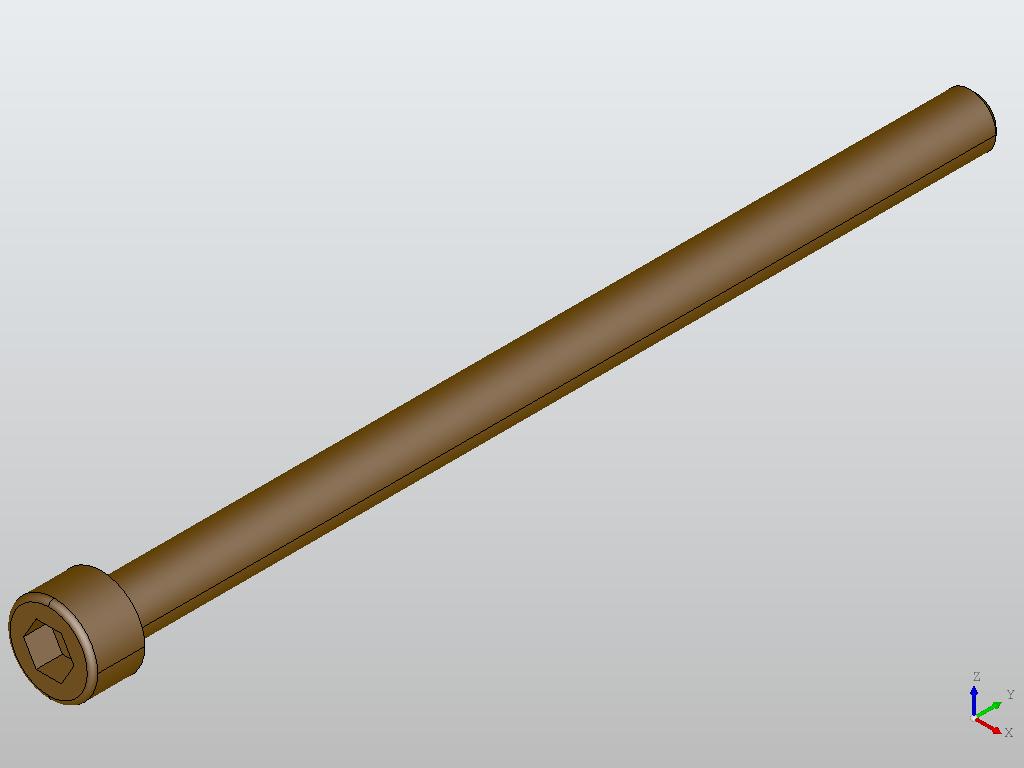}} &
\subfloat{\includegraphics[width = .12\linewidth]{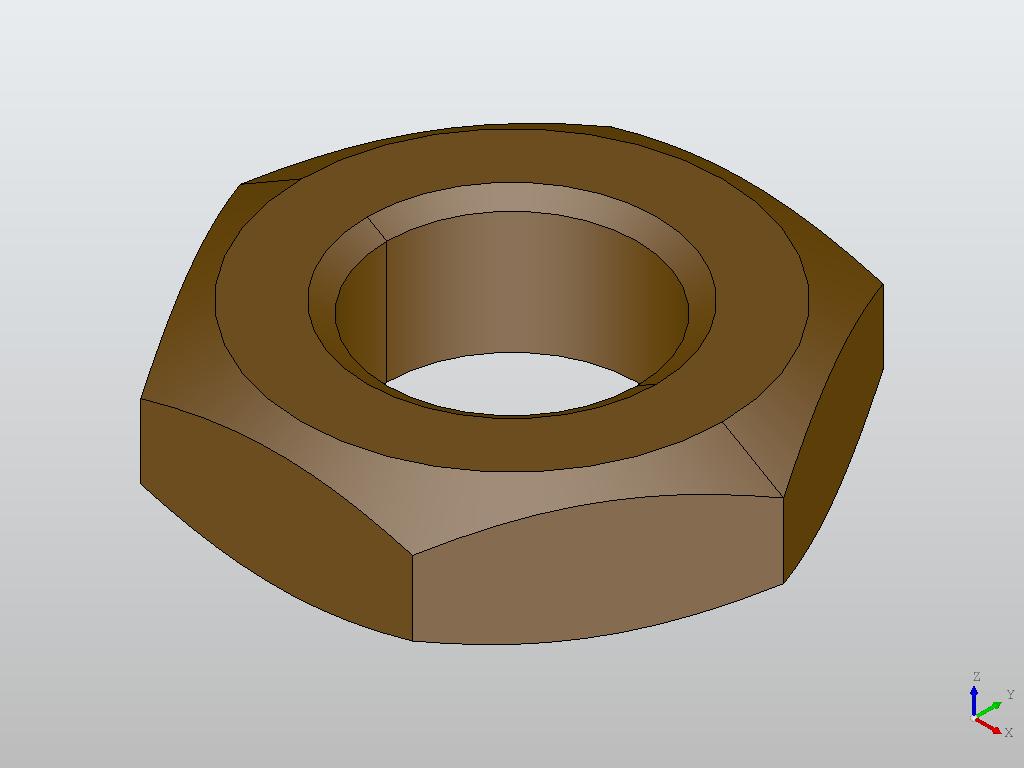}} &
\subfloat{\includegraphics[width = .12\linewidth]{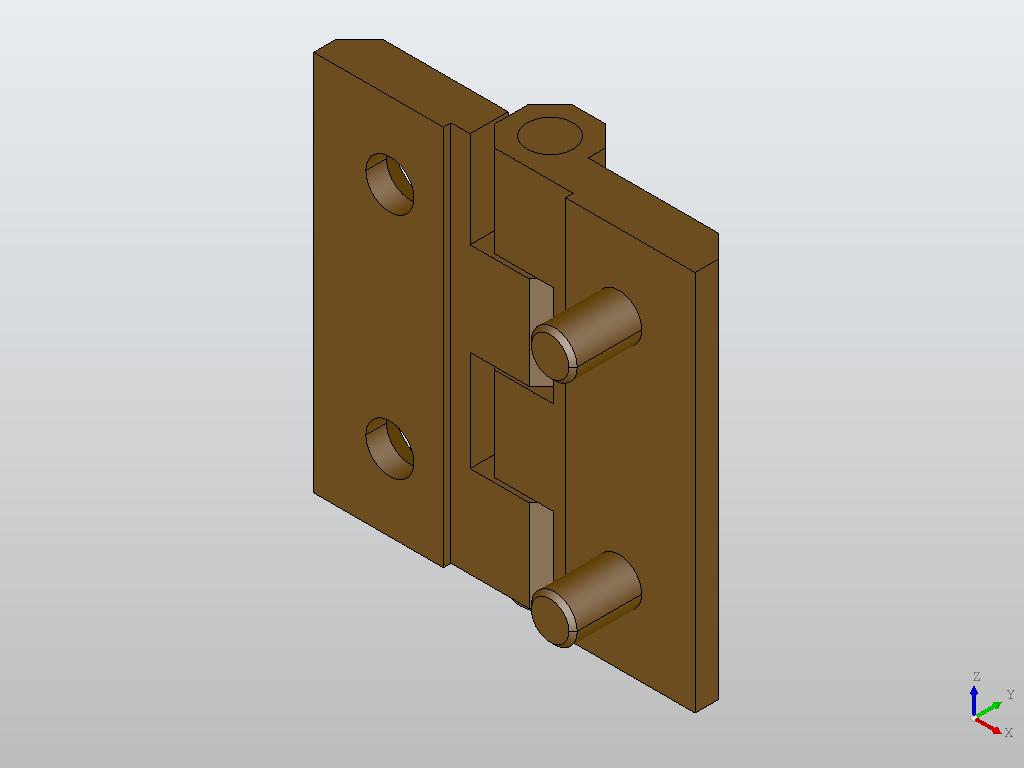}} &
\subfloat{\includegraphics[width = .12\linewidth]{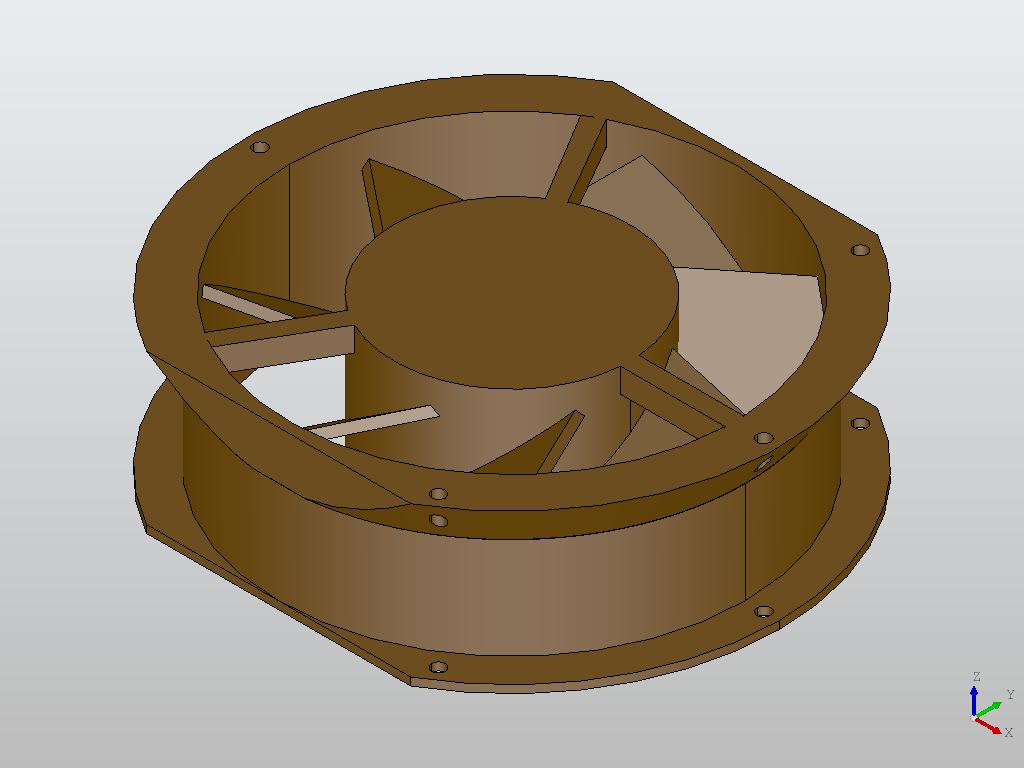}} &
\subfloat{\includegraphics[width = .12\linewidth]{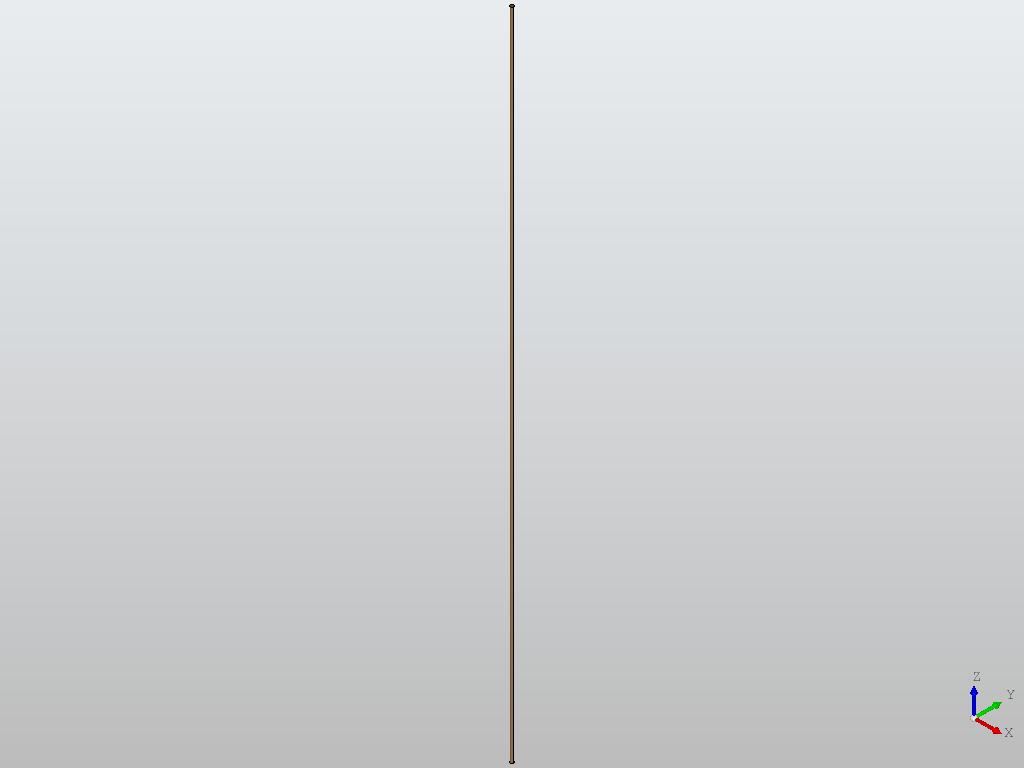}} &
\subfloat{\includegraphics[width = .12\linewidth]{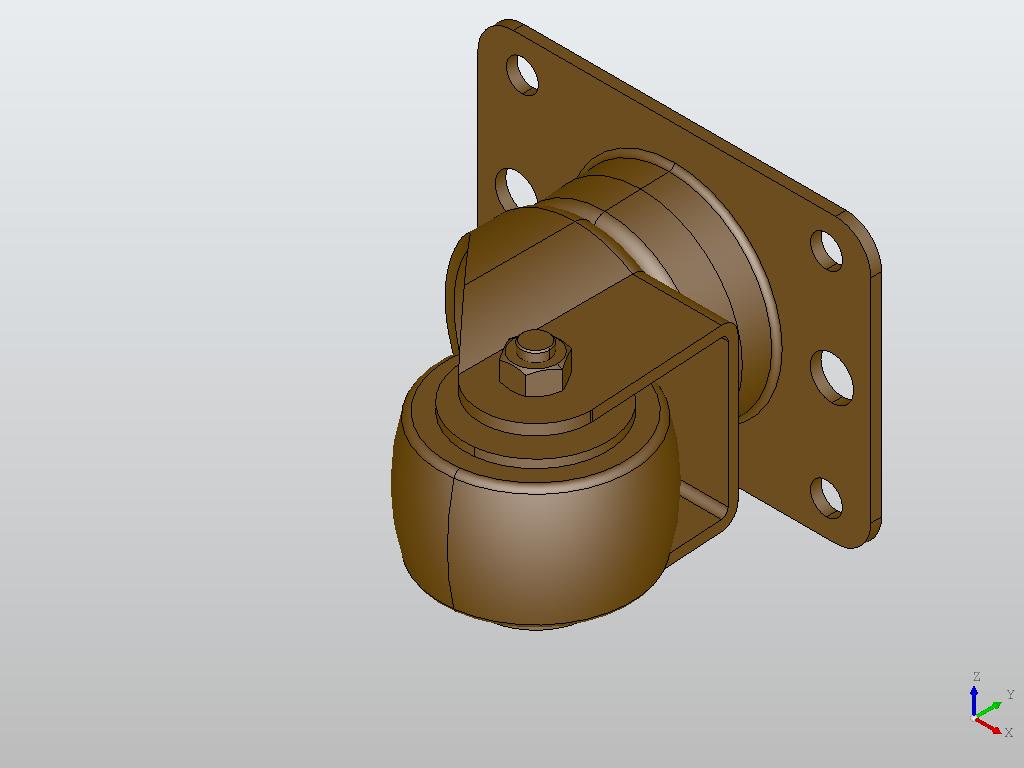}} \\
\subfloat{\includegraphics[width = .12\linewidth]{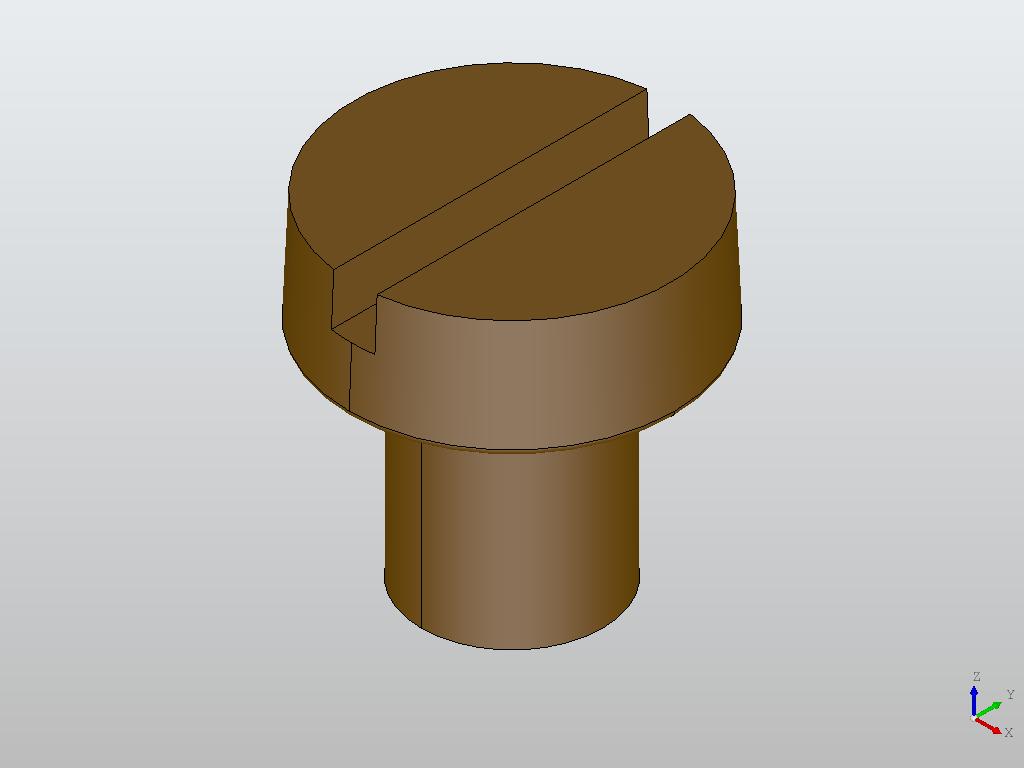}} &
\subfloat{\includegraphics[width = .12\linewidth]{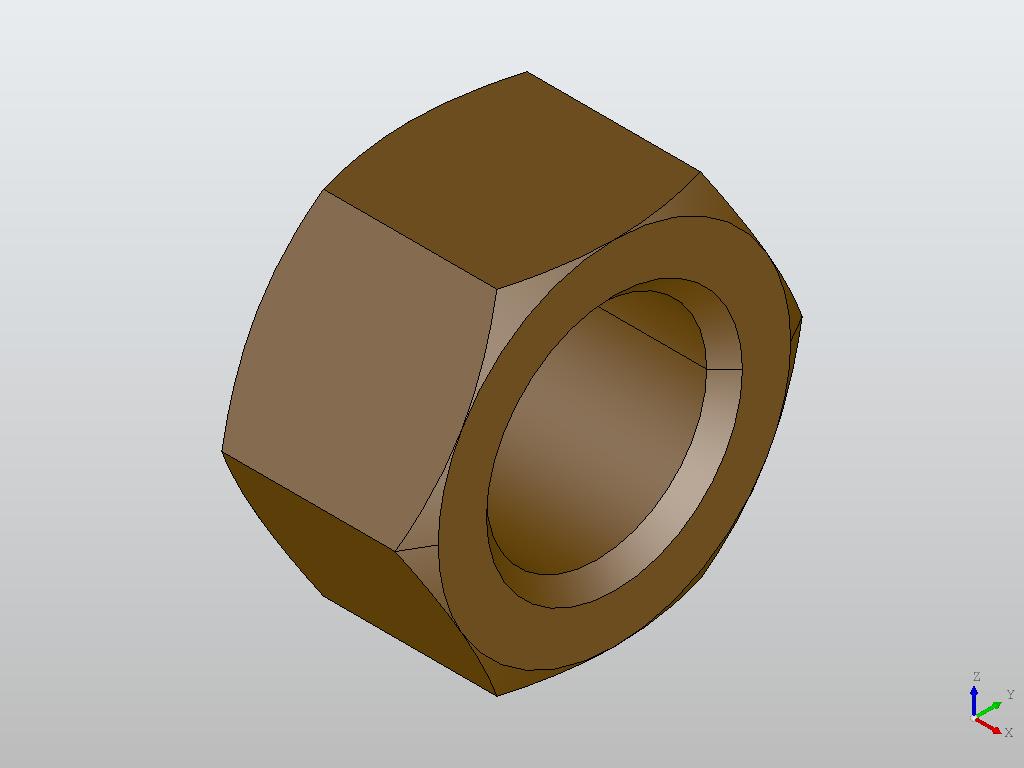}} &
\subfloat{\includegraphics[width = .12\linewidth]{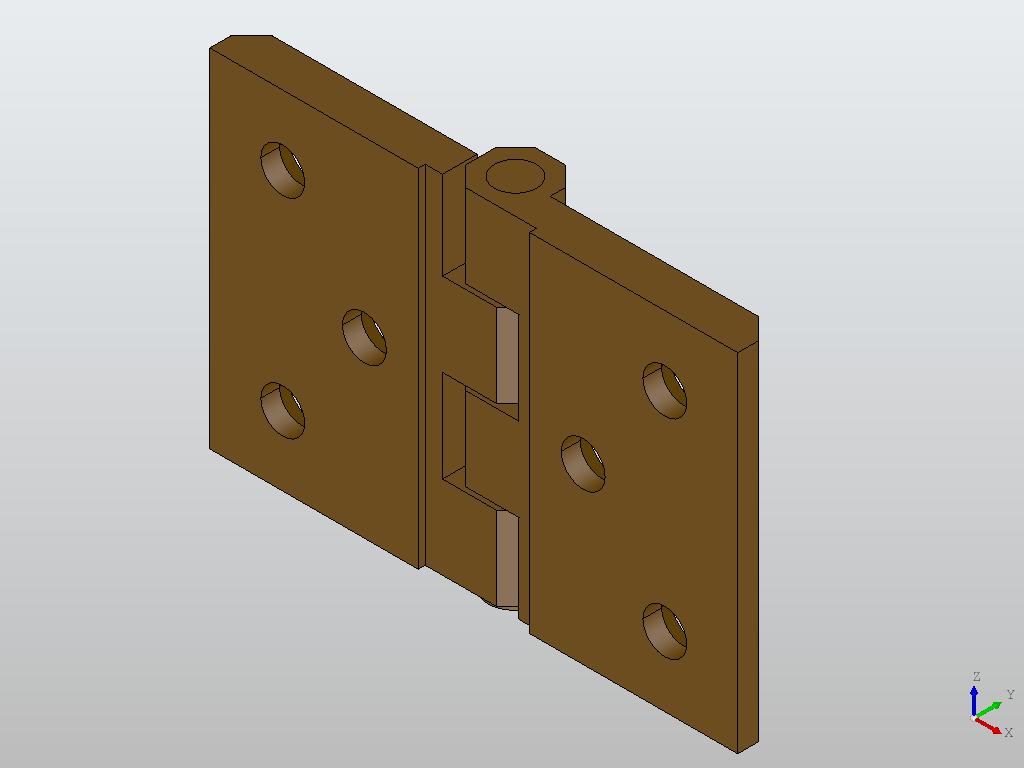}} &
\subfloat{\includegraphics[width = .12\linewidth]{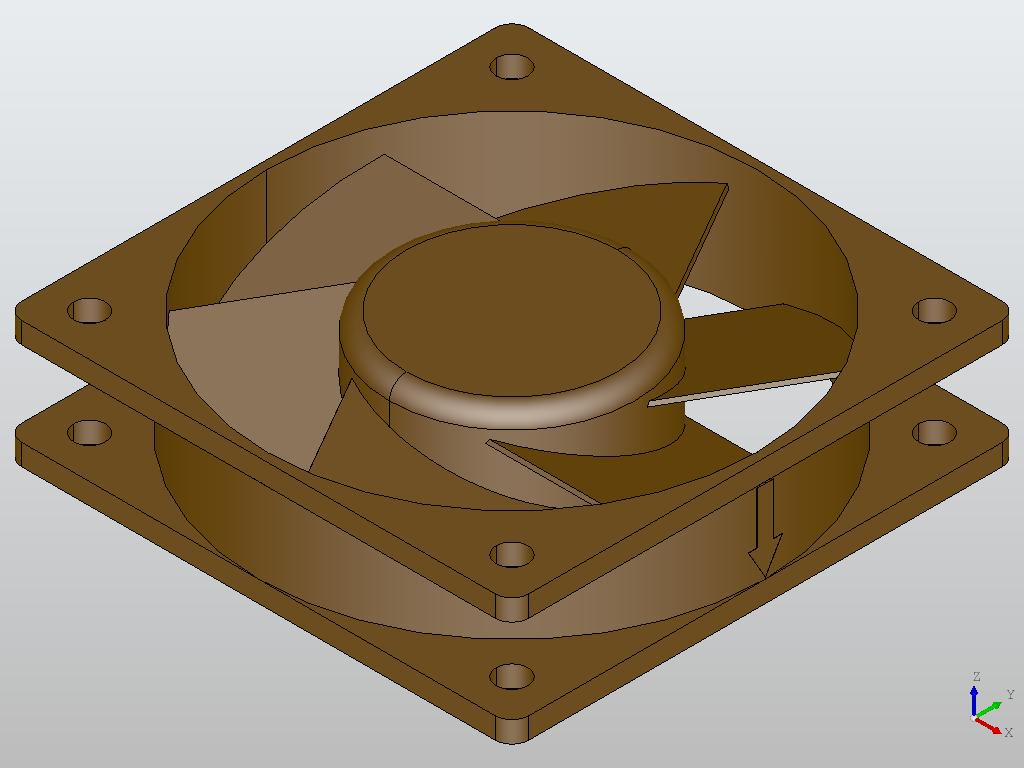}} &
\subfloat{\includegraphics[width = .12\linewidth]{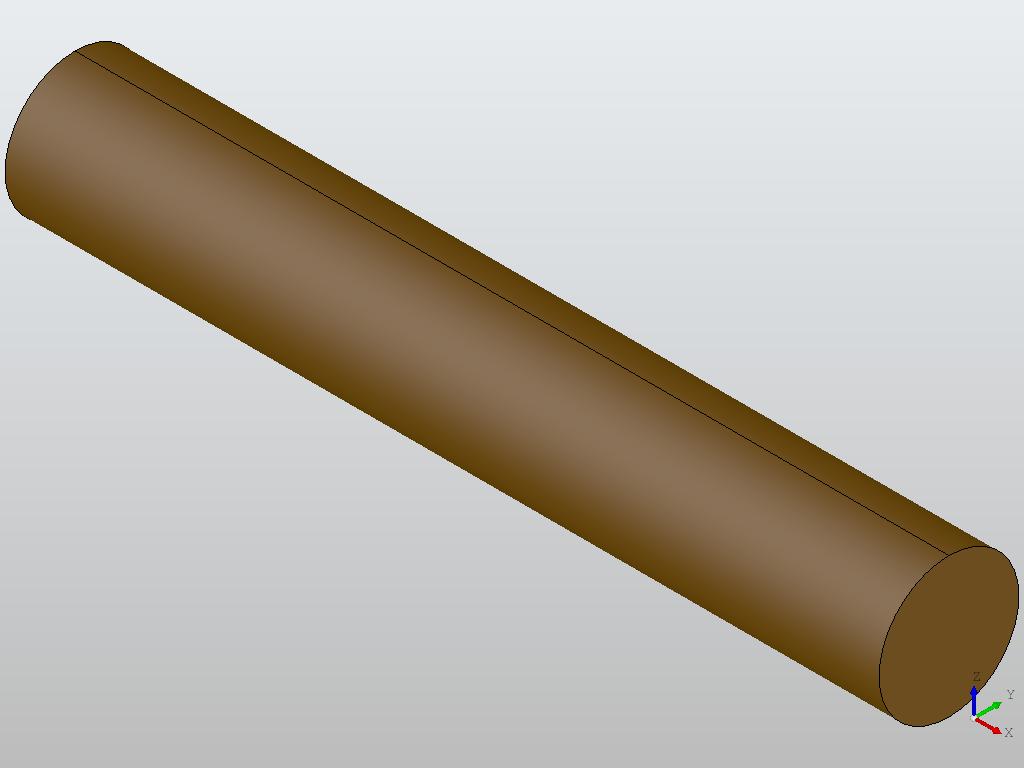}} &
\subfloat{\includegraphics[width = .12\linewidth]{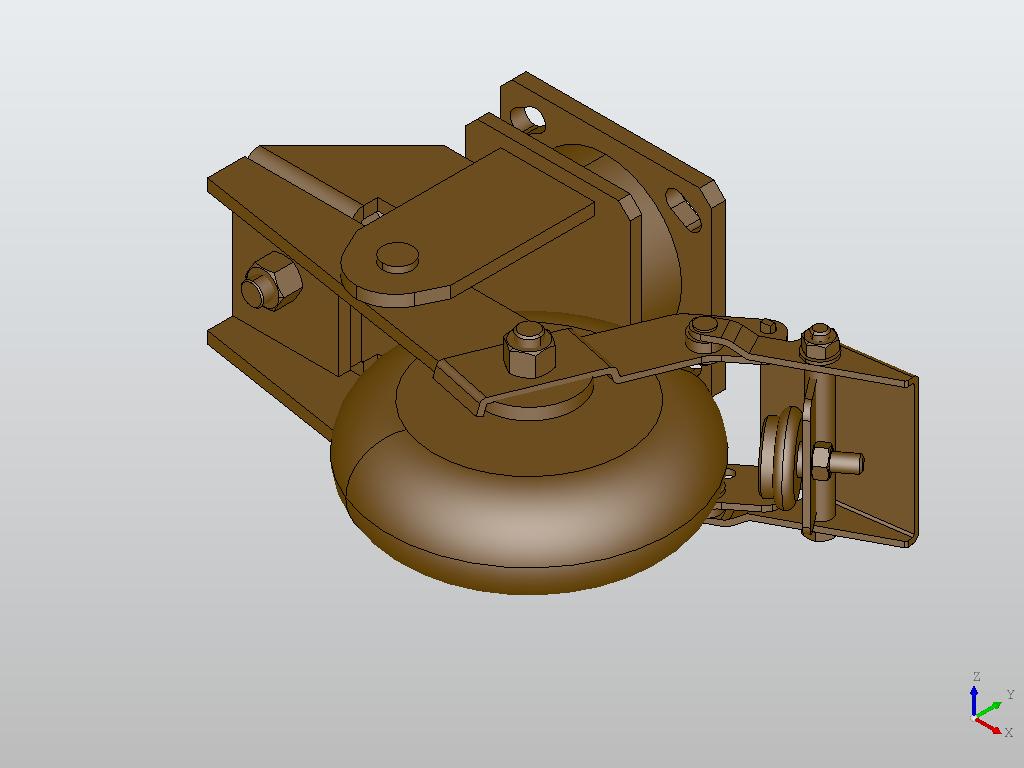}} \\
\hline
\end{tabular}
\caption{Examples of the models contained in the collected Traceparts CAD dataset.}
\label{fig:examplesTracepart}
\end{figure}

The classes considered belong to the domain of mechanical components and are given by, respectively, \textit{screws}, \textit{nuts}, \textit{hinges}, \textit{fans}, \textit{iron bars}, and \textit{wheels}.
Examples of the six classes of the dataset are shown in Figure~\ref{fig:examplesTracepart}. 

The Traceparts library collects templates from different companies. In this way, two visually very similar models that belong to the same class but that were produced by different companies can be generated in different ways and therefore have different graph structures. For example, a sphere can be defined by a company $A$ starting from a single sphere, while a company $B$ can define it as the union of two half spheres.

\begin{figure}[ht]
\centering
\begin{tabular}{cccc}
\textbf{(a)} & \textbf{(b)} & \textbf{(c)} & \textbf{(d)}\\
\subfloat{\includegraphics[width = .2\linewidth]{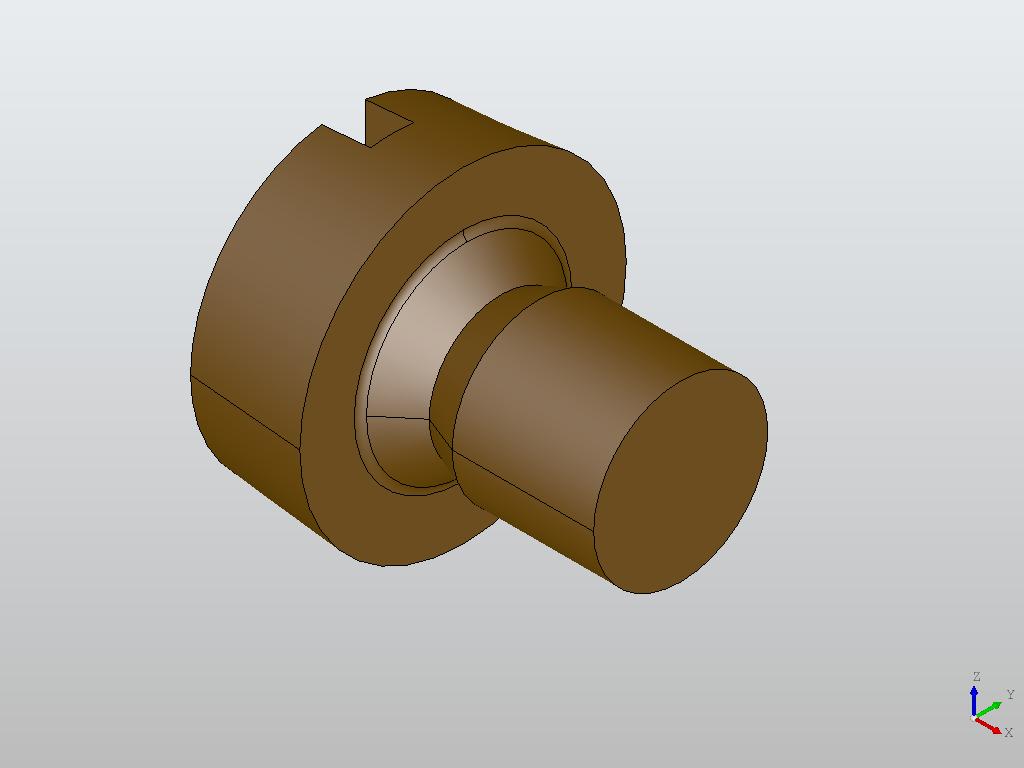}} &
\subfloat{\includegraphics[width = .2\linewidth]{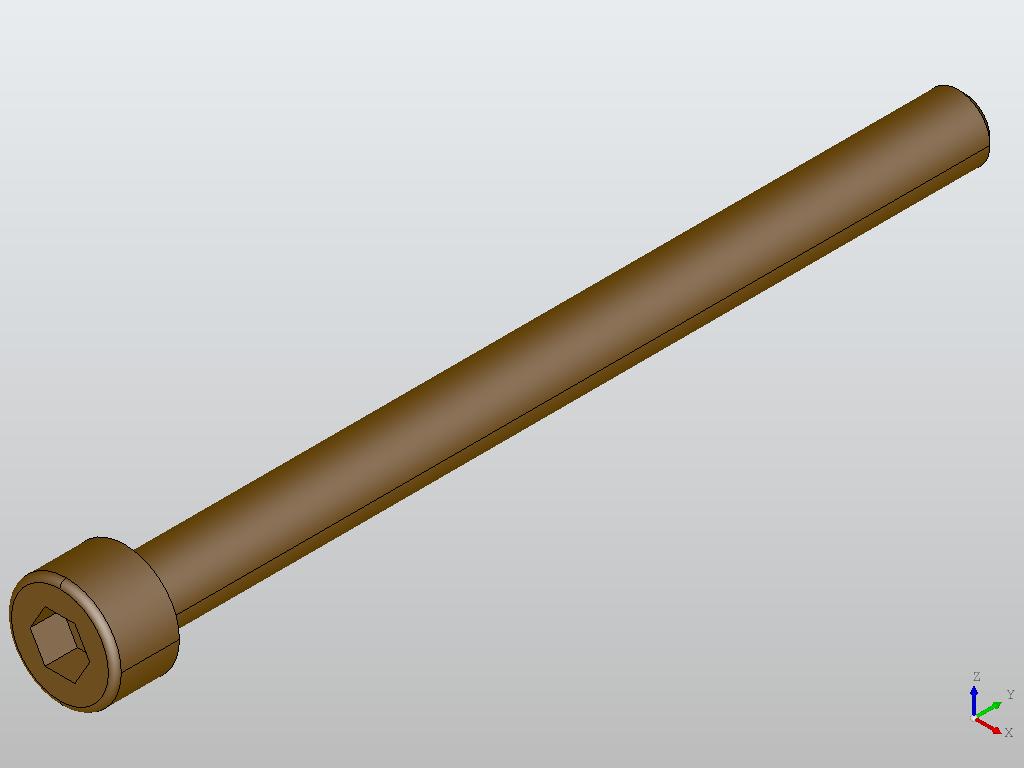}} &
\subfloat{\includegraphics[width = .2\linewidth]{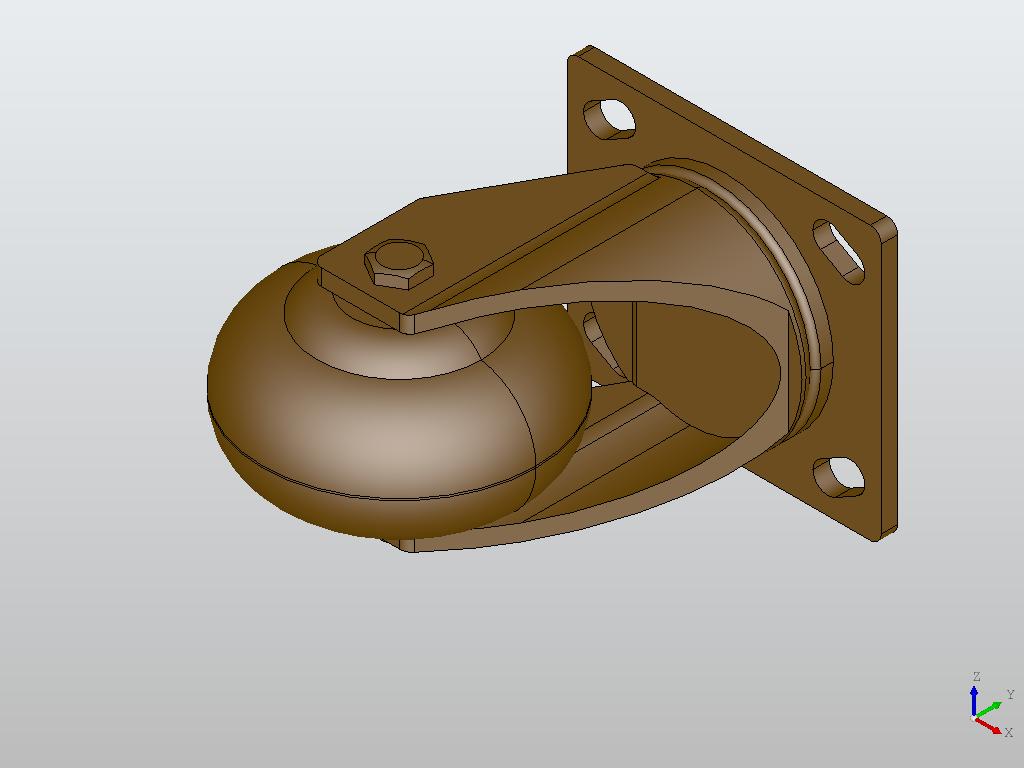}} &
\subfloat{\includegraphics[width = .2\linewidth]{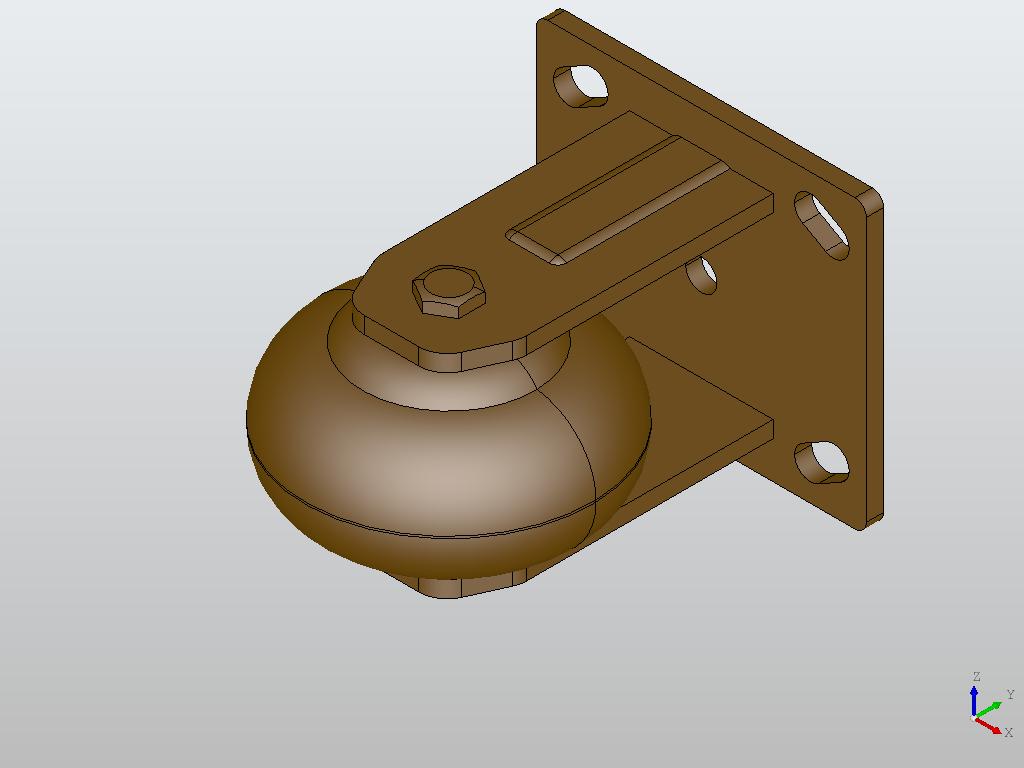}} \\
\subfloat{\includegraphics[width = .2\linewidth]{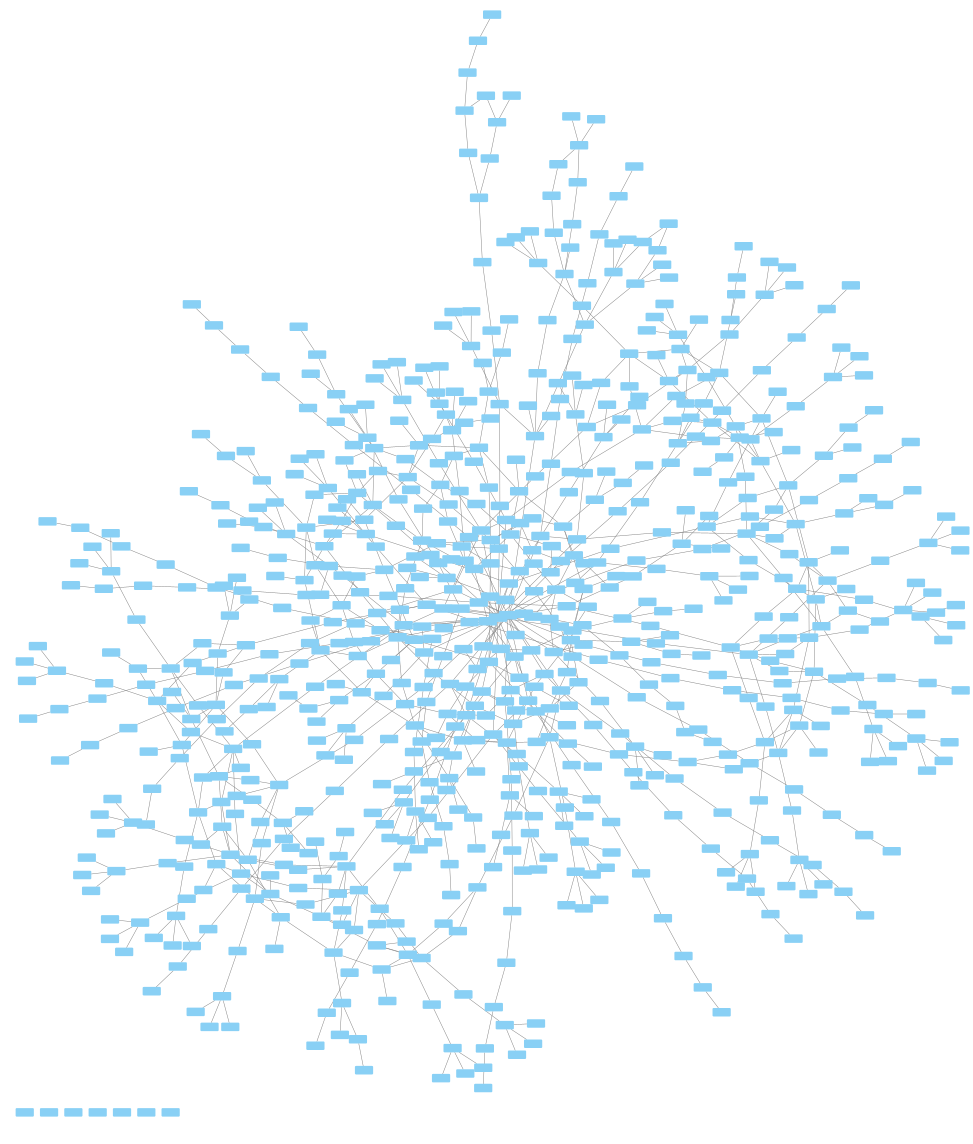}} &
\subfloat{\includegraphics[width = .2\linewidth]{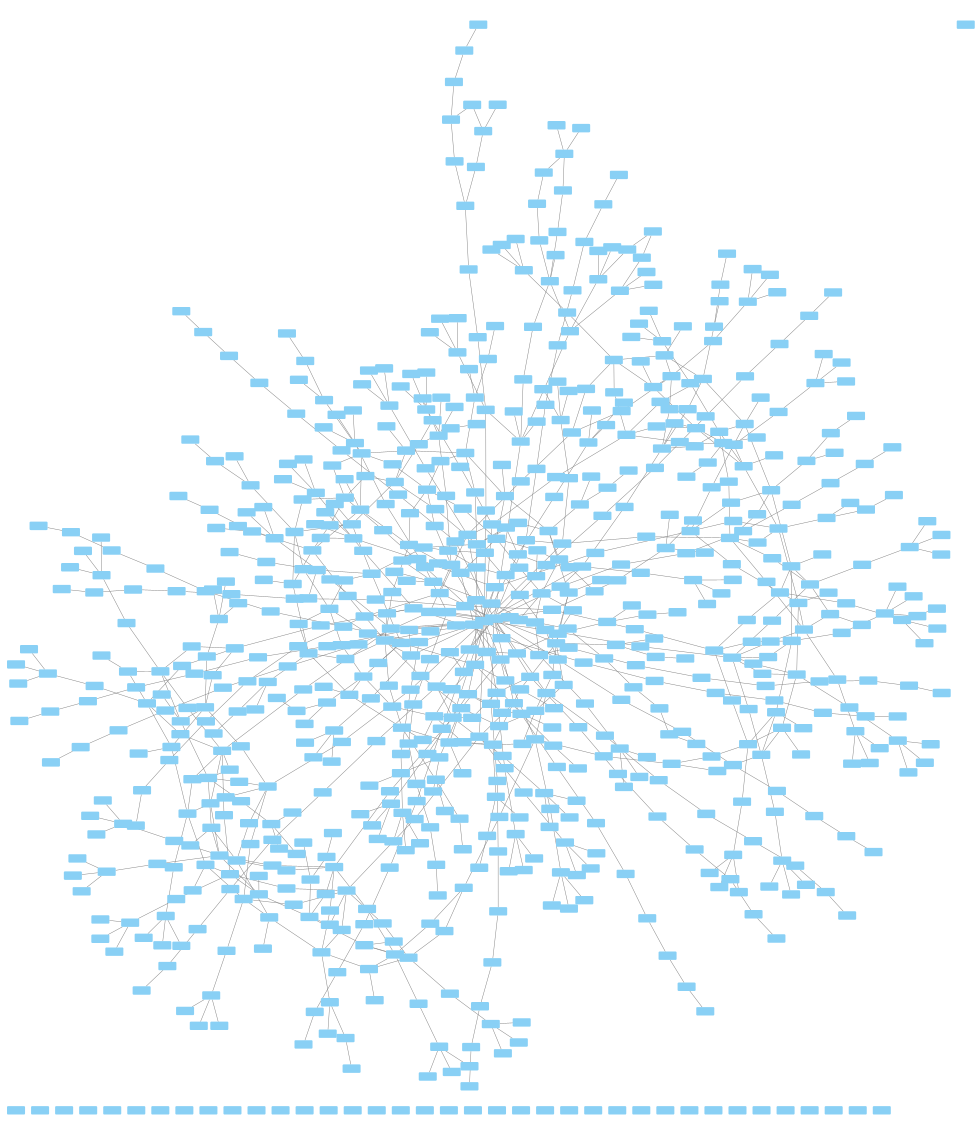}} &
\subfloat{\includegraphics[width = .2\linewidth]{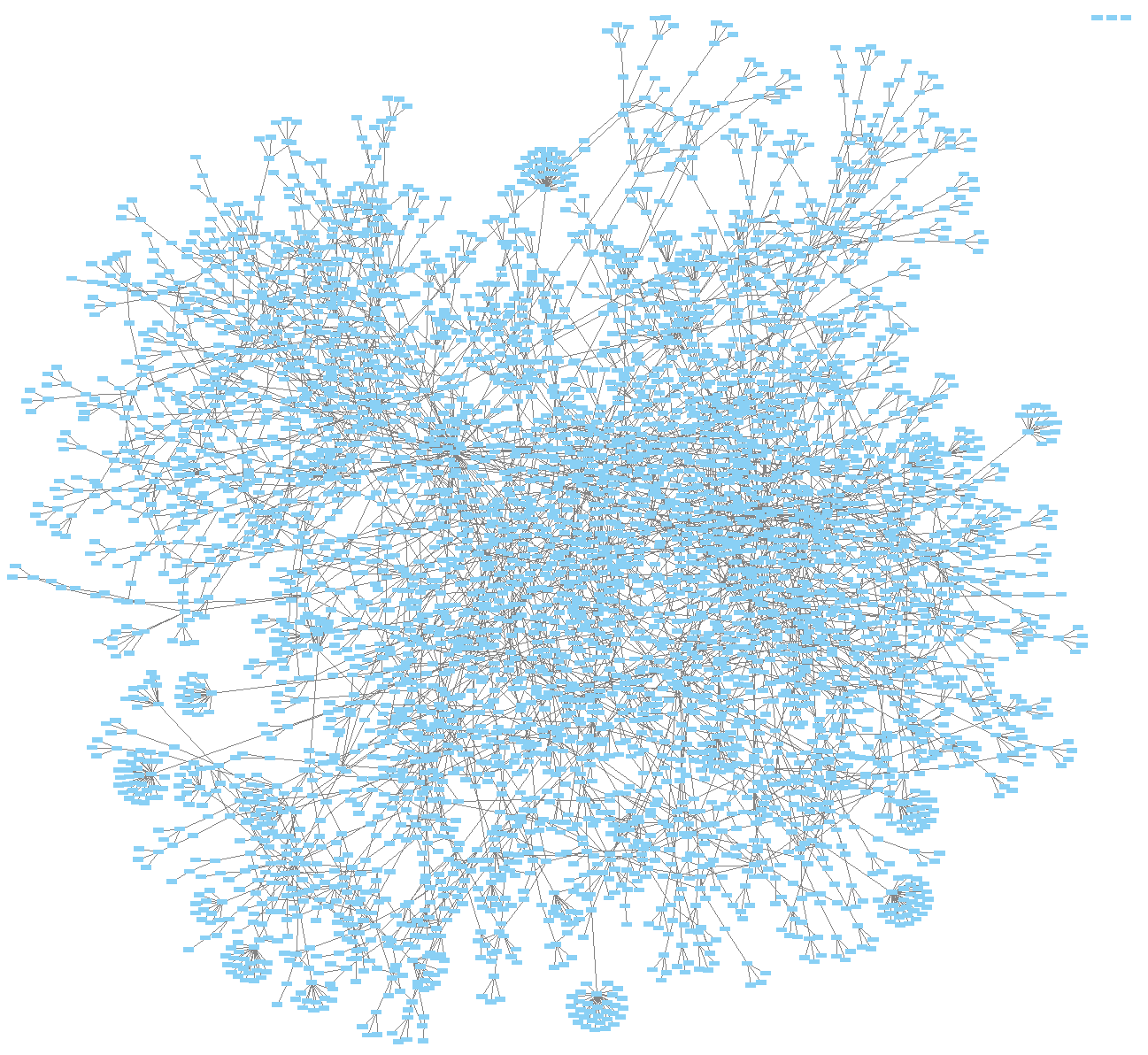}} &
\subfloat{\includegraphics[width = .2\linewidth]{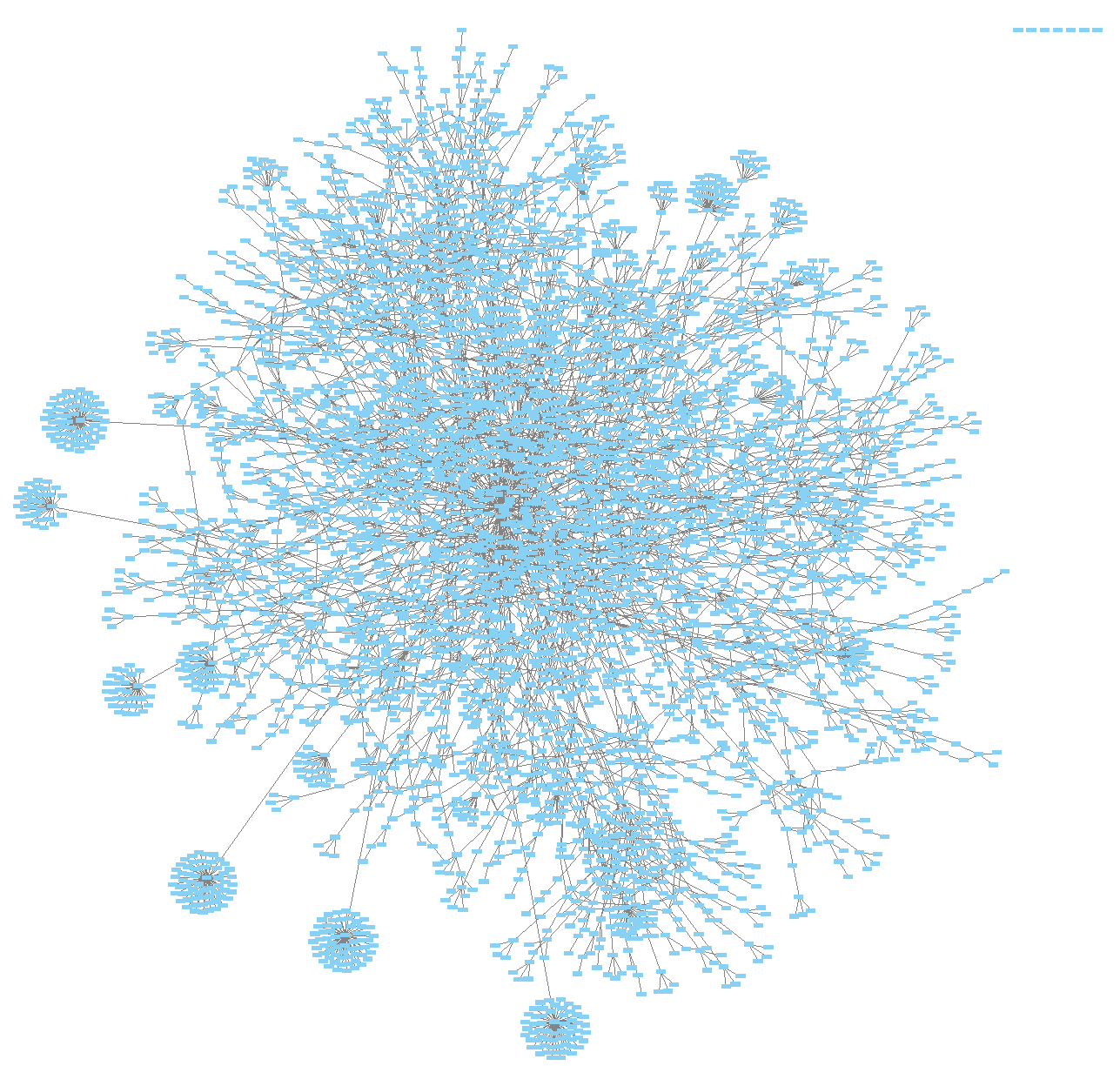}} \\
\end{tabular}
\caption{Examples of CAD models in the dataset and their respective generated graphs. 3D models belonging to the same class, \textbf{(a)}-\textbf{(b)} and \textbf{(c)}-\textbf{(d)},  can be represented by graphs with similar structure even if they are visually different such as \textbf{(a)}-\textbf{(b)} or by graphs with different structure even if they are visually similar such as \textbf{(c)}-\textbf{(d)}.}
\label{fig:graph_vs_aspect}
\end{figure}

On the other hand, two models belonging to the same class but visually different, if generated by the same company, can be represented by graphs with similar topology. In this case, the visual differences between the models are given by the different attributes of the nodes in the graphs. For example, two different spheres, one very large and one very small, will be characterized by the same graph structure but different radius values in the node that defines the sphere. In Figure~\ref{fig:graph_vs_aspect}, we show some examples of visual similarity compared to graph similarity in the case of models made by the same company and different companies.

Table~\ref{tab:Tracepart} shows the characteristics of the graph nodes in each class. The average number of nodes varies greatly depending on the class: screws, nuts, and iron bars have hundreds of nodes, whereas hinges, fans, and wheels have thousands. Some classes are therefore represented by a greater number of nodes, corresponding to more complex graphs. While a screw or iron bar can be represented with a relatively small graph, a wheel needs 10 times the number of nodes to be represented with a graph derived from the original STEP file.

The variance in the number of nodes also varies considerably across classes. In particular, the classes of screws and iron bars are characterized by zero variance, despite the fact that models within them are visually different. This is due to the fact that all models in those classes are made by a single company and are therefore characterized by a single style of model creation, which corresponds to a very similar graph structure. 

\begin{table}[ht]
\centering
    \begin{tabular}{ l r  r } 
        \hline
        \textbf{Class} & \textbf{Mean} \# \textbf{nodes} & \textbf{Variance} \# \textbf{nodes} \\ 
        \hline
        screws &  773.0  & 0.0\\ 
        nuts   &  981.4  & 403428.7\\ 
        hinges &  3282.6 & 2987419.7\\ 
        fans   &  3782.2 & 18587.3\\ 
        iron bars & 130.0 & 0.0\\
        wheels &  9307.2 & 8803938.1\\
        \hline
        \end{tabular}
    \caption{Properties of the models in the classes of the proposed Traceparts CAD dataset.}
    \label{tab:Tracepart}
\end{table}

\begin{figure}[ht]
\centering
\setlength{\tabcolsep}{4.5pt} 
\begin{tabular}{cccccccc}
\hline
\textbf{C 0} & \textbf{C 1} & \textbf{C 2} & \textbf{C 3} & \textbf{C 4} & \textbf{C 5} & \textbf{C 6} & \textbf{C 7}\\
\hline \\
\subfloat{\includegraphics[width = .09\columnwidth]{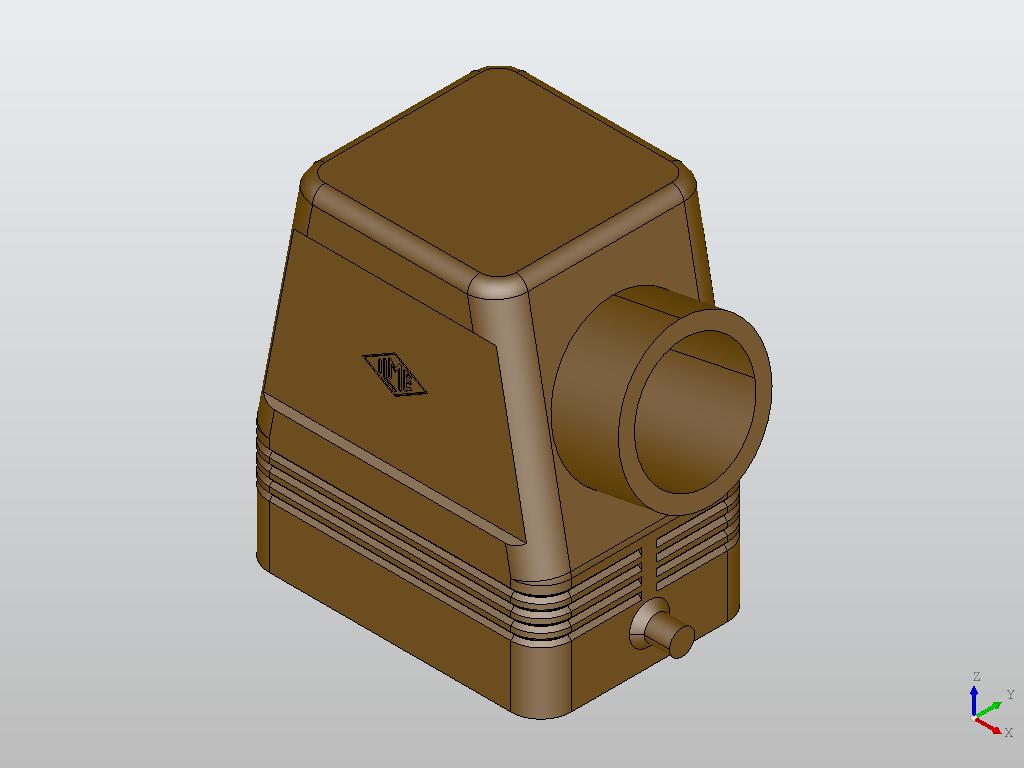}} &
\subfloat{\includegraphics[width = .09\linewidth]{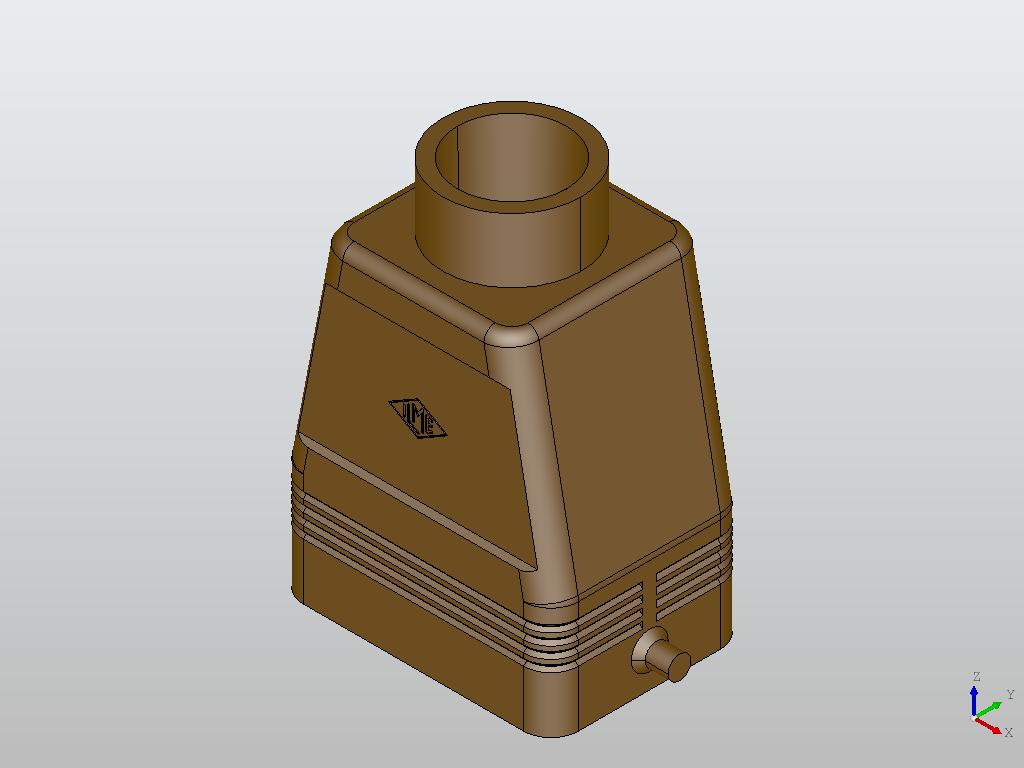}} &
\subfloat{\includegraphics[width = .09\linewidth]{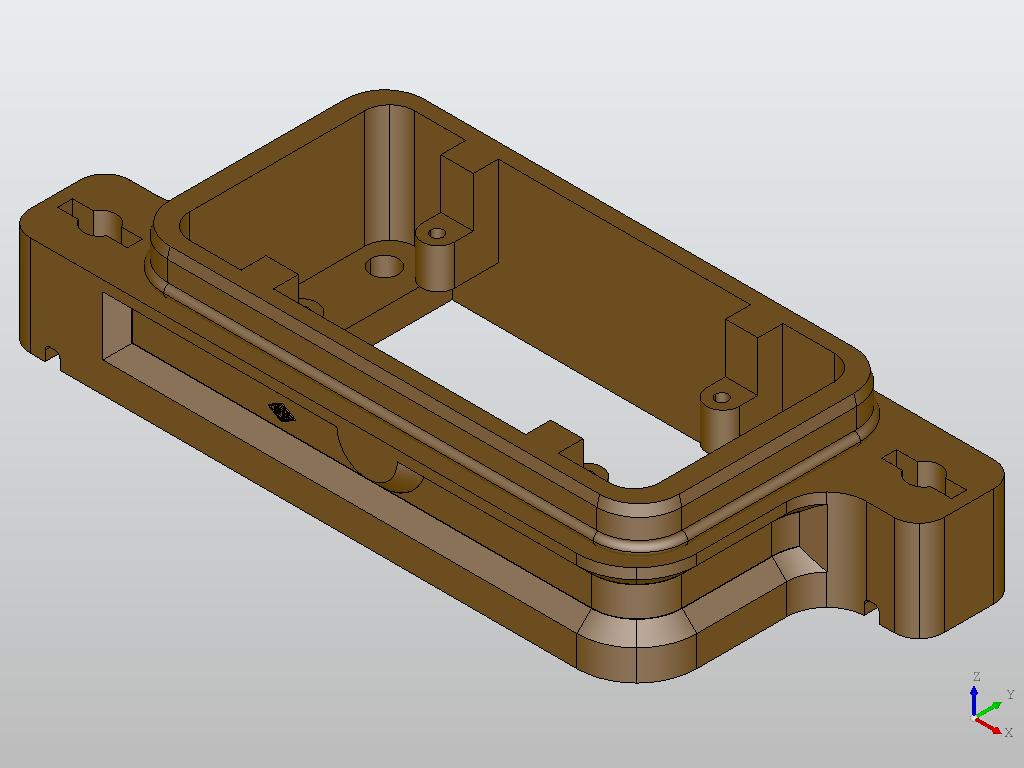}} &
\subfloat{\includegraphics[width = .09\linewidth]{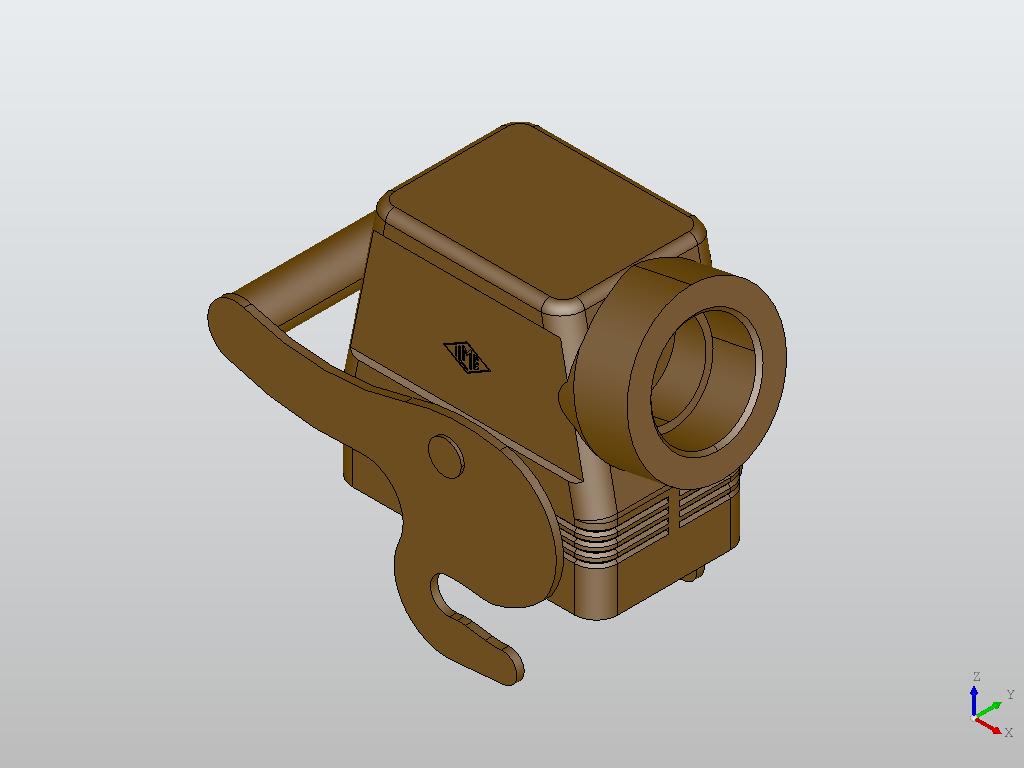}} &
\subfloat{\includegraphics[width = .09\linewidth]{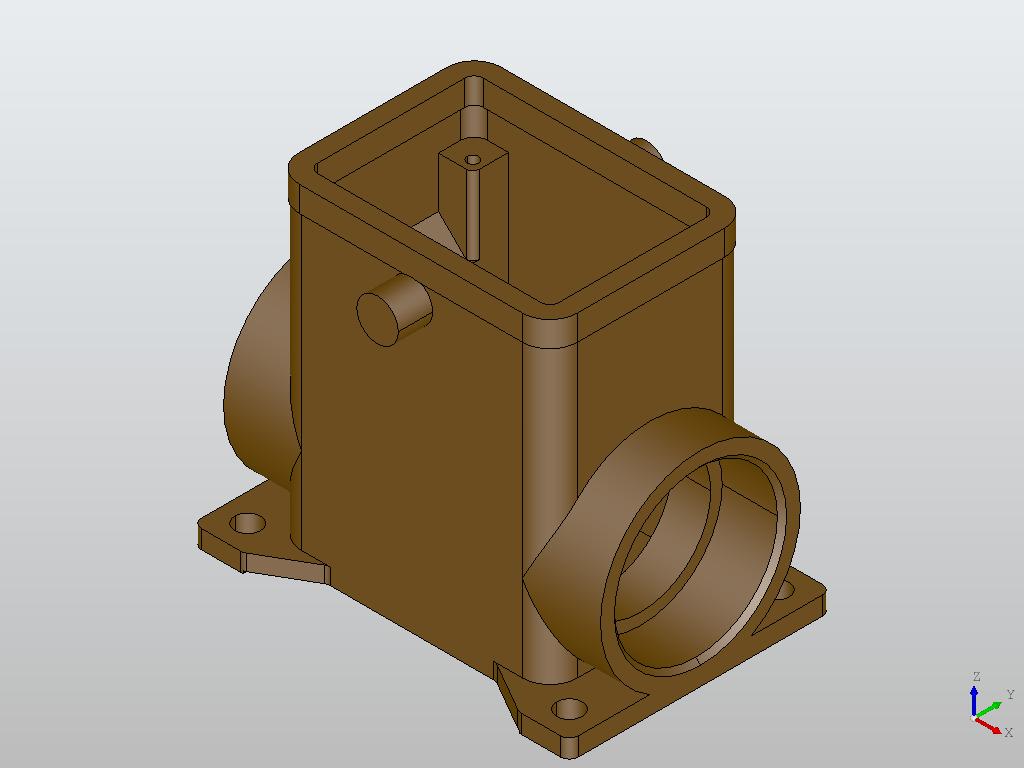}} &
\subfloat{\includegraphics[width = .09\linewidth]{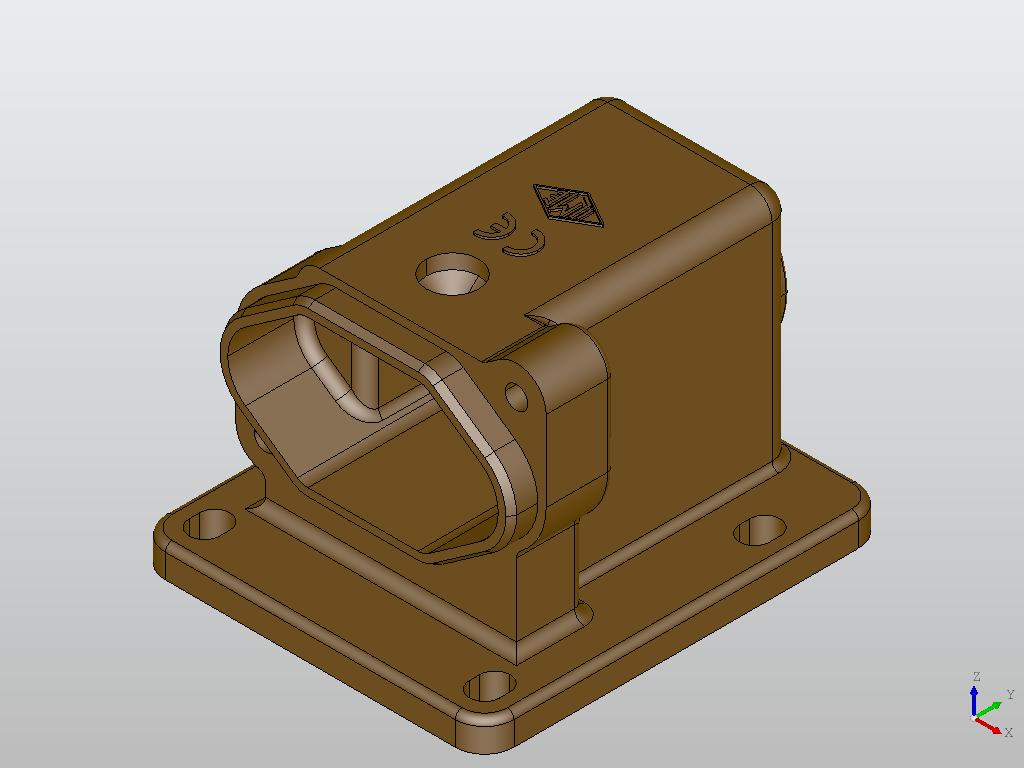}} &
\subfloat{\includegraphics[width = .09\linewidth]{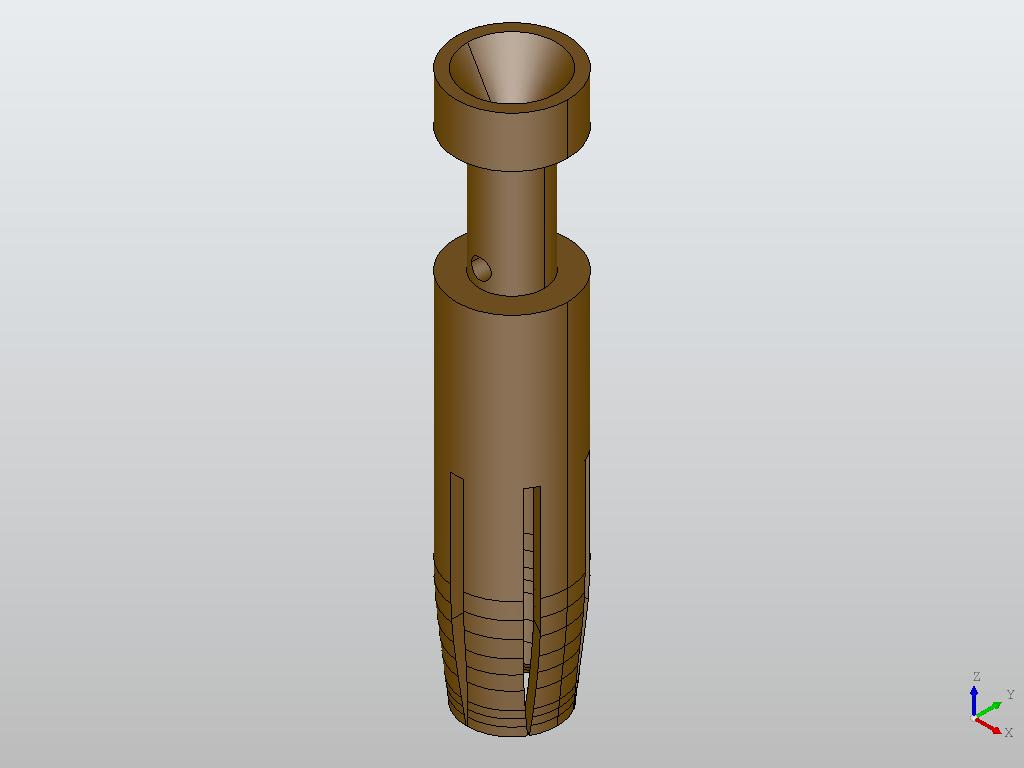}} &
\subfloat{\includegraphics[width = .09\linewidth]{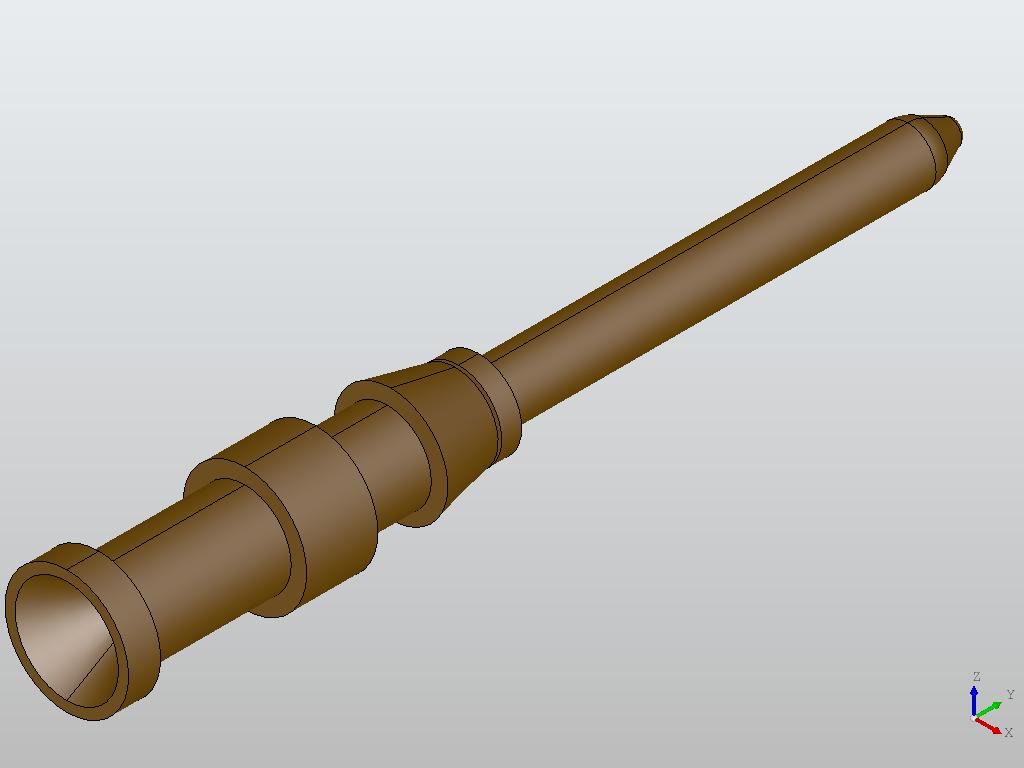}} \\
\subfloat{\includegraphics[width = .09\linewidth]{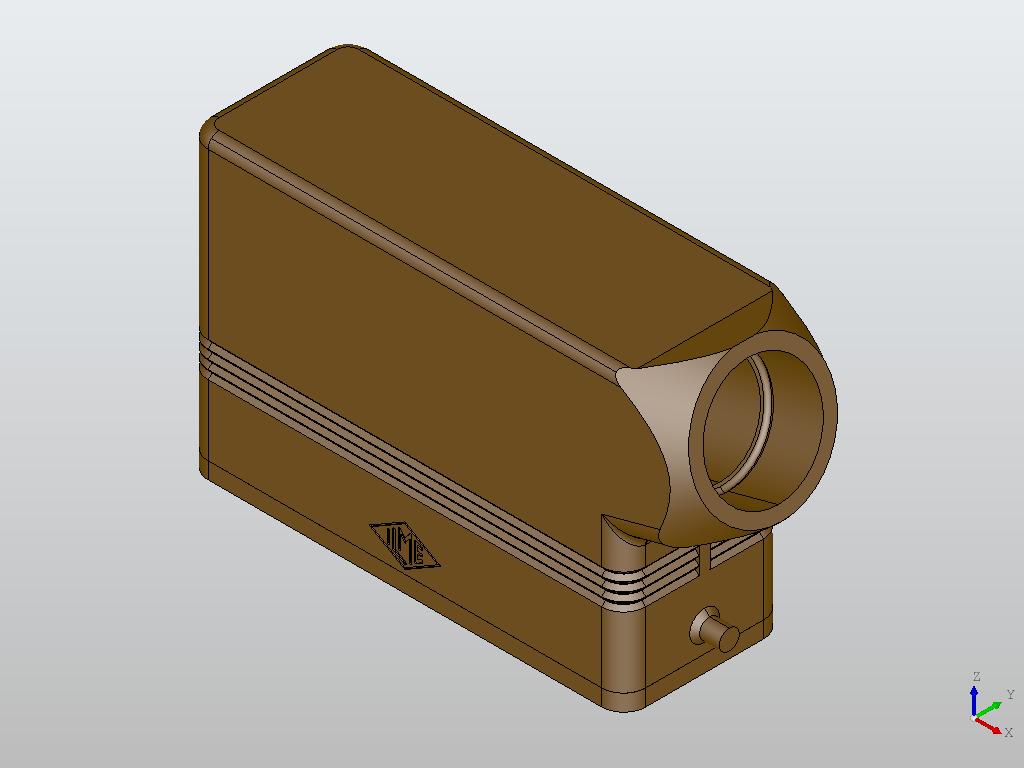}} &
\subfloat{\includegraphics[width = .09\linewidth]{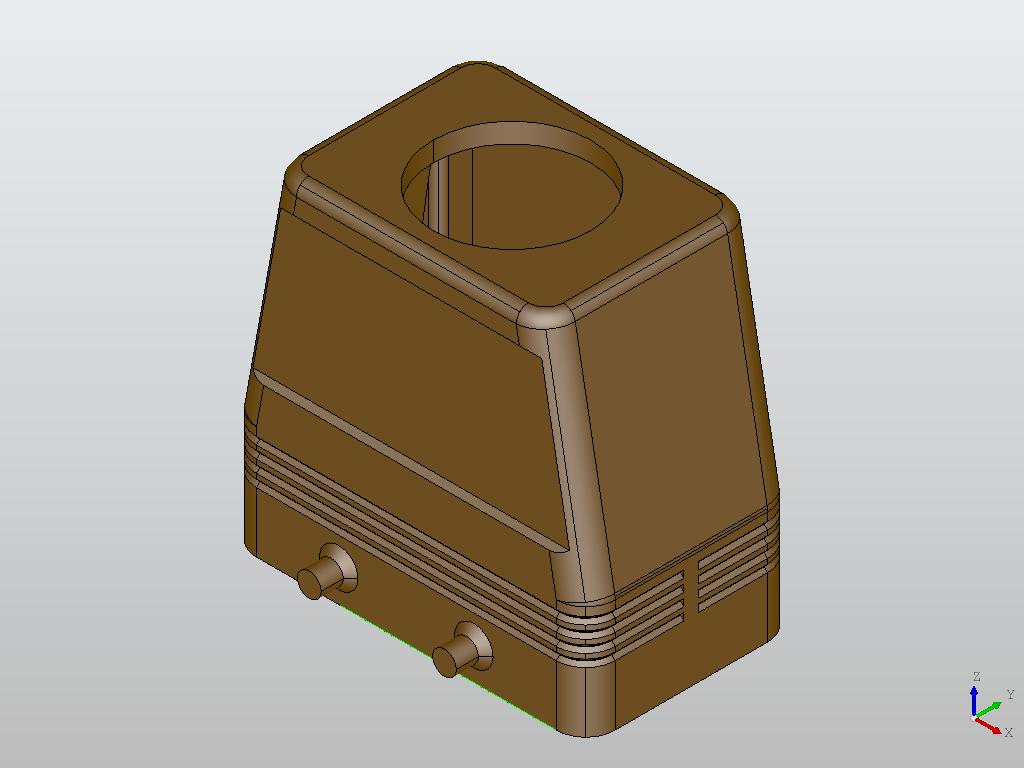}} &
\subfloat{\includegraphics[width = .09\linewidth]{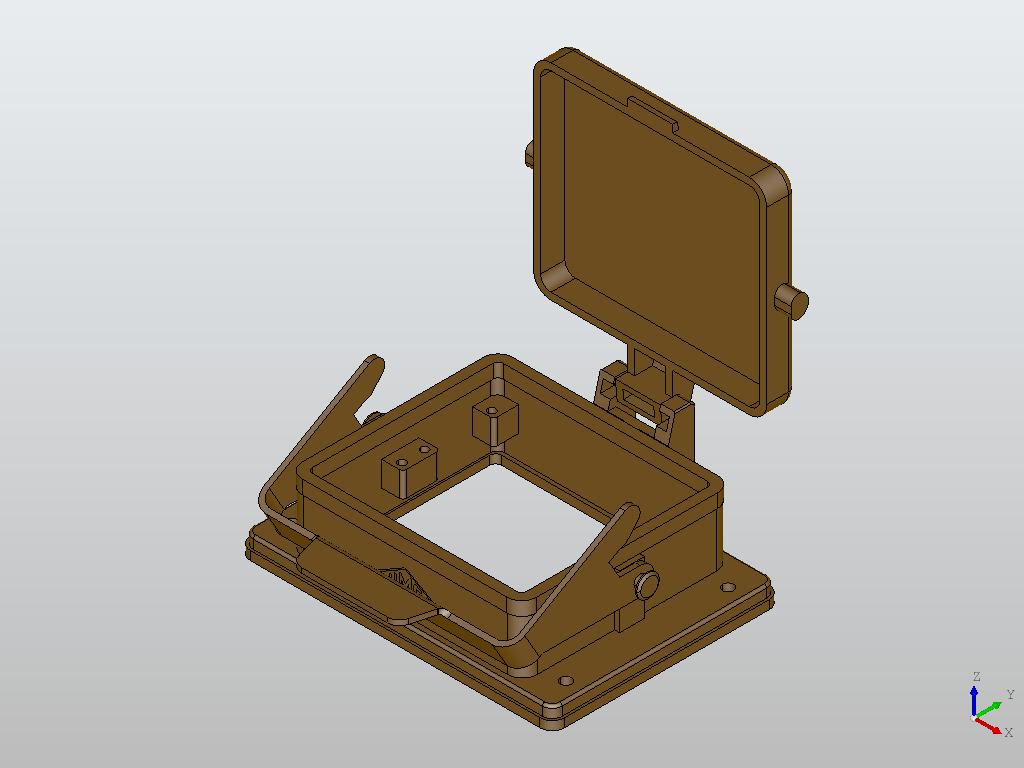}} &
\subfloat{\includegraphics[width = .09\linewidth]{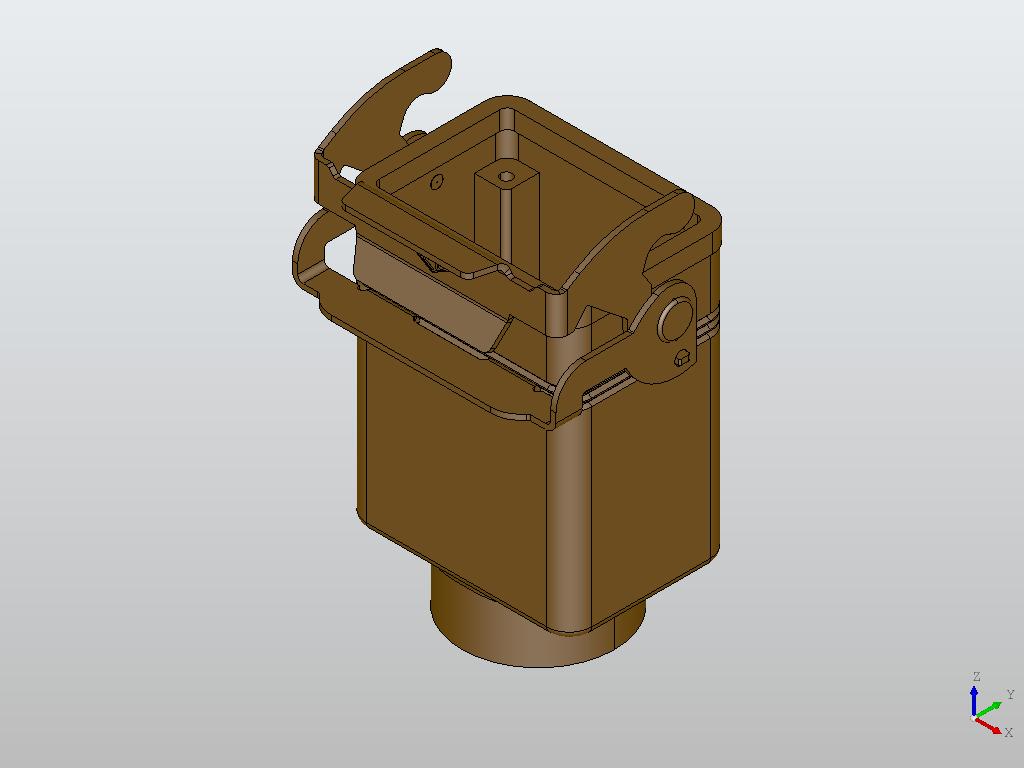}} &
\subfloat{\includegraphics[width = .09\linewidth]{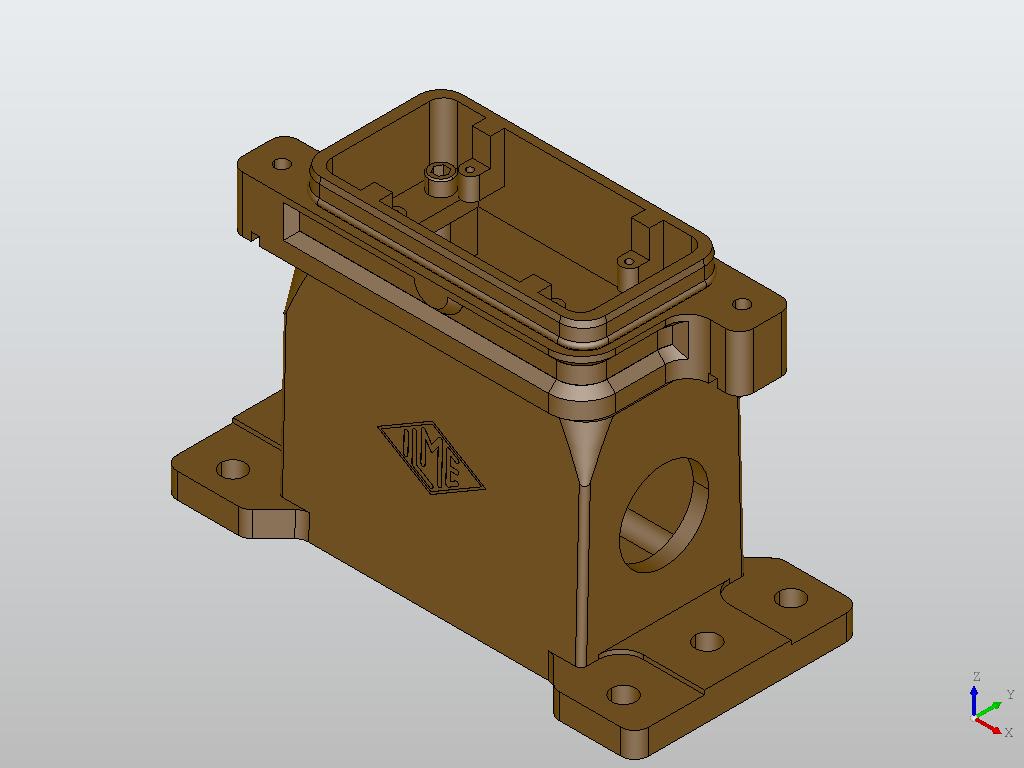}} &
\subfloat{\includegraphics[width = .09\linewidth]{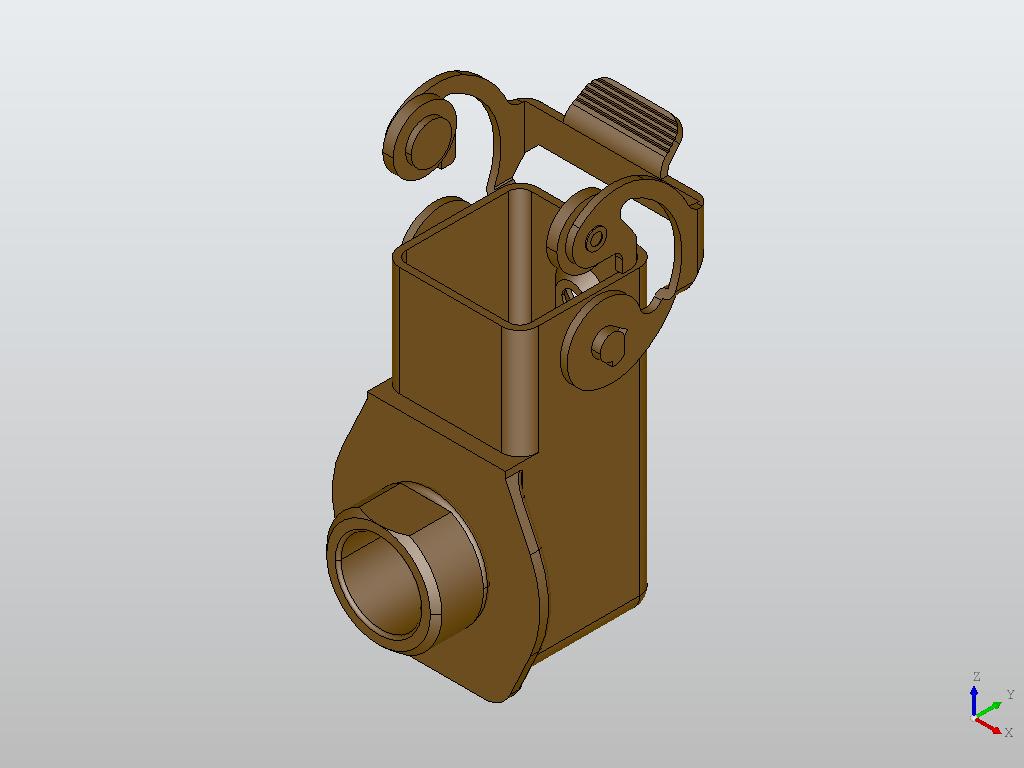}} &
\subfloat{\includegraphics[width = .09\linewidth]{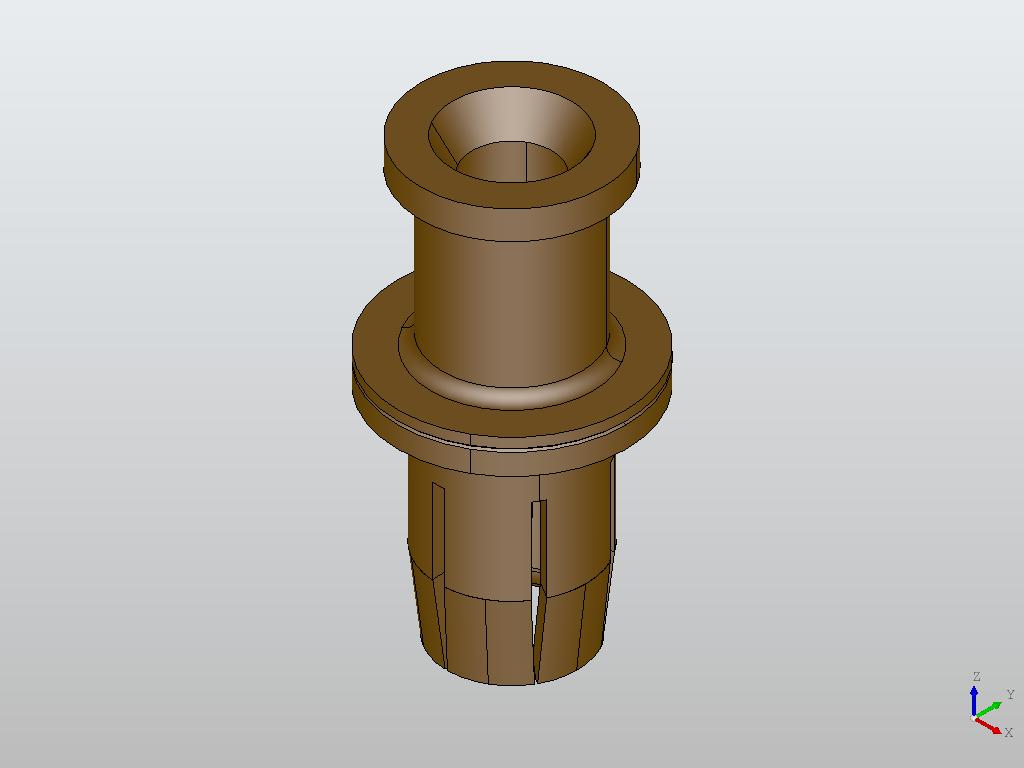}} &
\subfloat{\includegraphics[width = .09\linewidth]{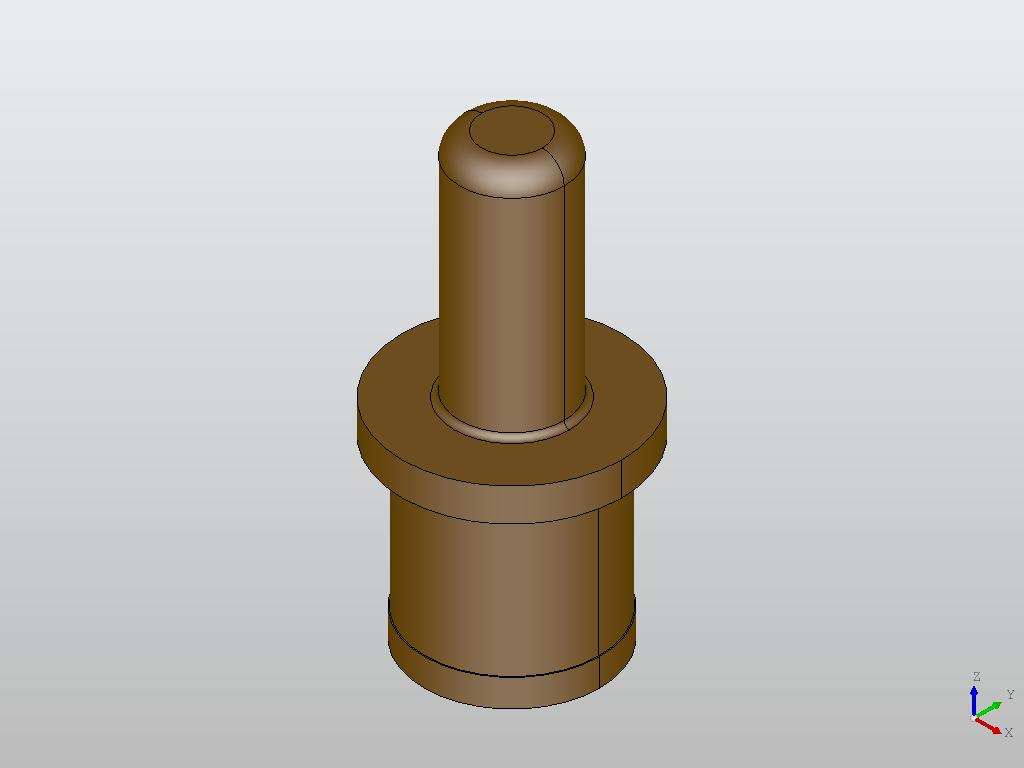}} \\
\subfloat{\includegraphics[width = .09\linewidth]{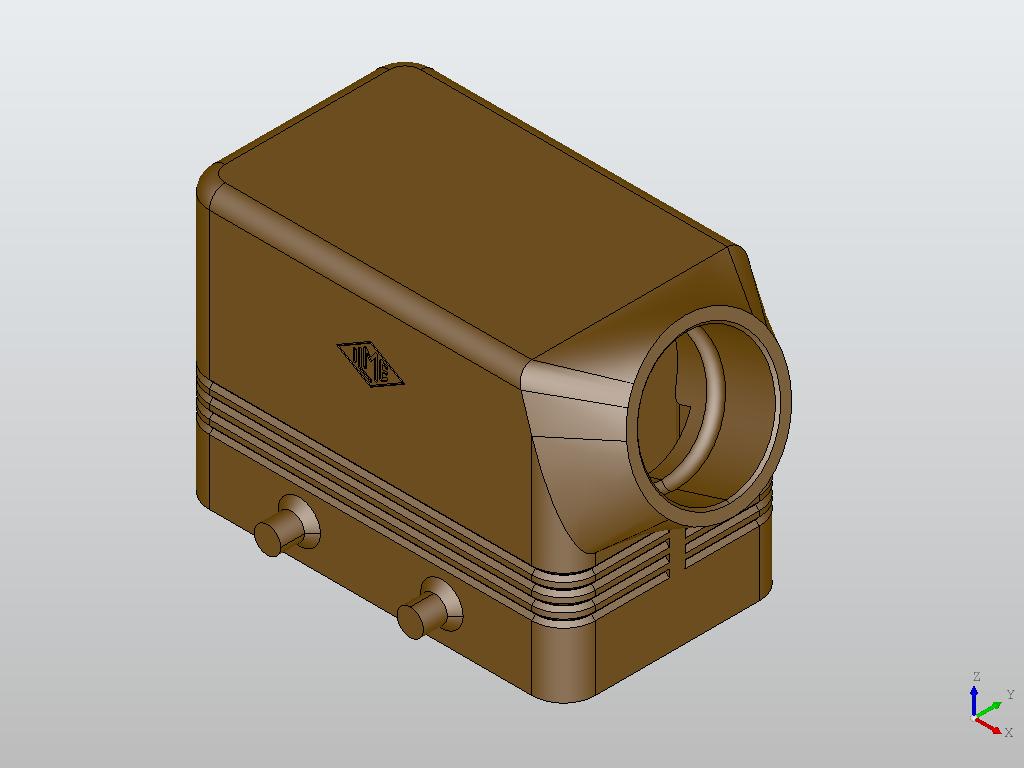}} &
\subfloat{\includegraphics[width = .09\linewidth]{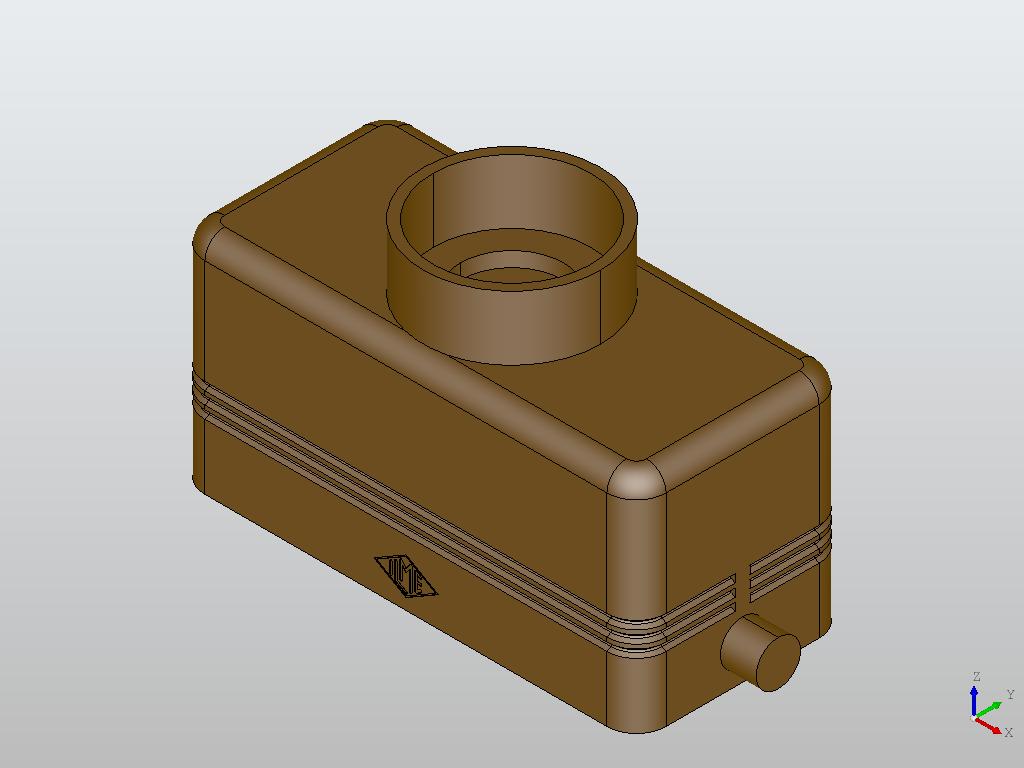}} &
\subfloat{\includegraphics[width = .09\linewidth]{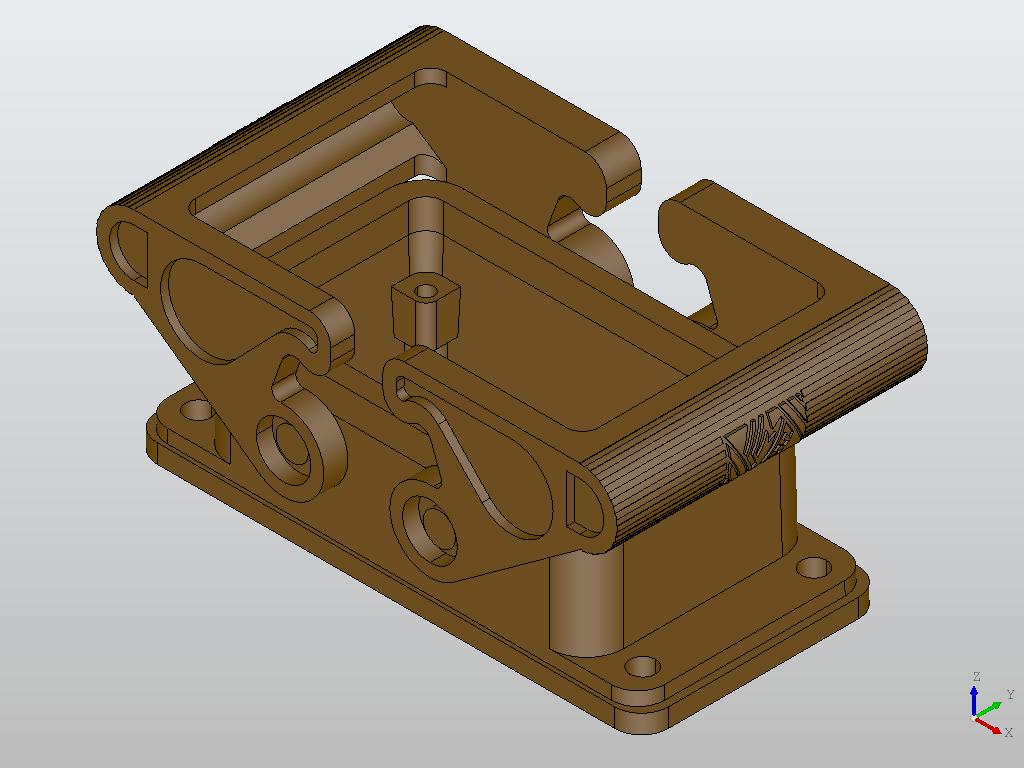}} &
\subfloat{\includegraphics[width = .09\linewidth]{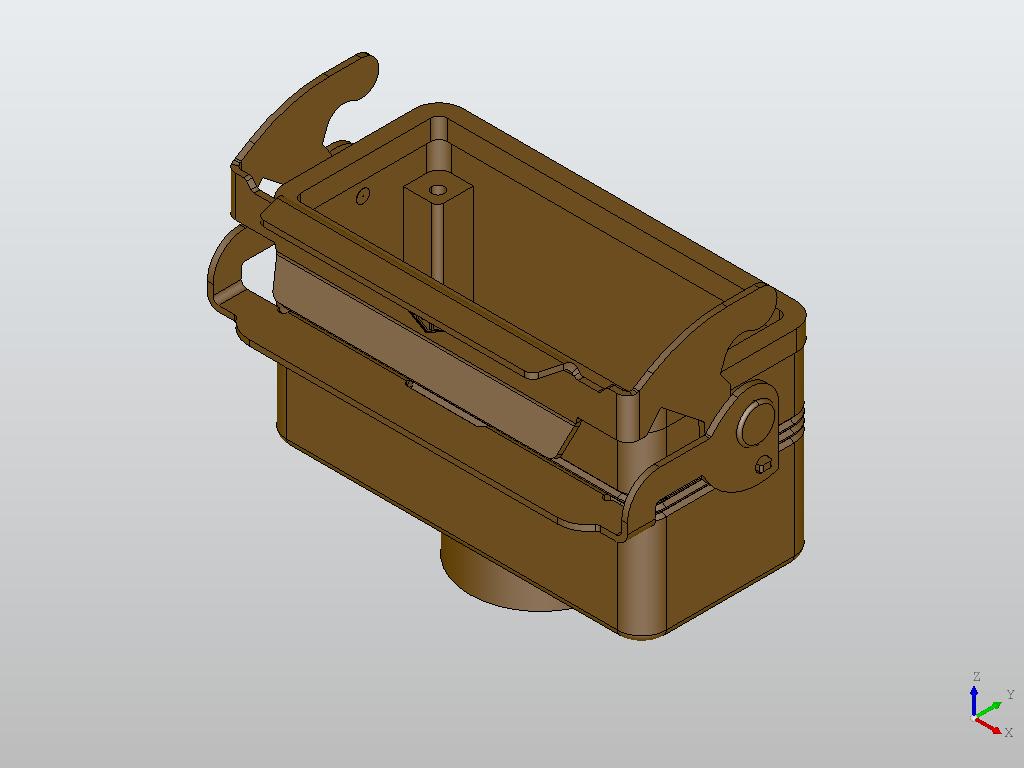}} &
\subfloat{\includegraphics[width = .09\linewidth]{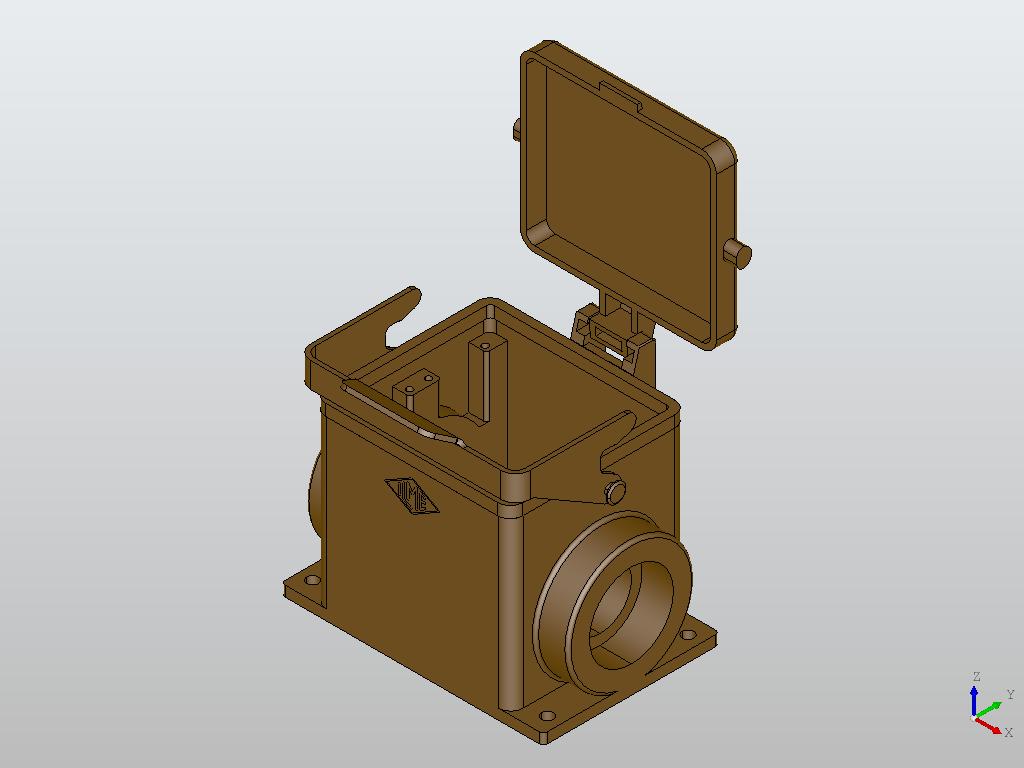}} &
\subfloat{\includegraphics[width = .09\linewidth]{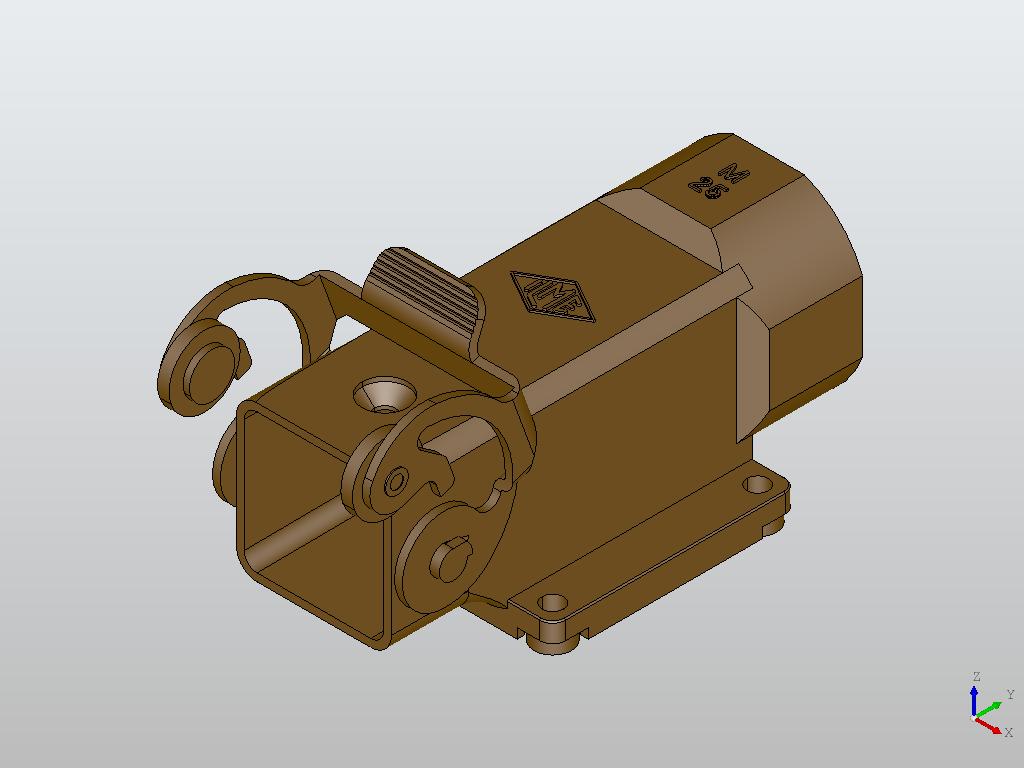}} &
\subfloat{\includegraphics[width = .09\linewidth]{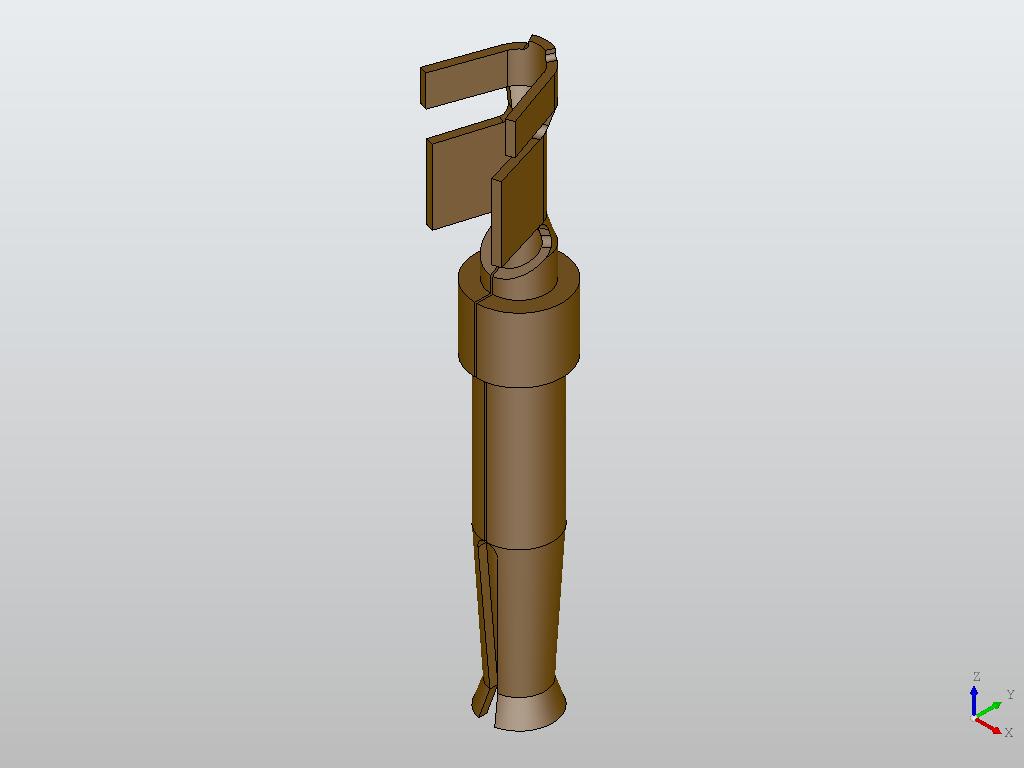}} &
\subfloat{\includegraphics[width = .09\linewidth]{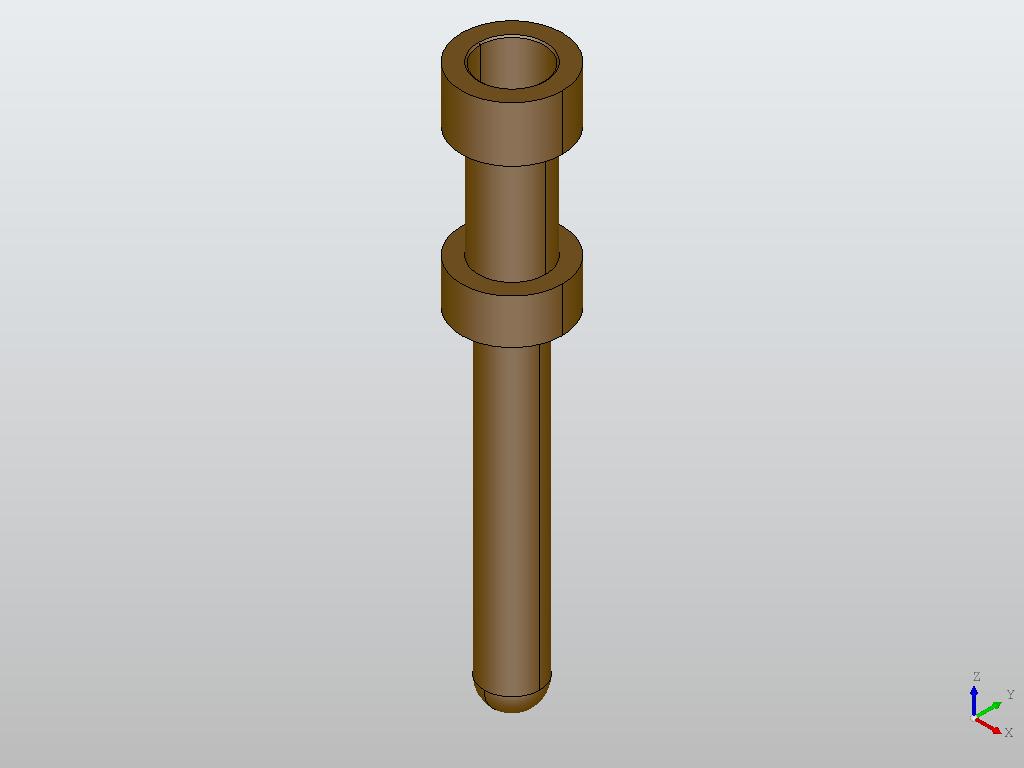}} \\
\hline
\end{tabular}
\caption{Examples of the models contained in the collected Configurators CAD dataset.}
\label{tab:ConfiguratoriExamples}
\end{figure}

\textbf{Configurators CAD dataset}: The second dataset was created from the models made available thanks to the software modeling company Configurators~\cite{Configuratori}. 
It is made up of 8 classes of 50 elements each, belonging to the STEP 214 AP. Examples of models for each class are shown in Figure~\ref{tab:ConfiguratoriExamples}. In Table~\ref{tab:nodes_Configuratori}, we show the characteristics of the graphs for each class. 

\begin{table}[ht]
\centering
    \begin{tabular}{ l r r } 
        \hline
        \textbf{Class} & \textbf{Mean} \# \textbf{nodes} & \textbf{Variance} \# \textbf{nodes} \\ 
        \hline
        0 &  11244.0 & 3112635.5\\ 
        1 &  10300.4  & 1799380.8\\ 
        2 &  13027.9 & 15810147.7\\ 
        3 &  18241.6 & 12651546.6\\ 
        4 &  14707.9 & 14708304.2\\
        5 &  15748.2 & 23589269.2\\
        6 &  6784.2 & 10737704.7\\
        7 &  958.9 & 34371.9\\
        \hline
        \end{tabular}
    \caption{Properties of models for the classes of the Configurators CAD dataset.}
    \label{tab:nodes_Configuratori}
\end{table}

This dataset contains models taken from a real-world working application in which each class represents a different component of a single construct. On average, these 3D CAD models are more complex than those included in the TraceParts CAD dataset and the classes are more similar to each other. For example, classes 0 and 1 vary only by the position, vertical or lateral, of a hole on the structure. This allows us to test our proposed method in a more challenging and real world application scenario that can also stress more the differences with other existing solutions in the literature. 

\subsection{GCN-based CAD model classification}\label{sec:classification}
After transforming the STEP models into graph data, the information contained in the nodes is summarized into vectors. 
The total number of different types of STEP entities contained in the datasets models is 80, therefore each node is represented by a vector of dimension 80 obtained through the one-hot encoding of its entity type.

We implemented the GCN network described in Section~\ref{sec:GraphClassificationRetrieval} using the PyTorch and PyTorch Geometric libraries, which provide the graph convolution and the fully connected modules. The code is available at the link \url{https://github.com/divanoLetto/3D_STEP_Classification}.  We then tested the network on the two graph datasets. The network has been trained for 50 epochs for the first dataset and 100 epochs for the second one with an initial learning rate (\textit{lr}) equal to $0.0005$ and a scheduler mechanism that decays the \textit{lr} by a gamma value equal to $0.1$, when the current loss value is greater than the average of the last 6 epochs.

Each dataset was divided into a train set and a test set with a 90/10 split; furthermore, 10\% of the train set models were taken to build a validation set. 
We therefore get, for the first dataset, 486 models for the train set, 60 models for the test set and 54 models for the validation set.  
For the second dataset we obtained 324 models for the train set, 40 models for the test set, and 36 models for the validation set. 
We set the GCN with 3 convolutional layers with dimensions of 64, 32 and 32, followed by the attention mechanism and two fully connected layers with size 32 and 6 or 8, equal to the number of classes for the first and second datasets, respectively.

This configuration was decided after a set of tests performed on the first dataset as shown in Table~\ref{tab:GCN bottleneck}; in these test, the network performance was compared while varying the size of the first FC layer (\textit{i.e.}, the bottleneck layer).
By increasing the size of the bottleneck, we were able to increase the expressive power of the network to 100\% accuracy.

\begin{table}[ht]
\centering
    \begin{tabular}{ c  r  r  r} 
        \hline
        \textbf{GCN bottleneck size} & \textbf{Accuracy} & \textbf{Loss} & \textbf{Precision} \\ 
        \hline
        8 & 91.66\% & 0.84 & 0.93 \\ 
        16 & 100.0\% & 0.55 & 1.00 \\ 
        \textbf{32} & \textbf{100.0}\% & \textbf{0.51} & \textbf{1.00} \\ 
        \hline
    \end{tabular}
    \caption{Study on the effect of varying the size of the GCN bottleneck layer when testing on the first dataset.}
    \label{tab:GCN bottleneck}
\end{table}

Regarding the attention mechanism, we studied how it helps to obtain a general graph representation from its nodes embedding. We compared our network with an attention mechanism with two other networks that merge nodes information through an unweighted average of node values and a weighted sum, where the weight associated with a node is determined by its degree. Results reported in Table~\ref{tab:attention} indicate that the attention mechanism allows us to significantly increase the accuracy of the model compared to other non-learning-based methods. 

\begin{table}[ht]
    \centering
    \begin{tabular}{ l r r r r } 
        \hline
        \textbf{Network module} & \textbf{Accuracy} & \textbf{Loss} & \textbf{Precision} & \textbf{Recall} \\ 
        \hline
        Unweighted average & 93.33\%  & 0.68 & 0.942 &  0.933 \\ 
        Degree weighted sum & 93.33\%  & 0.64 & 0.952 & 0.933 \\ 
        \textbf{Attention} & \textbf{100.0}\% & \textbf{0.51} & \textbf{1.000} & \textbf{1.000} \\ 
        \hline
    \end{tabular}
    \caption{Study on the attention mechanism on the first dataset to obtain an overall representation of the graph starting from the embedding of its nodes compared to: the use of a unweighted average; a weighted sum based on the degree of the nodes.}
    \label{tab:attention}
\end{table}

We compared our method with other existing approaches that operate in the domain of 3D but using different formats for the data. We considered the following methods:
\begin{itemize}
    \item \textbf{PointNet++}: we sampled from each of our CAD models a point cloud in order to train the PointNet++ network~\cite{PointNet++} that represents a standard in 3D model classification. As the number of sampling points, we considered 1024, 2500 and 10,000 points, but we did not find any significant variation. The network was trained for 50 epochs with a learning rate equal to $0.001$.
    \item \textbf{MVCNN}: 12 2D views for each model were generated through a mobile camera that rotates around the model. Each image is resized to $224\times224$, passed through the convolutional layers and then aggregated at a viewpooling layer with the representations of the other images of the same model. The Multi View CNN (MVCNN)~\cite{MVCNN} network was trained for 30 epochs with a learning rate equal to $5e^{-5}$. The same setting was used for the Single View CNN (SVCNN) variant that considers each image as a separate input data.
\end{itemize} 

\begin{figure}[ht]
\centering
\begin{tabular}{cccc}
\textbf{CAD} & \textbf{Graph} & \textbf{PointCloud} & \textbf{Multi-views}\\
\subfloat{\includegraphics[width = .2\columnwidth]{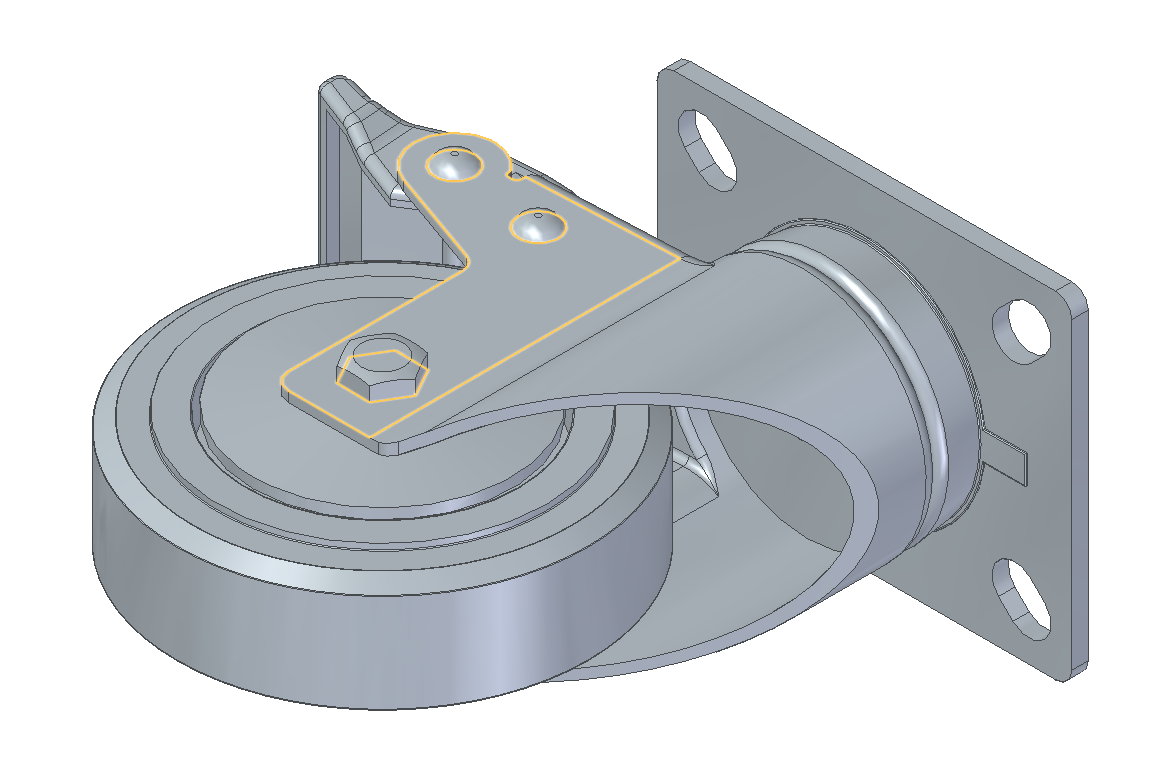}} &
\subfloat{\includegraphics[width = .2\columnwidth]{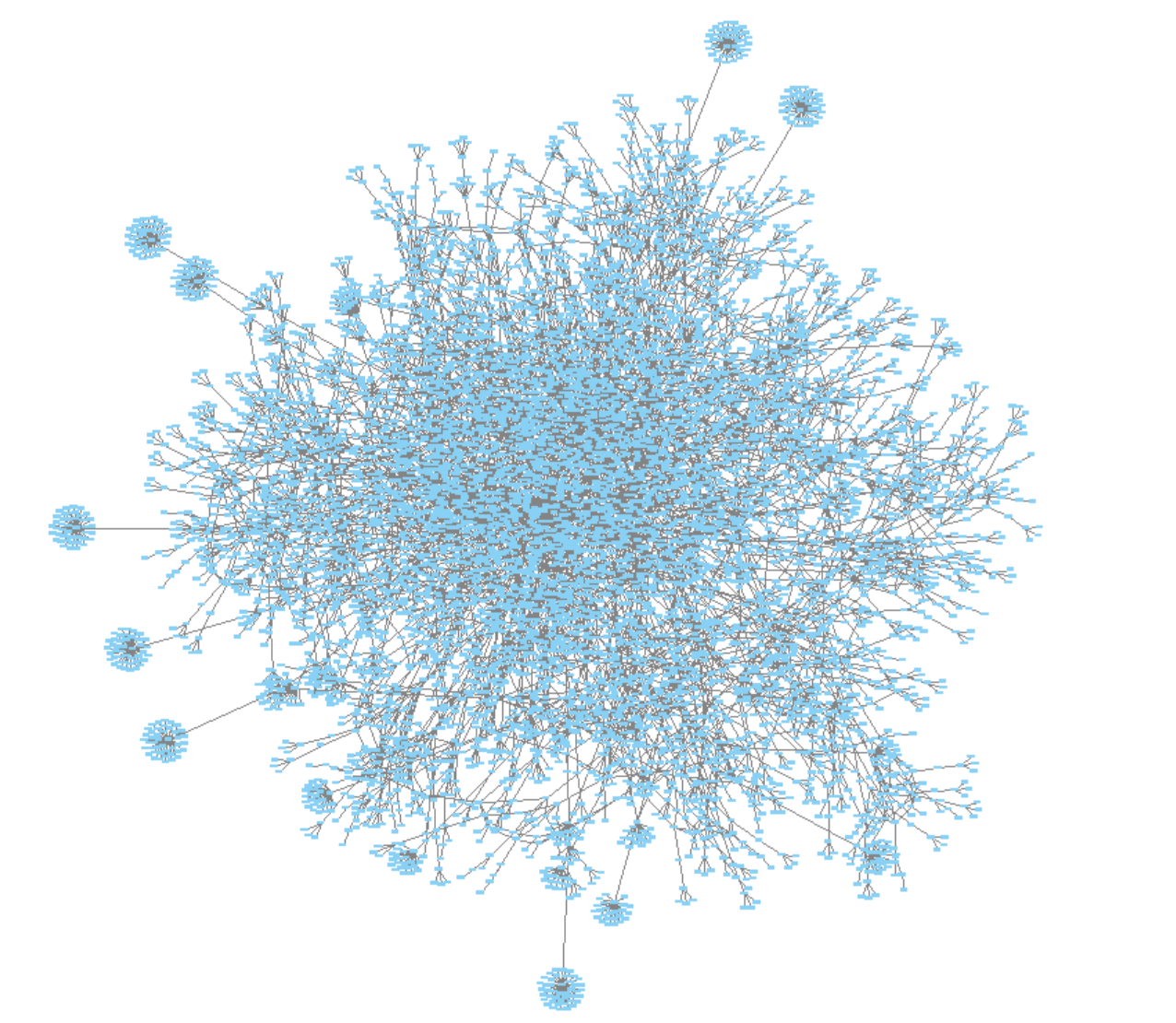}} &
\subfloat{\includegraphics[width = .2\columnwidth]{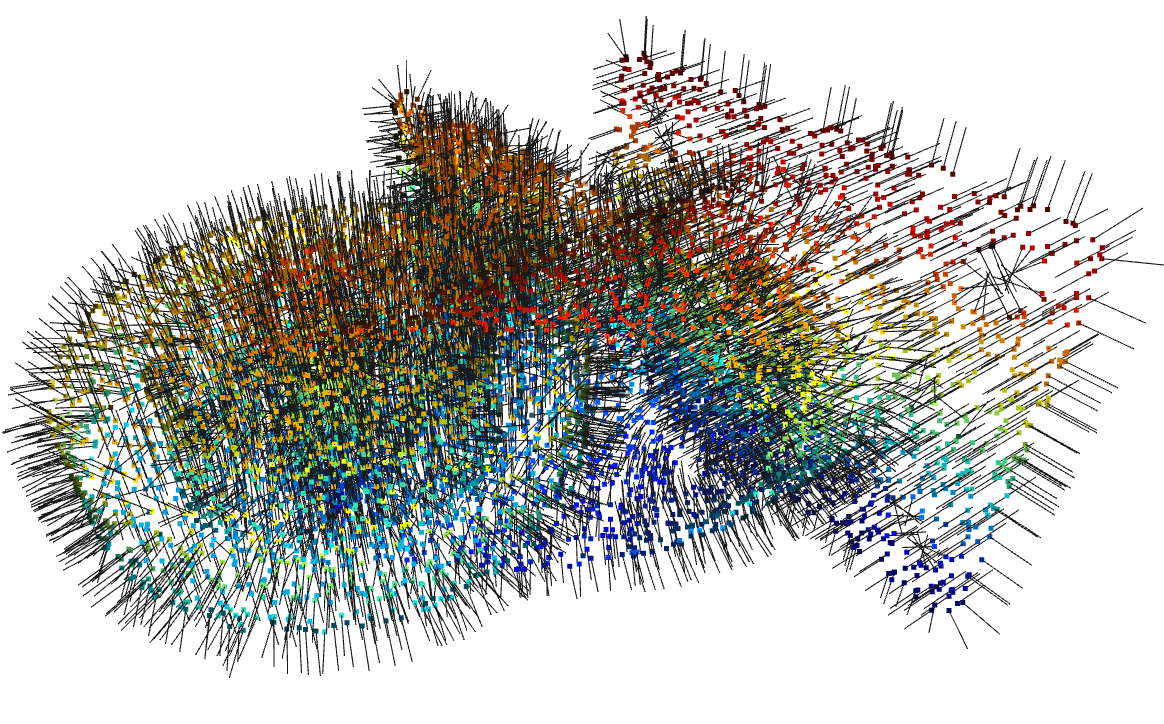}} &
\subfloat{\includegraphics[width = .2\columnwidth]{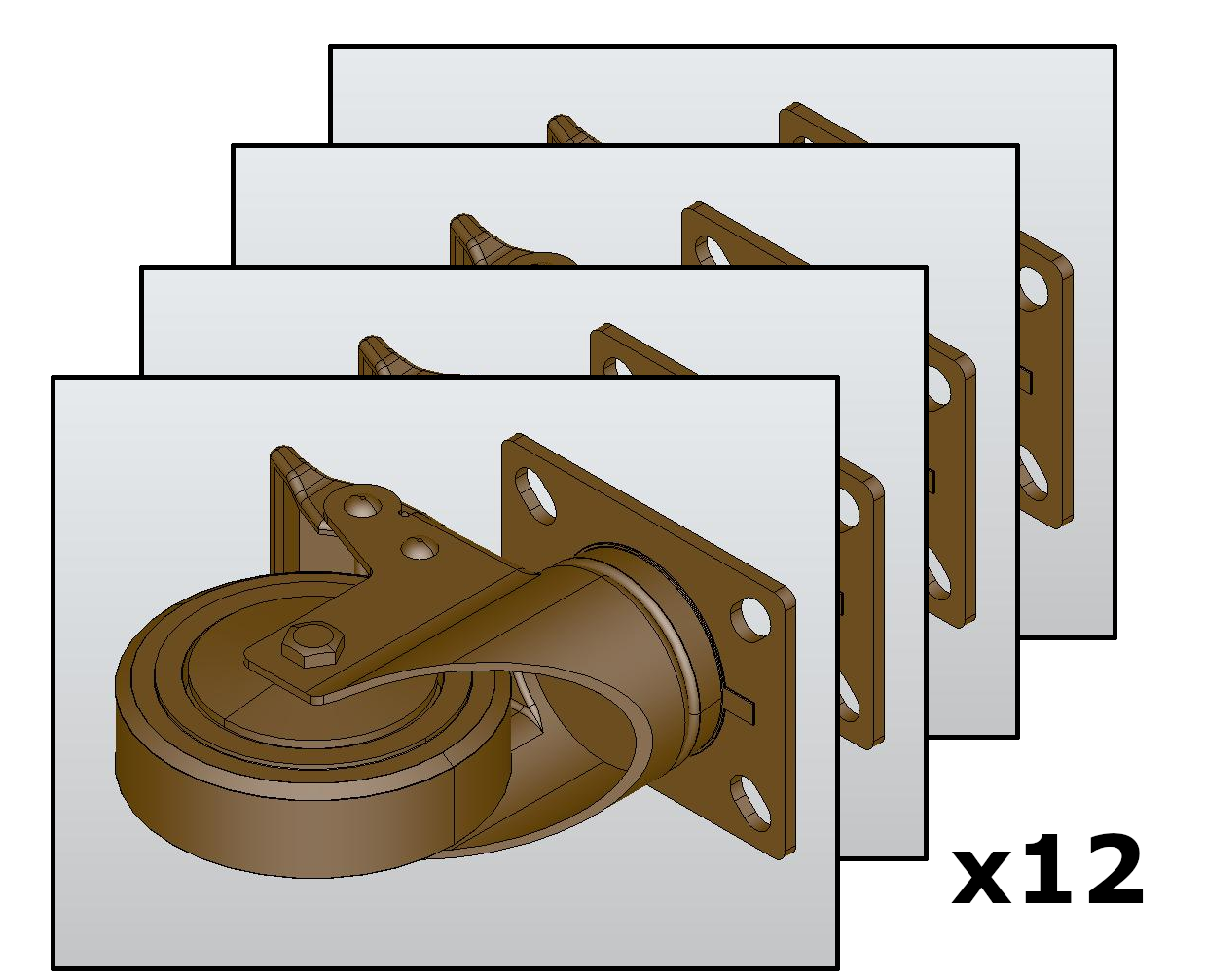}} 
\end{tabular}
\caption{Summary of the different input types of the tested approaches for a single CAD model.}
\label{fig:inputs}
\end{figure}

Figure~\ref{fig:inputs} highlights the different types of inputs for the different approaches.

Results of the comparison of those methods with our approach for the classification task are shown in Table~\ref{tab:pointNetcompare} for the Traceparts dataset and in Table~\ref{tab:Configuratori_results} for the Configurators dataset. 
Regarding the first dataset, our GCN achieved an accuracy score of 100\% outperforming the PointNet and PointNet++ networks trained on the same data, slightly beating the Single-View CNN (SVCNN) variant, and achieving the same perfect accuracy score as the MVCNN method. 

\begin{table}[ht]
\centering
    \begin{tabular}{l r} 
        \hline
        Method & Accuracy \\ 
        \hline
        PointNet & 28.7\% \\ 
        PointNet$++$ & 31.6\% \\ 
        SVCNN & 99\% \\ 
        \textbf{MVCNN} & \textbf{100\%} \\ 
        \textbf{GCN (ours)} & \textbf{100\%} \\
        \hline
    \end{tabular}
    \caption{Comparison on the Tracepart dataset of our approach based on STEP data to others state-of-the-art methods that works on different data formats.}
    \label{tab:pointNetcompare}
\end{table}

In the second dataset, our approach confirmed its validity achieving the value of 97.5\% of accuracy, therefore exceeding the results of the MVCNN method. This is due to the fact that, unlike the 2D view method, the graph-based approach is able to exploit information that is not visible externally and is not affected if two models belonging to different classes are visually similar, as for example, happens for the classes 0 and 1 or for the classes 3, 4 and 5.

\begin{table}[ht]
\centering
    \begin{tabular}{ l r } 
        \hline
        Method & Accuracy \\ 
        \hline
        PointNet & 18.7\% \\ 
        PointNet$++$ & 28.5\% \\ 
        SVCNN & 91.4\% \\ 
        MVCNN & 93.3\% \\ 
        \textbf{GCN (ours)} & \textbf{97.5\%} \\
        \hline
    \end{tabular}
    \caption{Comparison on the Configurators dataset of our approach based on STEP data to others state-of-the-art methods that works on different data formats.}
    \label{tab:Configuratori_results}
\end{table}

\begin{figure}[ht]
    \centering
    \subfloat{{\includegraphics[width=0.48\columnwidth]{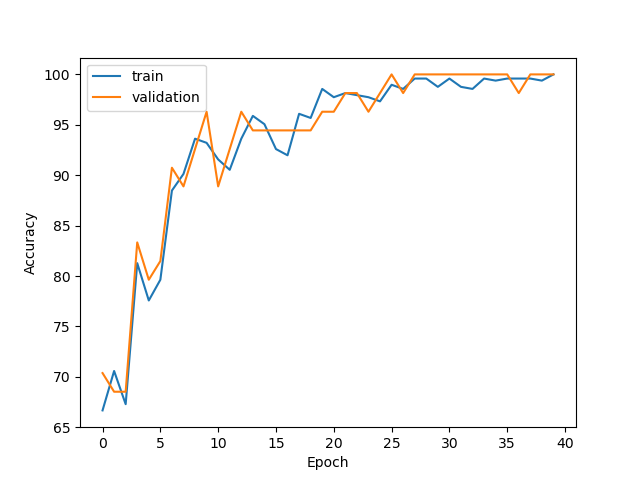} }}
    \hfill
    \subfloat{{\includegraphics[width=0.48\columnwidth]{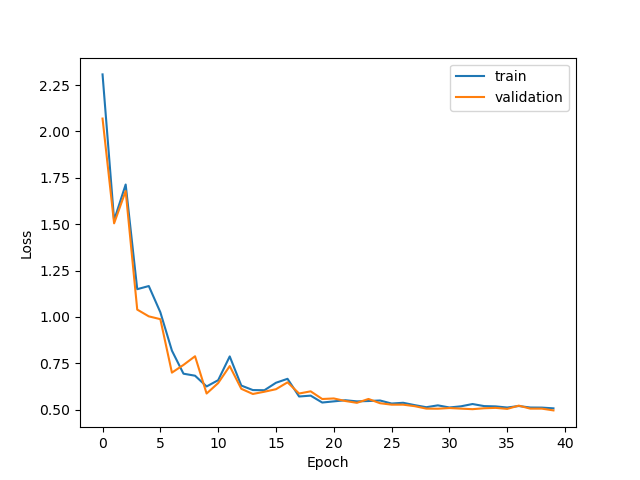} }}
    \caption{GCN training accuracy and loss obtained on the Traceparts CAD dataset.}
    \label{fig:GCNflow}
\end{figure}

\begin{figure}[ht]
    \centering
    \subfloat{{\includegraphics[width=0.48\columnwidth]{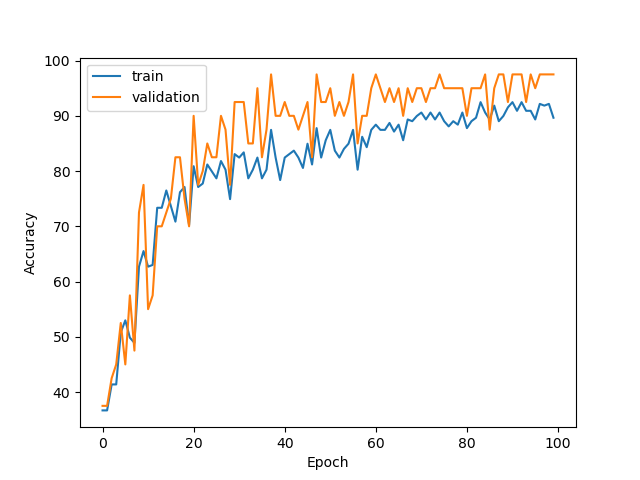} }}
    \hfill
    \subfloat{{\includegraphics[width=0.48\columnwidth]{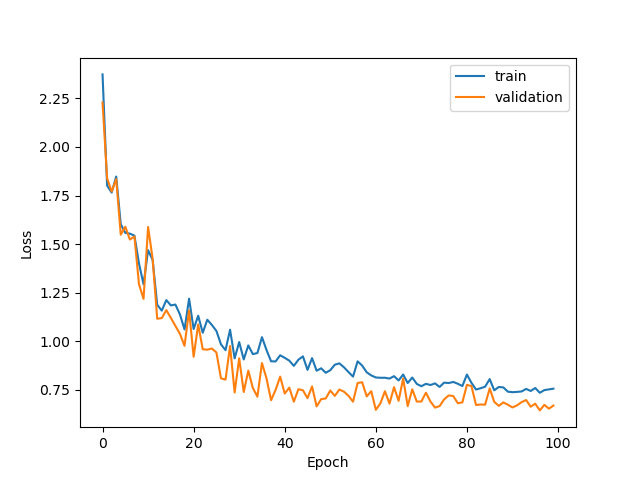} }}
    \caption{GCN training accuracy and loss obtained on the Configurators CAD dataset.}
    \label{fig:GCNflow_configuratori}
\end{figure}

The GCN training accuracy and loss for both the datasets are shown in Figures~\ref{fig:GCNflow} and~\ref{fig:GCNflow_configuratori}, respectively.

\subsection{GCN-based CAD model retrieval}\label{sec:retrieval}
The GCN network trained for the classification task was then used for the retrieval problem. A layer is chosen from the network, and a vector of features containing the model's salient features is extracted for each element of the dataset. The vectors extracted from the test set are then compared with those extracted from the train set through various metrics such as cosine distance, Euclidean distance or histogram intersection. We expect that models of the same class be represented by similar feature vectors. Then, by comparing the feature vectors, we should obtain a smaller distance if they belong to the same class and a greater one if they belong to different classes. We then use these distances to solve the retrieval task.

As a distances between extracted features, we compared the Euclidean distance, the cosine distance and the histogram intersection. The \textit{mean average precision} (mAP) obtained for the first dataset through the metrics with the final softmax layer as output layer is shown in Table~\ref{tab:retrieval_metric}. 
The cosine distance resulted in the best score, so we used that distance in the experiments.

\begin{table}[ht]
\centering
    \begin{tabular}{ l r } 
        \hline
        \textbf{Metric} & \textbf{mAP} \\ 
        \hline
        Euclidean distance & 0.976\\ 
        \textbf{Cosine distance} & \textbf{0.989}\\ 
        Histogram intersection & 0.976\\ 
        \hline
        \end{tabular}
    \caption{Comparison between different metrics when matching feature vectors extracted from the last FC layer of the GCN model for the first dataset.}
    \label{tab:retrieval_metric}
\end{table}

We also studied which layer produces a more representative feature vector, as shown in Table~\ref{tab:retrieval_output_layer}. The last layer, \textit{i.e.}, the softmax layer, performed the best mAP score equal to $0.989$. 

\begin{table}[ht]
\centering
    \begin{tabular}{l r} 
        \hline
       \textbf{Output layer} & \textbf{mAP} \\ 
        \hline
        Attention layer & 0.872\\ 
        First FC layer without ReLu & 0.902 \\ 
        First FC layer with ReLu & 0.919 \\ 
        Second FC layer & 0.948 \\ 
        \textbf{Softmax layer} & \textbf{0.989} \\ 
        \hline
    \end{tabular}
    \caption{Comparison of mAP values with the cosine metric and different output layers for the generation of feature vectors for the first dataset.}
    \label{tab:retrieval_output_layer}
\end{table}

\begin{figure}[ht]
  \includegraphics[width=\linewidth]{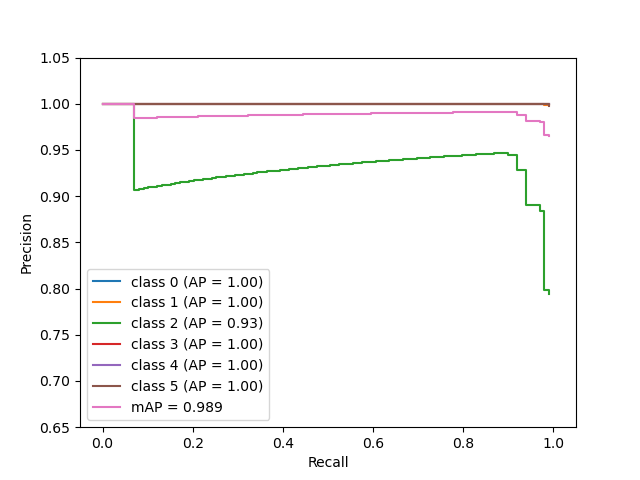}
  \caption{Precision-recall curve for the retrieval of CAD models with features vectors extracted from the softmax layer and cosine distance as metric for comparing feature vectors of the first dataset.}
  \label{fig:precision-recall}
\end{figure}

Based on the above tests, we fixed the cosine metric and the softmax layer as the output layer and generated the precision-recall curve shown in Figure~\ref{fig:precision-recall}.
Our method obtained excellent values of mAP, and only for class 2, characterized by the greater variance in the number of nodes of the graphs, it did not reach the perfect score.

\subsection{Limitations} 
The large amount of information which is present in the CAD files involves some redundancy and therefore a large amount of memory is needed to store the dataset models as graphs. To avoid having to convert the models to graphs at each run, we used the \textit{networkx} python library~\cite{networkx} to save the graph data as a \textit{.graphml} file. 
As shown in Table~\ref{tab:storage_graphs}, the size of the storage files increases almost linearly as the sum of nodes and arcs increases.

\begin{table}[ht]
\centering
    \begin{tabular}{ c c c } 
        \hline
        \# \textbf{nodes} & \# \textbf{arcs} & \textbf{space} \\ 
        \hline
        1K & 1K & 200 KB\\ 
        3K & 35K & 600 KB\\ 
        9K & 10K & 2,3 MB\\ 
        15K & 17K & 3,5 MB\\
        20K & 25K & 4,1 MB\\
        30K & 35K & 7,0 MB\\
        35K & 40K & 8,3 MB \\
        40K & 43K & 11,0 MB \\
        \hline
    \end{tabular}
    \caption{Increase in size of graph storage \textit{.graphml} files as the number of nodes and arcs increases.}
    \label{tab:storage_graphs}
\end{table}

This implies that for complex models corresponding to large graphs, heavy files are generated. In contrast to other formats such as point clouds, voxel grids and multiple 2D scans, in which the number of elements can be fixed as well as the size of the files, in the graph approach, it is not possible to regulate the size of the input data, therefore large graphs can be difficult to manage for GPU computation and RAM memory storage. 

\section{Conclusions}\label{sec:conclusion}
In this paper, we have introduced a new approach to solve the retrieval and classification problems of 3D CAD data that operates directly in native format without the need to convert the models to other extensions.
We exploited the linked structure of STEP files, which represents a standard format in the CAD domain to create a graph in which the nodes are the primitive geometric elements and the arcs are the connections between them. The graphs created are then evaluated through convolutional graph neural networks.

Since there is no dataset containing a sufficient number of 3D models in a native CAD format, we created two datasets by downloading data from the \textit{Traceparts} model library and by collecting 3D models from a real-world software modeling company and dividing them into classes. 
We have used these datasets to validate our approach and compare it with state-of-the-art methods that consider other 3D formats such as \textit{PointNet++} for the point clouds and \textit{MVCNN} for the models expressed as multiple 2D views. The obtained results demonstrate the validity of our method for both the classification task, where we obtained 100\% and 97.5\% accuracy for the relative datasets, and for the retrieval task, where we achieved results close to 100\% of mAP for the first dataset. Our approach resulted in an equivalent or better performance with \textit{MVCNN}, greatly exceeding the performance of \textit{PointNet++} on the same data expressed in the respective formats.

This preliminary work has introduced a new approach for the analysis of CAD models, but its full potential is still to be discovered. In fact, the collected datasets did not allow us to exploit the information on the hierarchy of the components of the models and in the classification task the information of the nodes was represented only by their geometric type without considering the related parameters. Theoretically, using the hierarchy of components would allow us to improve the classification of complex objects, while introducing the parameters of the nodes would make it possible to distinguish even very similar objects. In future work, it will be critical to exploit these two characteristics in order to conduct experiments on larger datasets, allowing us to maximize the potential of classification and retrieval tasks of CAD files.

\section*{Acknowledgment}
We thank the Configurators s.r.l. company for providing us with the CAD models included in the second dataset.



\printbibliography                

@article{Bo:2014,
    author = {Ding, Bo},
    year = {2014},
    month = {10},
    pages = {},
    title = {3D CAD Model Representation and Retrieval based on Hierarchical Graph},
    volume = {9},
    journal = {Journal of Software},
    doi = {10.4304/jsw.9.10.2499-2506}
}

@article{Giannini:2017,
    author = {Giannini, Franca and Lupinetti, Katia and Monti, Marina},
    title = "{Identification of Similar and Complementary Subparts in B-Rep Mechanical Models}",
    journal = {Journal of Computing and Information Science in Engineering},
    volume = {17},
    number = {4},
    year = {2017},
    month = {05},
    doi = {10.1115/1.4036120},
}

@article{TAO:2013,
title = {Partial retrieval of CAD models based on local surface region decomposition},
journal = {Computer Aided Design},
volume = {45},
number = {11},
pages = {1239-1252},
year = {2013},
doi = {https://doi.org/10.1016/j.cad.2013.05.008},
author = {Songqiao Tao and Zhengdong Huang and Lujie Ma and Shunsheng Guo and Shuting Wang and Youbai Xie},
}

@article{Li:2011,
author = {Li, Min and Zhang, Y.F. and Fuh, J.Y.H. and Qiu, Z.M.},
title = {Design reusability assessment for effective {CAD} model retrieval and reuse},
journal = {International Journal of Computer Applications in Technology},
volume = {40},
number = {1-2},
pages = {3-12},
year = {2011},
doi = {10.1504/IJCAT.2011.038546},
}

@article{You:2010,
title = {3D solid model retrieval for engineering reuse based on local feature correspondence},
author = {Chun-Fong You and Yi-Lung Tsai}, 
journal = {The International Journal of Advanced Manufacturing Technology},
volume = {46}, 
pages = {649-661},
year = {2010},
}

@article{BAI:2010,
title = {Design reuse oriented partial retrieval of {CAD} models},
journal = {Computer Aided Design},
volume = {42},
number = {12},
pages = {1069-1084},
year = {2010},
doi = {https://doi.org/10.1016/j.cad.2010.07.002},
author = {Jing Bai and Shuming Gao and Weihua Tang and Yusheng Liu and Song Guo},
}

@article{ELMEHALAWI:2003-II,
title = {A database system of mechanical components based on geometric and topological similarity. Part II: indexing, retrieval, matching, and similarity assessment},
journal = {Computer Aided Design},
volume = {35},
number = {1},
pages = {95-105},
year = {2003},
doi = {https://doi.org/10.1016/S0010-4485(01)00178-6},
author = {Mohamed El-Mehalawi and R {Allen Miller}},
}

@article{ELMEHALAWI:2003-I,
title = {A database system of mechanical components based on geometric and topological similarity. Part I: representation},
journal = {Computer Aided Design},
volume = {35},
number = {1},
pages = {83-94},
year = {2003},
doi = {https://doi.org/10.1016/S0010-4485(01)00177-4},
author = {Mohamed El-Mehalawi and R {Allen Miller}},
}

@INPROCEEDINGS{Osada:2008,  
author={Kunio Osada and Takahiko Furuya and Ryutarou Ohbuchi},  
booktitle={IEEE Int. Conf. on Shape Modeling and Applications},   
title={SHREC—08 entry: Local 2D visual features for CAD Model retrieval},   
year={2008},  
volume={},  
number={},  
pages={237-238},  
doi={10.1109/SMI.2008.4547985}
}

@inproceedings{Hua:2017,
booktitle = {Eurographics Workshop on 3D Object Retrieval},
title = {RGB-D to CAD Retrieval with ObjectNN Dataset},
author = {Hua, Binh-Son and Truong, Quang-Trung and Tran, Minh-Khoi and Pham, Quang-Hieu and Kanezaki, Asako and Lee, Tang and Chiang, HungYueh and Hsu, Winston and Li, Bo and Lu, Yijuan and Johan, Henry and Tashiro, Shoki and Aono, Masaki and Tran, Minh-Triet and Pham, Viet-Khoi and Nguyen, Hai-Dang and Nguyen, Vinh-Tiep and Tran, Quang-Thang and Phan, Thuyen V. and Truong, Bao and Do, Minh N. and Duong, Anh-Duc and Yu, Lap-Fai and Nguyen, Duc Thanh and Yeung, Sai-Kit},
year = {2017},
publisher = {The Eurographics Association},
DOI = {10.2312/3dor.20171048}
}

@article{Tao:2015,
author = {Song-qiao Tao},
title = {CAD Model Retrieval Based on Graduated Assignment Algorithm},
journal = {3D Research},
volume = {6}, 
number = {21}, 
year = 2015, 
}

@article{Tao:2011,
author = {Song-qiao Tao and Zheng-dong Huang and  Tan-guang Zheng},
title = {3D CAD model retrieval based on attributed adjacency graph matching},
year = {2011},
journal = {Computer Integrated Manufacturing System},
volume = {17},
number = {04},
pages = {0-0},
}

@INPROCEEDINGS{Ma:2019,  
author={Ma, Weifang and Wang, Peiyan and Cai, Dongfeng and Wang, Dahan},  
booktitle={IEEE Int. Conf. on Ubiquitous Computing \& Communications (IUCC) and Data Science and Computational Intelligence (DSCI) and Smart Computing, Networking and Services (SmartCNS)},   
title={Research on 3D CAD Model Retrieval Algorithm Based on Global and Local Similarity}, 
year={2019},  
volume={},  
number={},  
pages={349-355},  doi={10.1109/IUCC/DSCI/SmartCNS.2019.00085}
}

@misc (Traceparts,
    AUTHOR ={Trace Group},
	TITLE = {Traceparts online library, Product Content Everywhere},
	note = {Online available: https://www.traceparts.com/it.}
)

@ARTICLE (CADClassification1,
    AUTHOR = {Cheuk Yiu Ip},
	TITLE = {Automatic Classification of CAD Models},
	YEAR = {2005}
)

@ARTICLE (zernike,
    AUTHOR = {M. Novotni and R. Klein},
	TITLE = {Shape retrieval using 3D zernike descriptors},
	YEAR = {2004},
	DOI = {36(11):1047–1062}
)

@ARTICLE (MRG,
    AUTHOR = {M. Hilaga and Y. Shinagawa and T. Kohmura and T. L. Kunii},
	TITLE = {Topology matching for fully automatic similarity estimation of 3D shapes},
	YEAR = {2001},
	JOURNAL = {In SIGGRAPH}, 
	pages = {203-212}, 
)

@ARTICLE (CADClassification3,
    AUTHOR = {Cheuk Yiu Ip and William C. Regli},
	TITLE = {Content-Based Classification of CAD Models with Supervised Learning}, 
	journal = {Computer Aided Design and Applications},
	YEAR = {2005}
)

@article{ShapeDistribution,
    AUTHOR = {Osada R. and Funkhouser T. and Chazelle B. and Dobkin D.},
	TITLE = {Shape distributions}, 
	journal = {ACM Transactions on Graphics},
	YEAR = {2002}
}

@ARTICLE (CADClassification4,
    AUTHOR = {Chen and Ding-Yun and Tian, Xiao-Pei and Shen and Yu-Te and Ouhyoung and Ming},
	TITLE = {On Visual Similarity Based 3D Model Retrieval},
	YEAR = {2003}
)

@ARTICLE (CADClassification2,
    AUTHOR = {Qin and Fw. and Li and Ly. and Gao and Sm. et al},
	TITLE = {A deep learning approach to the classification of 3D CAD models},
	YEAR = {2014},
	NOTE = {https://doi.org/10.1631/jzus.C1300185}
)

@ARTICLE (Frequency,
    AUTHOR = {Li and Wenjin and Mac and Gary and Tsoutsos and Nektarios Georgios and Gupta, Nikhil and Karri, Ramesh},
	TITLE = {Computer aided design (CAD) model search and retrieval using frequency domain file conversion},
	YEAR = {2020}
)

@ARTICLE (VoxNet,
    AUTHOR = {D. Maturana and S. Scherer},
	TITLE = {VoxNet: A 3D Convolutional Neural Network for real-time object recognition},
	YEAR = {2015},
	DOI = {10.1109/IROS.2015.7353481},
	JOURNAL ={IEEE/RSJ International Conference on Intelligent Robots and Systems (IROS)}
)

@ARTICLE (ROCA,
    AUTHOR = {Gümeli and Can and Dai and Angela and Nießner and Matthias},
	TITLE = {ROCA: Robust CAD Model Retrieval and Alignment from a Single Image},
	YEAR = {2021}
)

@ARTICLE (LightNet,
    AUTHOR = {Shuaifeng Zhi and Yongxiang Liu and Xiang Li and Yulan Guo},
	TITLE = {Toward real-time 3D object recognition: A lightweight volumetric CNN framework using multitask learning},
	YEAR = {2018}
)

@ARTICLE (PointNet,
    AUTHOR = {Qi and Charles R. and Su and Hao and Mo and Kaichun and Guibas, Leonidas J},
	TITLE = {PointNet: Deep Learning on Point Sets for 3D Classification and Segmentation},
	YEAR = {2016}
)

@inproceedings{PointNet++,
 author = {Qi, Charles Ruizhongtai and Yi, Li and Su, Hao and Guibas, Leonidas J},
 booktitle = {Advances in Neural Information Processing Systems},
 editor = {I. Guyon and U. Von Luxburg and S. Bengio and H. Wallach and R. Fergus and S. Vishwanathan and R. Garnett},
 pages = {},
 publisher = {Curran Associates, Inc.},
 title = {PointNet++: Deep Hierarchical Feature Learning on Point Sets in a Metric Space},
 volume = {30},
 year = {2017}
}

@ARTICLE (KnowledgeGraph,
    AUTHOR = {W. Nie and Y. Wang and D. Song and W. Li},
	TITLE = {3D Model Retrieval Based on a 3D Shape Knowledge Graph},
	YEAR = {2020},
	DOI = {10.1109/ACCESS.2020.3013595}
)

@ARTICLE (Kd-Network,
    AUTHOR = {Klokov and Roman and Lempitsky and Victor},
	TITLE = {Escape from Cells: Deep Kd-Networks for the Recognition of 3D Point Cloud Models},
	YEAR = {2017},
)

@ARTICLE (MVCNN,
    AUTHOR = {Su and Hang and Maji and Subhransu and Kalogerakis and Evangelos and Learned-Miller and Erik},
	TITLE = {Multi-view Convolutional Neural Networks for 3D Shape Recognition},
	YEAR = {2015}
)

@ARTICLE (PANORAMA,
    AUTHOR = {Sfikas and Konstantinos and Theoharis and T and Pratikakis and Ioannis},
	TITLE = {Exploiting the PANORAMA Representation for Convolutional Neural Network Classification and Retrieval},
	YEAR = {2017},
	DOI = {10.2312/3dor.20171045}
)

@ARTICLE (PANORAMA_ENN,
    AUTHOR = {Konstantinos Sfikas and Ioannis Pratikakis and Theoharis Theoharis},
	TITLE = {Ensemble of PANORAMA-based convolutional neural networks for 3D model classification and retrieval},
	YEAR = {2018},
)

@ARTICLE (RotationNet,
    AUTHOR = {Kanezaki and Asako and Matsushita, Yasuyuki and Nishida and Yoshifumi},
	TITLE = {RotationNet: Joint Object Categorization and Pose Estimation Using Multiviews from Unsupervised Viewpoints},
	YEAR = {2016}
)

@ARTICLE (CADSketch1,
    AUTHOR = {Hou S. and Ramani K},
	TITLE = {Classifier combination for sketch-based 3D part retrieval
    Comput. Graph},
	YEAR = {2007}
)

@ARTICLE (CADSketch2,
    AUTHOR = {Hou S. and Ramani K},
	TITLE = {Sketch-based 3D engineering part class browsing and retrieval SBM},
	YEAR = {2006},
)

@ARTICLE (CADSketch3,
    AUTHOR = {Feiwei Qin and Shi Qi and Shuming Gao and Jing Bai},
	TITLE = {3D CAD model retrieval based on sketch and unsupervised variational autoencoder},
	YEAR = {2022}
)

@ARTICLE (networkx,
    AUTHOR = {Aric Hagberg <hagberg@lanl.gov> Dan Schult <dschult@colgate.edu> Pieter Swart <swart@lanl.gov>},
	TITLE = {NetworkX Python Library},
	note = {https://github.com/networkx/networkx}
)

@ARTICLE (SimGNN,
    AUTHOR = {Yunsheng Bai and Hao Ding and Song Bian and Ting Chen and Yizhou Sun and Wei Wang},
	TITLE = {SimGNN: A Neural Network Approach to Fast Graph Similarity Computation},
)

@ARTICLE{ShapeNet,
    AUTHOR = {Angel X. Chang and Thomas Funkhouser and Leonidas Guibas and Pat Hanrahan and Qixing Huang and Zimo Li and Silvio Savarese and Manolis Savva and Shuran Song and Hao Su and Jianxiong Xiao and Li Yi and Fisher Yu}, 
    TITLE = {ShapeNet: An Information-Rich 3D Model Repository}, 
    note = {Technical Report arXiv:1512.03012 Stanford University - Princeton University - Toyota Technological Institute at Chicago 2015}
}

@ARTICLE (ModelNet,
    AUTHOR = {Z. Wu and S. Song and A. Khosla and F. Yu and L. Zhang and X. Tang and J. Xiao},
	TITLE = {3D shapenets: A deep representation for volumetric shapes},
	JOURNAL = {IEEE Conference on Computer Vision and Pattern Recognition}, 
	pages = {1912–1920}, 
	year = {2015},
)

@misc (Configuratori,
	AUTHOR = {Configuratori Srl},
	TITLE = {Configuratori},
	note = {website: https://www.configuratori.com/}
)


\end{document}